\documentclass[reqno]{amsart}
\usepackage{amsmath}
\usepackage{amsopn}
\usepackage{amssymb}
\usepackage{amsfonts}
\usepackage{graphicx}
\usepackage{boldline}
\usepackage{graphicx,booktabs}
\usepackage[table]{xcolor}
\usepackage{subfigure}
\usepackage{fancyhdr}
\usepackage[section]{placeins}
\usepackage[autostyle, english = american]{csquotes}
\usepackage{multicol}
\usepackage{booktabs,dcolumn,caption}
\usepackage{url}
\usepackage[colorlinks,allcolors=blue]{hyperref}
\numberwithin{table}{section}
\usepackage{pdflscape} 
\usepackage{multirow}
\usepackage{longtable}
\usepackage{rotating}
\usepackage{slashbox}
\usepackage{hhline}
\numberwithin{figure}{section}

\newcommand{\rr}{\raggedright}
\newcommand{\tn}{\tabularnewline}
\newcommand{\rl}{\raggedleft}
\newcommand{\rc}{\centering}
\newcolumntype{R}[1]{>{\raggedleft\let\newline\\\arraybackslash\hspace{0pt}}m{#1}}
\newcolumntype{L}[1]{>{\raggedright\let\newline\\\arraybackslash\hspace{0pt}}m{#1}}
\newcolumntype{C}[1]{>{\centering\let\newline\\\arraybackslash\hspace{0pt}}m{#1}}

\usepackage[foot]{amsaddr}

\definecolor{maroon}{cmyk}{0,0.87,0.68,0.32}

\begin{document}

\title{\textbf{Informational Space of Meaning \\ for Scientific Texts} }

\author{N. S\"uzen$^{1}$}
\author{E. M. Mirkes$^{1}$}
\author{A. N. Gorban$^{1}$}

\address{$^{1}$School of Mathematics and Actuarial Science, University of Leicester, Leicester LE1 7RH, UK}

\email{ns433@le.ac.uk (N. S\"uzen)}
\email{em322@le.ac.uk (E.M. Mirkes)}
\email{ag153@le.ac.uk (A.N. Gorban)}

\maketitle

\begin{abstract}

In Natural Language Processing, automatic extracting the meaning of texts constitutes an important problem. Our focus is the computational analysis of meaning of short scientific texts (abstracts or brief reports). In this paper, a vector space model is developed for quantifying the meaning of words and texts. We introduce the \textit{Meaning Space}, in which the meaning of a word is represented by a vector of \textit{Relative Information Gain} (RIG) about the subject categories that the text belongs to, which can be obtained from observing the word in the text.

This new approach is applied to construct the Meaning Space based on Leicester Scientific Corpus (LSC) and Leicester Scientific Dictionary-Core (LScDC). The LSC is a scientific corpus of 1,673,350 abstracts and the LScDC is a scientific dictionary which words are extracted from the LSC. Each text in the LSC belongs to at least one of 252 subject categories of Web of Science (WoS). These categories are used in construction of vectors of information gains.   

The Meaning Space is described and statistically analysed for the LSC with the LScDC. The usefulness of the proposed representation technique is evaluated through top-ranked words in each category. The most informative $n$ words are ordered. We demonstrated that RIG-based word ranking is much more useful than ranking based on raw word frequency in determining the science-specific meaning and importance of a word.  The proposed model based on RIG is shown to have ability to stand out topic-specific words in subject categories. The most informative words are presented for 252 subject categories. The new scientific dictionary and the $103,998\times 252$ Word-Category RIG Matrix are available online.
   
Analysis of the Meaning Space provides us with a tool to further explore quantifying the meaning of a text using more complex and context-dependent meaning models that use co-occurrence of words and their combinations.

\vspace{5mm}


\end{abstract}

\tableofcontents

\section{Introduction}\label{intro}
\subsection{The Problem and Preliminaries\nopunct}\hspace*{\fill} \\\\

Automatic analysis of text meaning is one of the main problems in Natural Language Processing (NLP). This work is focused on the computational analysis of the meaning of short scientific texts (abstracts or brief reports). The starting point is a combination of a simple Bag of Word (BoW) model with the holistic approach to the text meaning: the text is considered as a collection of words, the meaning of the text is hidden in a situation of use, which is evaluated as a whole. A space of meaning for words is created  from the analysis of situations of their use  and then, after detailed analysis of this space (including dimensionality reduction and clustering) we will return to the texts and introduce more complex models including words co-occurrence analysis, combination of word's meaning etc.

First of all, we have to consider  the ``meaning of meaning''. This is an extremely deeply discussed topic, since antiquity till modern time  (see, for example, \cite{ogden,putnam,carston,michaelis}), but the consensus is still on the way. We start from the  Wittgenstein formulation: ``Meaning is use'' or, in more detail, ``For a large class of cases – though not for all – in which we employ the word `meaning' it can be defined thus: the meaning of a word is its use in the language''\cite[\S 43]{Wittgenstein2009}.  

\begin{figure} 
\centering
 \includegraphics[width=\textwidth]{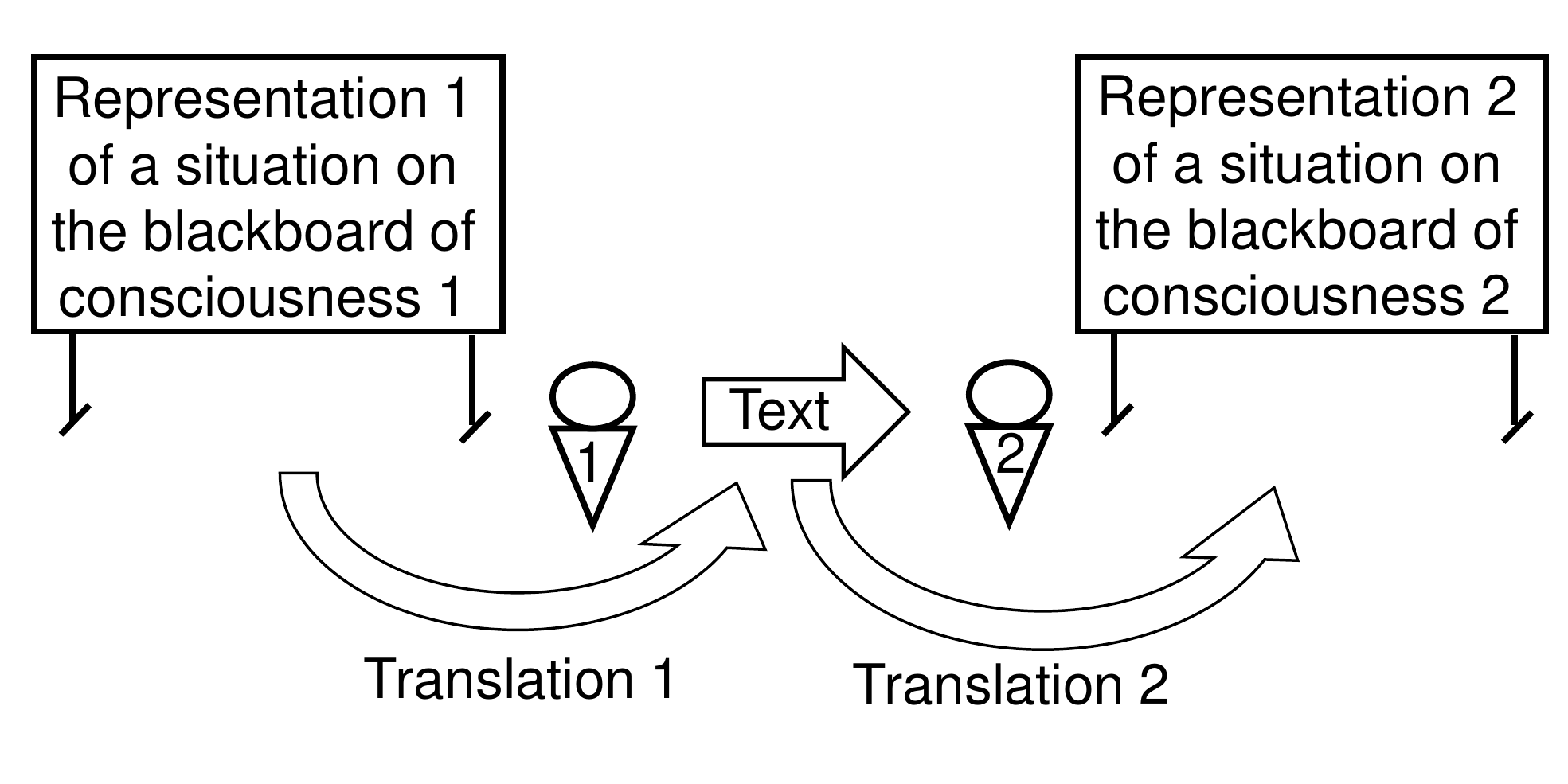}
   \caption{The idealised scheme of the act of communication. There is a representation of a situation  on the sender's `blackboard of the consciousness' (a representation 1 of a situation 1). A text related to this situation is generated by the sender (translation 1). This text is transmitted to the receiver and transformed by him into  a representation of a situation (representation 2 of a situation 2). 
The situation can be the situation in a real world, the imaginary situation in a possible world, an impossible situation in an impossible world, a chimeric situation combined from several possible or imaginary situation, and so on. We do not study the relations of representation to reality, but only consider the chain: $\mbox{Representation 1}\to \mbox{Text} \to \mbox{Representation 2}$.}
  \label{fig:communication} 
\end{figure}

This idea was widely discussed. This paper  aims to propose an approach to computational analysis of meaning for a large family of texts. The texts we work with (abstracts or brief reports) have well defined dominant communicative function: this is the representative function. An elementary basic scheme of the correspondent act of communication is presented in Figure ~\ref{fig:communication}. In this scheme, we see two representations of the situation on the blackboard of consciousness: the sender's representation of the situation and the receiver's representation of the situation. Moreover, the represented situations can be different. In fact, they are always different, and special tools are invented and used to make them as close as possible when necessary.  The situations are not compulsory real. They   can be real, possibly real or imaginary, and even impossible.

It is necessary to stress that the sender's and the receiver's representations never coincide and:
\begin{itemize}
\item Do not represent any   situation  `in detail' and, therefore, can represent parts (or projections, let us recall the  Plato's Cave allegory) of many different situations at the same time;
\item Can include internal contradictions and, therefore, can represent nothing possible in reality;
\item Can partially represent different situations, that is, can be `chimeric' combinations of different possible real or imaginary situations.
\end{itemize}
There are two `translation' operations in the scheme Figure ~\ref{fig:communication}: (i) from the sender's representation of the situation to the  text of the message and (ii) from the text of the message to the the receiver's representation. Both these operations depend on much wider context of the communication including experience of the sender and receiver. Of course, the standard scientific communication assumes that there may be many receivers and the sender can be not a single person. This one-to-many or even many-to-many communication adds more situations and representations and may also add some less trivial multi-agent structures with additional communication channels.

We consider using the language to transmit information about the represented situations (Figure ~\ref{fig:communication}) and neglect many other uses of the language, from military orders to psychological manipulations. The scheme of the act of communication  (Figure ~\ref{fig:communication}) includes just very basic elements and can be elaborated in much more detail. Here we should refer to the classical works of G.P. Shchedrovitsky \cite{Shchedrovitsky1974,Shchedrovitsky1995} and J. Habermas \cite{Habermas1984,Habermas1987}. For our purposes in this work, the basic scheme (Figure ~\ref{fig:communication}) is sufficient.

According to Shchedrovitsky \cite{Shchedrovitsky1974}, at the level of `simple communication' there is no `meaning' different from the processes of understanding themselves, which correlate and connect the elements of the text message with each other and with the elements of the situation being restored.

Meaning, for our analysis, is hidden in the relationship between the representation of situations on the `blackboard of the consciousness' and the texts of the messages.  That is, a formal analysis of meaning requires the formalisation of translation operations presented in the scheme of a communication act (Figure ~\ref{fig:communication}). Moreover, we can state that we understand the {\em meaning of meaning} if and only if we can produce such a translation. This translation is context-dependent, the unique experience of the sender and the receiver is involved in this context, so the task  of ``reproducing the translation'' is not fully feasible. Moreover, understanding can be represented as a reflexive game \cite{Lefebvre2010} with different levels (The sender prepares a message taking into account the experience of the receiver, his goals and tools, and guesses that the receiver takes into account the experience of the sender, his goals and tools, and... Analogously, the receiver tries to understand the message taking into account..., etc.) 

The relation between the text and the representation of the situation cannot be considered as a bijection (both for sender's and receiver's representation). It is many-to-many correspondence: each text corresponds to many situations and each situation can have many representing texts. Moreover, the further consistent formalisation requires the notion of {\em fuzzy many-to-many correspondence} elaborated for relational databases \cite{Chen2012}.

According to Mel'čuk \cite{Melchuk1970,Melchuk2012,Melchuk2013,Melchuk2015}, the natural language is ``{\em the meaning to text and text to meaning transformer}''. He accepted a very strong hypothesis that we are able to describe meaning in a special {\em semantic language}. 

We prefer to be more flexible at this point and characterise a situation ``behind the text'' by a set of attributes, the method of this characterisation can be changed and does not give a unique and exhaustive presentation of it.

Despite  the multiplicity of possible translations, creating of a plausible translation (one of many possible versions) and description of the cloud of such versions of translation  can be  challenging. This problem resembles the translation problem for natural languages. Now, after impressive progress of machine translation, it seems to be a very attractive idea to apply the modern machine learning tools and encoder-decoder approach \cite{ChoAtAl2014} to analysis and simulation the  translation operations  $\mbox{Representation 1}\to \mbox{Text} \to \mbox{Representation 2}$ (Figure ~\ref{fig:communication}). 

Huge digitized collections of texts exist and are available online. On the contrary, unfortunately, there is no generally accepted common tools for working directly with  representations of situations. Various philosophical and logical aspects of this problem were discussed previously by many authors (see, for example, the book  `Representation and reality' \cite{Putnam1988}). 

We do not have a universal toolbox for work with all representations of situations and cannot propose  a general solution to this problem. Such a solution, perhaps, is impossible in a finite closed form despite many efforts over decades.  Our goal is more modest. We will  provide  computational analysis of relations between texts of messages and representations of situations for a large collection of brief scientific texts. To do this, these representations must be standardised, at least in part, and expressed in the form of diagrams, specially organized texts or other means. 

The simplest approach is to replace the situation representations with the values of some attributes. This approach is not only the simplest, but also quite universal. Many forms of more specific descriptions of situations can be transformed into vectors of attributes. The choice of attributes can be very broad. A classical collection of examples is provided by various version of   sentiment analysis.  We aim to provide another basic example specific to scientific texts: a list of scientific subject categories that the text belongs to.  The list of 252 possible categories is generally accepted and standardised by WoS. Of course, the variety of possible extensions and modifications of the set of attributes characterising the situation is virtually infinite.

\begin{figure} 
\centering
 \includegraphics[width=0.8\textwidth]{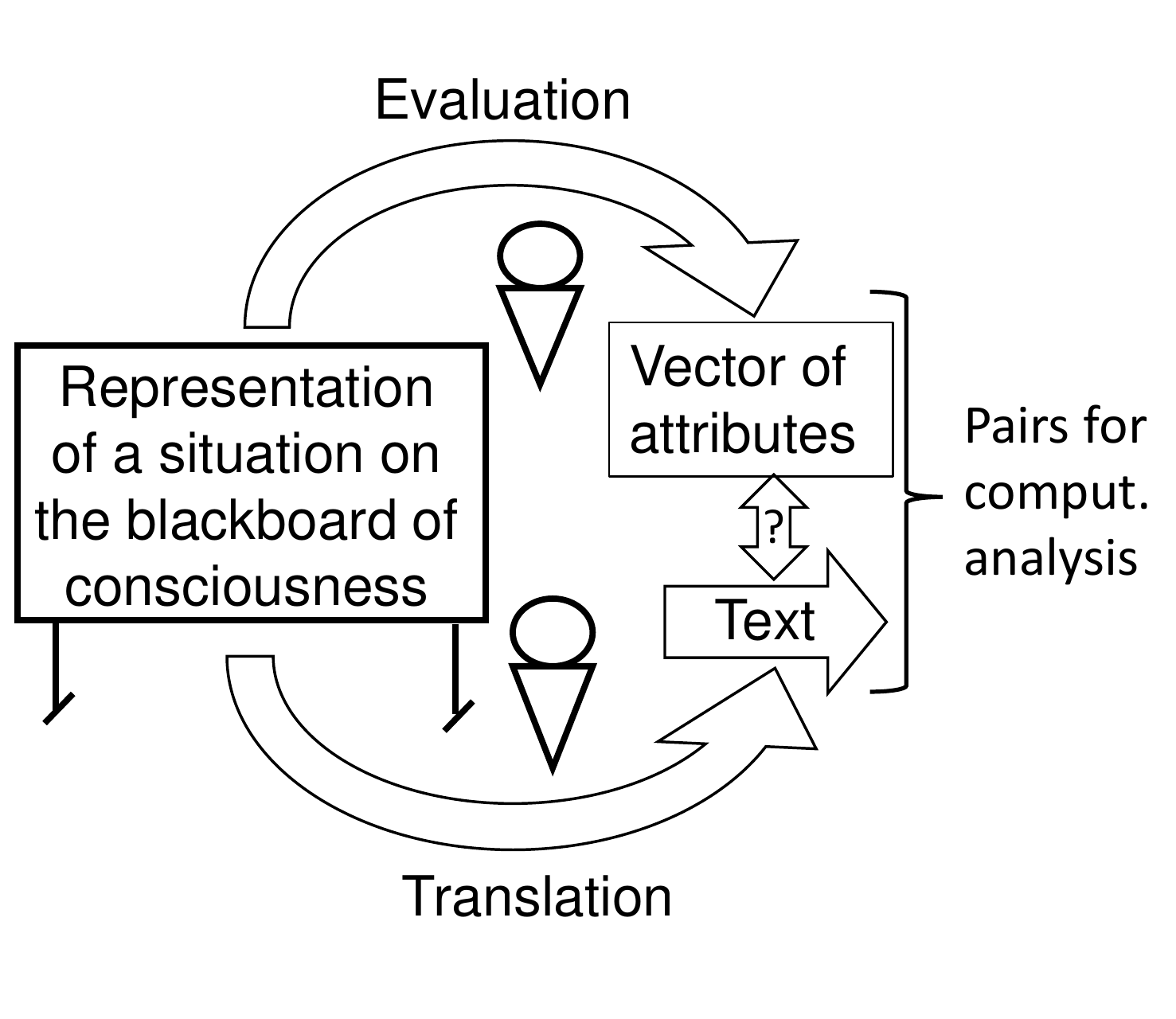}
   \caption{Parallel (sender's) generation of the learning set for computational analysis and quantification of meaning. One of the main problems is the relationship between the content of texts and the evaluated attributes of the situation.}
  \label{fig:trainingset} 
\end{figure}

\begin{figure} 
\centering
 \includegraphics[width=0.85\textwidth]{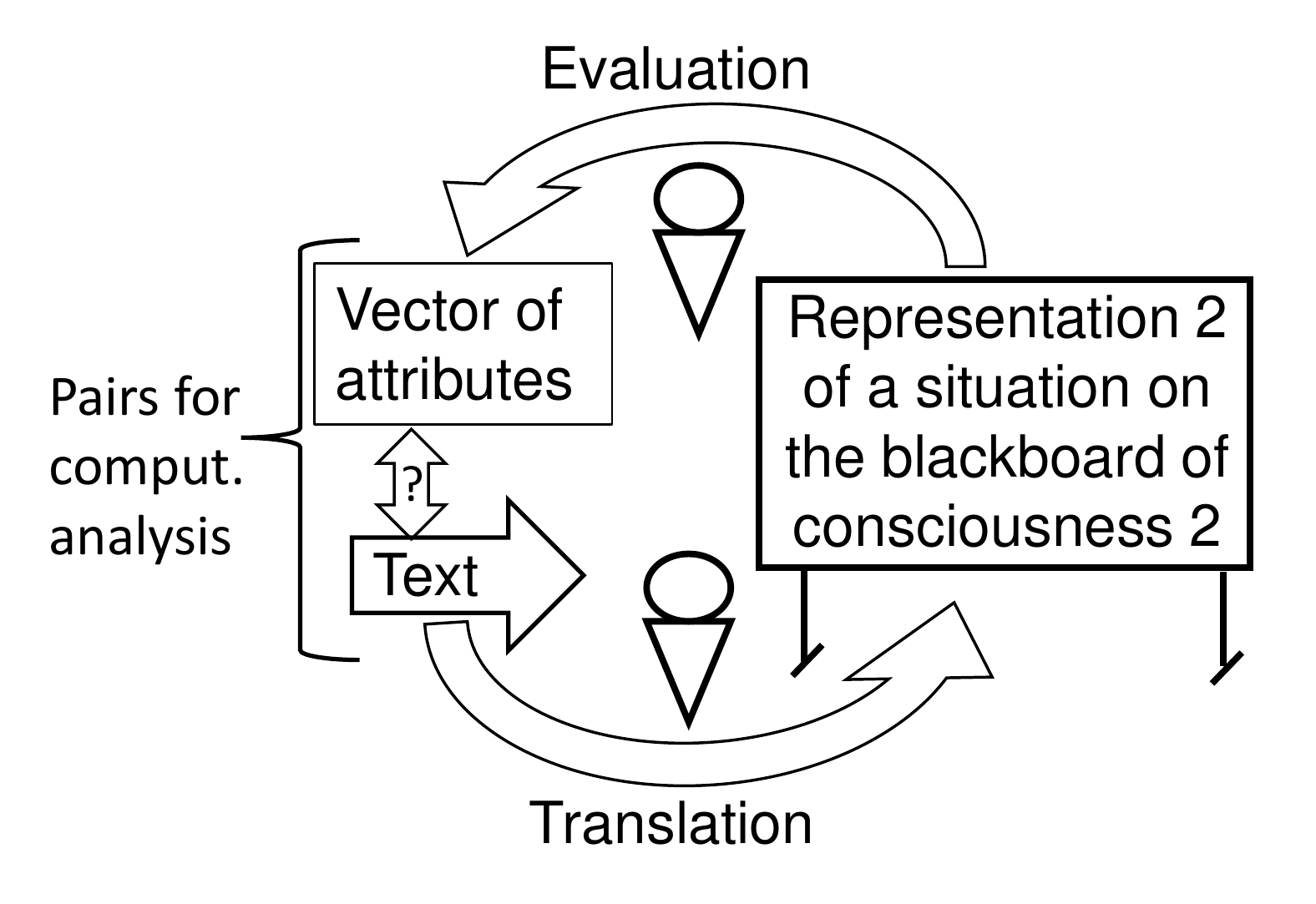}
   \caption{Antiparallel (receiver's) generation of the learning set for computational analysis and quantification of meaning.  One of the main problems is the relationship between the content of texts and the evaluated attributes of the situation.}
  \label{fig:trainingset2} 
\end{figure}

Initially, in the act of communication, the situation is not represented by a universally conventional set of attributes. The introduction of attributes is an additional operation external to the communication and is not included in the scheme of simple communication (Figure ~\ref{fig:communication}). Moreover, an additional operation must be performed for the selected set of attributes: evaluating their values. This operation can be done either on the sender's side (Figure ~\ref{fig:trainingset}), the receiver's side (Figure ~\ref{fig:trainingset2}), or by combinations of these approaches.  For example, categorisation of a brief scientific texts is a result of combined efforts: the authors select the categories by their choice of the journal, of the keywords, or  by the pointing the categories directly, then the editors can have their own choice, then WoS can finalise the list of subject categories for this text.

For most information services, the choice of subject categories is the result of an understanding of the text by many agents and conflicts of understanding are possible. Even on a famous and very `liberal' preprint server, arXiv, moderators can sometimes change the category selected by the authors. For example, an author may decide that his paper belong to the category `condensed matter', whereas the moderator may look through the paper and understand that the main category is not `condensed matter'  research but rather `nonlinear science' (this was a real life example). This simple example is important because it demonstrates that the content of the text may differ from its meaning: the text contained an explicit reference to  `condensed matter', but this {\em content} was questioned by the moderator, since in his {\em understanding} the research refers mainly to nonlinear science, and not to condensed matter.  There are important differences between the concepts of `meaning' and `content'  \cite{Shchedrovitsky1974}, which are often confused (just as understanding the situation behind the text is often confused with recognising the content of the text).

In the general case,   agents who are looking for the meaning of the text can be both humans or computer systems. The latter understand the text in the sense that they define the attributes of the situation behind the text. In our analysis below, the starting point is the combination of the text with the list of the subject categories the text belongs to (1,673,350 abstracts and 252 categories).

The core idea of this approach goes back to the lexical approach of Sir Francis  Galton. He selected the personality-descriptive terms and stated the problem of their interrelations for real persons. This work was continued by Thurstone  \cite{Thurstone1934}. He selected sixty adjectives (attributes of a person that are in common use). The respondents (1300 persons) were asked to imagine a person they knew well and to  select   the adjectives that can best describe this person. That is, a person was described by a 60-dimensional Boolean vector. The coordinates correspond to the attributes, the value is 1, if the attribute was selected to characterise the person, and 0 otherwise. Factor analysis gave five factors. After many years of development and discussions, the modern five-factor personality model became one of the common tools in psychodiagnosis \cite{McCrae04,FehrmanEgGorMir2019}. 

In psycholinguistics, Osgood with co-workers \cite{osgood2} used a similar approach for creation of  the 3D space of meaning by extraction of three `coordinates of meaning' from the evaluation of the `affective meaning' of words (objects) by people. These three coordinates are three extracted factors: Evaluation, Potency, and Activity.  Of course, the researches  started from many different scales and these three were extracted by factor analysis. 

Galton, Thurstone, Osgood and their followers asked respondents to evaluate a single object or person. Nevertheless, we can guess that these  evaluations were related to some situations with this single object or person, not just to an isolated abstract object. 
The people evaluated not the abstract `terms' but the psychologically meaningful situations behind these terms. These situations were the sources of the `affective meaning' or the personality evaluations. For example, if we evaluate a person as accurate, reliable, and friendly then we have in mind some situations where these properties were demonstrated. The same, if we evaluate a `dog' as strong, good, and active (or, say, weak, bad and active), we have in mind a dominant situation which we associate with a dog.

The `affective meaning' or psychological properties do not seem reasonable tools for description of the situations behind scientific texts. In our world of abstracts and brief scientific texts,  there is another, scientifically specific description of the situation of use -- the categories of the text. There are 252 Web of Science (WoS) categories for the Leicester Scientific Corpus (LSC), to which the text could belong (see Table \ref{table:categories}). These categories can intersect: a text can belong to several categories. We will use these 252 binary attributes (the text belongs to a given category, or does not belong to it) as a basic description of the situation.

The categories evaluate the situation (the research area) related to the text as a whole, not as a results of the combination of words' meaning. In this holistic approach, we define the general meaning of a word in short scientific texts as the information that the use of this word in  texts carries about the categories to which these texts belong. More specifically, that is the {\em Relative Information Gain} (RIG) about the subject categories that the text belongs to, which can be obtained from observing the word in the text. 
 This RIG is defined for each word and each category. Thus, a meaning of a word is represented by a 252-dimensional vector of RIGs. We create and study this space of meanings.

\begin{enumerate}
\item We intend to analyse the meaning of scientific texts.
\item We considered the specific world of the texts - the abstracts of research papers.
\item We narrowed the whole word of abstracts to a sample: 1,673,350 texts from the Leicester Scientific Corpus (LSC) \cite{LSCn}. 
\item We characterize the research situations behind the text by 252 binary attributes -- the scientific WoS categories. 
\end{enumerate}

Thus, to follow this way, we need a triad: dictionary, texts, and  multidimensional evaluation of the situation of use presented by the categories. We prepared the first two elements in the previous work and the results are available online \cite{our,LSC,LScDC}. Now, we start to create the space of meaning. 

In our case study, we employed very simple attributes for description of the text usage situation, the research subject categories of the text. This list of attributes can be modified and extended. The level of detail of the Meaning Space can vary greatly within the framework of the proposed approach.

\subsection{Approaches to Meaning of Words\nopunct}\hspace*{\fill} \\\\

Let us take a quick look at some relevant ideas about quantifying the meaning of words. Quantifying the meanings of words in a metric space might be used to measure the meanings of texts in the same metric as a BoW. A key issue in understanding the meaning of texts is to use a precise metric based on words' meanings. In classical psycholinguistic studies it is common to allocate words in a metric space based on their semantic connotations \cite{miller,samsonovic}. Semantic space model is a representation technique where each word is assigned to a point in high dimensional vector space. Vector Space Model (VSM) is one of the most attractive models for researchers since it makes semantics computable \cite{turney}. Osgood hypothesised 3-dimensional semantic space to quantify connotative meanings in his theory of \textit{Semantic Differential} concerning psychological and behavioural aspects  \cite{osgood2,osgood1,anderson}. The semantic space in his work was built by, in his words, `three orthogonal bipolar dimensions': Evaluation (E), Potency (P) and Activity (A) where each word is uniquely located on. Following this method of semantic differential, many studies have been attempted by both psychologists and linguists to identify new dimensions of semantic space and to measure the meaning  \cite{osgood2,samsonovic,bentler,lowe}. The structures of semantic spaces are constructed differently by various researchers.

From the perspective of distributional linguistics, a semantic space model is a representation technique for contextual similarity of words by their co-occurrence counts. Distributional hypothesis was introduced by Harris \cite{harris} and Distributional Semantic Models (DSM) were then proposed to represent word semantic by distributional vectors \cite{turney,sahlgren,lapesa,schutze}. This idea claims that words' similarity can be characterised by their distribution of contexts \cite{firth,lapata4}. The model offers that each word is represented by distribution of its contexts and the distribution of contexts can be learnt from the co-occurance. The axes in the space are determined by local word co-occurrences and the similarity of a word is measured by its position found by counting co-occurrences to other words in this semantic space \cite{lowe2}. This means that a word's distributional context is represented by a vector of co-occurrences with other context words in a window, where a window can be a certain number of words or lemmas (e.g. words, phrases, sentence, paragraph or document). 

Researchers in cognitive studies and information retrieval noted that usage of raw co-occurrence counts is problematic as semantic similarity will have frequency bias \cite{lowe}. It is proposed that degrees of similarity between word occurances can be assigned. Different approaches are used to avoid this problem by weighining of elements of the vector. Latent Semantic Analysis (LSA) is one of vector space models in NLP, in particular DSM, for estimating and representing the meaning of word based on statistical computations \cite{landauer2,landauer}. In LSA, word senses (or meanings) are approximated in high dimensional space by its effect on the meaning of contexts in which it occurs \cite{landauer3}. Relationship between texts based on their words and relationship between words based on their appearances in texts are analysed simultaneously in order to extract relations of words in terms of their contexts. 

LSA has been used for an adequate theory of word meaning by researchers from a wide range of research areas including psychology, philosophy, linguistics, information retrieval and cognitive science \cite{mccarty,murphy,lopukhina}. In cognitive science, the focus is to model human memory by activating the meaning potentials by other words in the context under the assumption that cognitive components of meaning of word are linked in a semantic-based network and changes dynamically \cite{hanks}. It is assumed that human knowledge acquisition actually follows the same process that LSA does: checking events in their internal and external environments and deriving the knowledge from a high dimensional semantic space by a procedure like dimension reduction \cite{kintsch,dumais}. Here, the semantic space is used as a basis for all cognitive processing. Although LSA supplies a usefull simulation of human cognitive processes, it is argued that LSA knowledge base does not provide a complete modelling of cognition \cite{kintsch}. There are limitations in modification of context and updating the model of semantic dimensions -- in this knowledge base -- which are characteristic of analytic thinking and dynamic structure of the human cognitive processes \cite{evangelopoulos}. Even if this problem is solved, there are other fundamental semantic problems for LSA such as polysemous words. In LSA, when each word is represented as a single context-free vector in the semantic space, different meanings or senses of a word is not taken into account \cite{landauer2}. 

This problem matches the task of characterisation of word meaning by its dictionary senses in Word Sense Disambiguation (WSD). WSD is defined as a task to determine the word sense (meaning) by the use of the word in a context in NLP and Machine Learning. In traditional word sense studies, meaning of a word is characterised by mutually disjoint senses covered in dictionaries as the best fit to the its dictionary senses \cite{erk}. By both linguistics and psychologists, it has been argued that clear distinctions of senses can be difficult in certain contexts due to fluctuations of meaning in context \cite{murphy, hanks, pustejovski}, especially for polysemous words. Hanks (lexigropher) pointed this problem in his paper where he questioned `Do word meaning exist?' \cite{hanks} as: 
\begin{quotation}
\textit{``...words have meaning potentials, rather than just meaning. The meaning potential of each word is made up of a number of components, which may be activated cognitively by other words in the context in which it is used. These cognitive components are linked in a network which provides the whole semantic base of the language, with enormous dynamic potential for saying new things and relating the unknown to the known.''} 
\end{quotation}

The problem of `fluctuation of meaning in context' is also important in theories of mental representation of word senses in Psychology. This was very well discussed by Kintsch \cite{landauer2,kintsch,kintsch2, kintsch3} who stressed the complexity of representation of polygamous words into a single vector in the semantic space in LSA. He questioned `How is the meaning of words represented in the mind?' and discussed the problem in the aspects of `mental lexicon' and `generative lexicon' approaches to the representation of meaning \cite{pustejovski}. He came up with the result that both mental lexicon and generative lexicon approaches have limitations in representation of the meanings when word meanings are constructed by their explicit definitions due to multiple senses of words and the flexibility of word meanings. He then discussed the implicit way to define word meaning: relations of the word to other words in the context. According to his research, LSA allows us to modify word meaning by situating the meaning as a vector in high-dimensional semantic space. In this case, the full meaning of the word is not defined, but it is explained in a relational system by only its semantic relationships with other words. He argued that standard composition rule for vectors in LSA does not distinguish the different meanings of a word; therefore, word meanings should be modified  according to the different context -- where it appears in -- by context-sensitive composition algorithms.

Polysemy is one of the characteristics of words in all natural languages. Psycholinguistic studies approach this phenomenon to answer questions of how to represent multiple senses in mental lexicon and how to activate senses during language comprehension \cite{lopukhina}. The mental lexicon here can be considered as a mental repertoire containing the list of meanings or senses in the mind. Linguistics proposed several approaches for sense representation in mental lexicon, basically classified as seperate sense representation and single core representation. Even though some polysemy studies argue the discreteness of sense storage in mental lexicon \cite{murphy,murphy2,murphy3}, the majority of studies suggests that polysemous senses can overlap in their mental representations \cite{erk,frazier,frazier2,pickering,farisson}.

Moreover, polysemy of words is one of major focuses in distributional semantics and it is yet to be studied \cite{erk5, van3, van4}. Some researches in distributional semantics have made modelling the differences of meanings in two occurrences of a word in different contexts possible by developing specialized models for word meaning \cite{erk3,erk2,lapata2,thater}. Such methods do not approach to word meaning by considering disjoint senses. Alternative models were purposed in which word meaning is not just extracted by pre-defined senses, but from the links between words and their window-based context words. To extract `contexeualised meaning' of a word or a set of words, co-occurance vectors are constructed and vector operations are used \cite{erk2,lapata2,lapata,lapata3}. A probabilistic method that word meaning is modelled as a probability distribution over latent dimensions (senses) was applied by \cite{lapata2,lapata}. Contexualized meaning was build as a change in original sense distribution. Cruys, Poibeau and Korhonen then purposed a model in which latent space is used to identify important dimensions for a context and adapt to vector of words constructed by the dependency relations with window-based context words \cite{van}.
 
\subsection{Quantification of Meaning and Space of Meaning for Scientific Texts\nopunct}\label{quan}\hspace*{\fill} \\\\

In academic disciplines, the notion of meaning of a word was analysed in many works ranging from psychology to linguistics, philosophy to pedagogy and computer science \cite{van2,paradis}. Technical innovations in computerised methods and extensive psycholinguistic and neurolinguistic experiments have made investigating word meanings in different perspectives and linking between the language and cognition, and the language in people's mind possible. 

There is no unique way to represent meanings that can be used in all theories of lexical semantics from different perspectives. Semantics studies require different semantic representations on the formalism for meaning of word \cite{erk4}. According to Kintsch, philosophers work with meaning of concepts instead of words, psychologists mostly study concept formation than vocabulary acquisition and linguistics work on meaning of word \cite{kintsch}. But, at this point precise representing and approximating the meaning of concept or specially a text such as sentences, passages or documents are still active problems in NLP and all other disciplines concerning with `meaning'. 

In this research, we specifically focus on meanings in scientific texts. We concern with how meaning can be extracted by analysing the large scientific corpus. Our fundamental assumption is that the meaning of a text can be extracted from the occurrence of its words in texts across the scientific categories. We hypothesize that there is a great connection between the meaning in a text and the vocabulary used in the text; however, we cannot say that each word has the same importance in all research disciplines. In fact, words have scientifically specific meaning in texts based on differences of use in subject categories and these meanings can be estimated from their occurrences in texts within categories. Difference in word meanings for categories correlates with the difference in distribution of words across categories. As they are scientific texts, we consider that occurrence of these words in texts of categories can be used in characterisation of word meaning for science.

Our approach to quantifying the meaning of a word differs from measuring its meaning on the basis of human sensations and feelings, as in psycholinguistic studies. Although measuring the meaning of a word in context by characterization through its dictionary meanings has many important implications in computational linguistics and psycholinguistic research, we do not focus here on dictionary meanings. Rather, we create a model for word representation that allows us to extract the meaning of a word through its importance in various scientific fields without distinguishing its dictionary meanings.
We approach the meaning of a word through the predictive power of a corpus analytical procedure under the assumption that the meaning of a word is determined by its use in scientific disciplines.
This actually matches the \textit{statistical semantics hypothesis} that `statistical patterns of human word usage can be utilised to figure out what people mean' \cite{turney}. We can also reword this as `statistical patterns of word usage in scientific fields can be used to figure out what a text means'.  

In these relations, the meaning of a word is defined as a vector of RIGs from the word to a category.  Given such information, meaning can be defined for each word and then for research text \cite{our}. A natural way to formalise this is to represent words as vectors and texts as sets of vectors in a specially constructed space. Differences in the distributions of vectors reflect differences in meaning of texts. This technique allowed us to represent each word by a distribution of numerical values over categories and meaning in text through a vector space model, that is, quantifying of meaning. 

In many semantic studies, the vector space is obtained by co-occurrence of words as discussed before. There are currently two broad VSMs based on co-occurrence: word-word and word-document where vectors are (normalised) frequency counts and dimensions are contexts (words or documents)\cite{magnus}. Vectors are called \textit{context vectors} in this case, and words are represented by the context vectors. In distributional hypothesis, these vectors are used to compute vector similarity. However, co-occurrence models are plagued with efficiency in real-word applications \cite{magnus}. There are two main problems in the usage of such approaches: first is the dimensionality in contexts vectors and the second is sparse data problem. In the first problem, the dimension of co-occurance matrix will tend to be extremely big for large data. In the second problem, as the vast majority of words occurs in a very small fraction of set of contexts \cite{zipf}, the majority of the entities of vectors will be zero. Therefore, the co-occurance matrix will  not give reliable results for large data and brief texts. Additional to these two problems, specifically, usage of co-occurrence is not appropriate for the representation of scientific texts due to multidisciplinary researches in the collection \cite{our}. Therefore, we introduce a new vector space to represent word meaning based on words' informational importance in the subject categories. 

We begin by creating a space to represent words meaning. The \textit{Meaning Space} is defined as a vector space, in which coordinates correspond to the subject categories. 
A word is represented by a vector of RIG  about the subject categories that the text belongs to, which can be obtained from observing the word in the text.  This approach allows us to identify the importance of the word for the corresponding category in terms of information gained when separating the corresponding category from its complement (like, for example, separating texts in category `algebra' from the text that do not belong to this category). 

To define RIGs, we consider the following  two attributes of text $d$  for a given word $ w_{j} $ and   a given category $c_{k}$:

\begin{description}
\item[$c_{k}(d)$] The text $d$ is in the category $c_{k}$: Attribute values are Yes ($c_{k}(d)=1$) or No ($c_{k}(d)=0$);
\item[$w_{j}(d)$] The word is in the text:  Attribute values are Yes ($w_{j}(d)=1$) or No ($w_{j}(d)=0$).
\end{description}

The corpus is considered as a probabilistic sample space (the space of equally probable elementary results, each of which is a random selection of text from the corpus).  RIG  measures the (normalized) information about the value of $c_{k}(d)$, which can be extracted from the value $w_{j}(d)$ (i.e. from observing the word  $ w_{j} $ in the text $d$) for a text $d$ from the corpus. 

As we have a number of word vectors, it is convenient to organise the vectors into a matrix. These vectors are used to construct \textit{Word-Category RIG Matrix}, in which rows correspond to words and columns correspond to categories. Each entry in the matrix corresponds to a pair \textit{(category,word)}. Its value for the pair $(c_{k},w_{j} )$ shows the RIG on the belonging of a text from the corpus to the category $c_{k}$ from observing the word $ w_{j} $ in this text. Word-Category RIG vectors estimate the meaning of words as their importance in the research fields. Thus, row vectors in the Word-Category RIG Matrix indicate words' scientific meanings.  

This approach computes a distributional representation (RIGs) for a word across all research subjects (RIGs in categories). Following to the distributional semantic hypothesis, if words have similar row vectors in the Word-Category RIG Matrix, they tend to have similar meanings. The hypothesis is that if texts have a similar distributions of word meanings -- similar clouds of word meanings vectors -- then they tend to have similar meanings.

We note that proposed hypothesis does not require an explicit distinguishing between homonymy and polysemy for words; it only requires linking the meaning of words to their importance in categories. With this approach, vocabulary meanings do not directly affect the representation of the meaning of the word. Rather, the meaning of a word is characterized through its measured information content in various scientific subject categories.

In this research, we present the first stage of `quantifying of meaning': construction of the Meaning Space and representing word meaning as a vector of RIGs for categories in this space. Such an understanding of meaning of words can help analyse the meaning of the texts. Having quantified meaning of words, one can represent all words in a corpus and then texts in the Meaning Space. Specifically, each text in the corpus is a cloud of RIG vectors and the text meaning can be later estimated and constructed by these distributions. Analysis of texts will be focused in the next stage of the research. Text analysis will be the next stage of the research. The earliest (preparatory) stage of the project was presented in \cite{LSC,our}.

The empirical analysis of this research is based on the Leicester Scientific Corpus which includes 1,673,350 texts \cite{LSCn} and Leicester Scientific Dictionary-Core (LScDC) of 103,998 words \cite{LScDCn}. The main hypothesis for construction of the Meaning Space is: meaning is the vector of information gains from the word to the categories assigned to the text. We used 252 categories of WoS.

We evaluated the Meaning Space and representation of word meaning in this space through top-ranked words in each category. We constructed the Word-Category RIG Matrix for the LSC \cite{wordcat}. The most informative words in each category are presented. It is shown that the proposed representation technique stands out topic-specific words in categories. We compared this approach with the representation technique where words are represented by vectors of their raw frequencies   in categories. Words are ranked by both frequencies and RIGs in categories. We demonstrated that frequencies are not much useful for identifying the most informative words in categories. We concluded that frequency is not much important in this sense.       
  
For each word in the LScDC, the sum and maximum of RIGs in categories are calculated and added at the end of the Word-Category RIG Matrix. Words can be ordered by their informativeness in scientific texts by these two criteria. The most informative $n$ words for scientific texts can be extracted by ordering/sorting words in column of the sum or maximum of RIGs. We compared these two ordering criteria by counting the number of matches in the top $n$ words, where $n$ ranges from 100 to 50,000. We concluded that the majority of the first 100 words do not match, with 28\% matched words. The intersection of words reaches to approximately 50\% for the top 1,000 words, and then 99\% for the top 50,000 words. 

Finally, we created a scientific thesaurus in which the most informative words were selected from the LScDC by their average RIGs in categories. The thesaurus was called \textit{Leicester Scientific Thesaurus (LScT)}. LScT contains the most informative 5,000 words in the corpus LSC. These words are considered as the most meaningful words in science. The full list of words in LScT is available online \cite{wordcat}.

\subsection{The Structure of This Paper\nopunct}\hspace*{\fill} \\\\

This paper is organised as follows. In Section \ref{rep},  the  Meaning Space is constructed and the representation of words by vectors in the Meaning Space is discussed. Given the representation of words by vectors of RIGs, we look at words ordered by their RIGs in each category. In Section \ref{exp}, we present the first findings of the new representation technique and the anomalies detected in the data by this model. To avoid a possible abnormal appearances of the words in the categories, we apply a further cleaning procedure of the LSC. The latest versions of the LSC, dictionaries Leicester Scientific Dictionary (LScD) and the LScDC are described \cite{LSCn,LScDCn,LScDn}. Finally, we construct the Word-Category RIG Matrix for the LSC \cite{wordcat} and discuss the experimental results in this section. In section \ref{thes}, we introduce the Leicester Scientific Thesaurus (LScT), in which there are 5,000 of the LScDC words selected by their average RIGs in categories. In Section \ref{conc}, the conclusion and outlook are summarised.

\section{Representation of Words by Vectors in the Meaning Space\nopunct}\label{rep}
In this section, we discuss the architecture of our approach to estimating the word meaning in a collection of documents. We assume that the dataset is a large corpus of natural language scientific texts and each text in the corpus belongs to at least one subject category. We hypothesize that words have scientifically specific meaning in categories and the meaning can be estimated by information gains from the word to the category. Before inquiring into the measurement of the meaning, we will mention how to represent each word as a vector of frequencies in categories. We then introduce a new approach to word meaning, in which each word is represented by a vector of RIGs in the \textit{Meaning Space}.     

\subsection{Representation of Words by   Vectors of Frequencies in Categories\nopunct}\hspace*{\fill} \\\\

In this section, we review how to represent a word in a vector space model by using  appearances this word in texts belonging to subject categories. A word representation method is defined in order to indicate term absence/presence in texts of categories. Each word is represented by a vector of frequencies in categories. That is, the number of presence of a word is calculated by how frequently this word is observed in texts belonging to the category. Each entry of the vector consists of the number of texts containing the word in the corresponding category. 

It is noteworthy that texts in a corpus do not necessarily belong to a single category as they are likely to correspond to multidisciplinary studies, specifically in a corpus of scientific researches. In other words, categories may not be mutually exclusive.

For every word $w_{j}$ from the dictionary ($j=1,...,N$) and every text $d_{i}$ from the corpus ($i=1,...,M$) the indicator $w_{j}(d_{i})$ is defined.  If the word $w_{j}$ occurs in the text $d_{i}$ (once or more), then $w_{j}(d_{i})=1$. Otherwise, $w_{j}(d_{i})=1$.

Let $D_{k}$ be a set of texts in the category $c_{k}$. The frequency of the word $w_{j}$ in the category  $c_{k}$ is 
$$w_{jk}=\sum_{d_{i}\in D_{k}} w_{j}(d_{i}) ,$$
This  $w_{jk}$ is the number of texts containing the word $w_{j}$ in the category $c_{k}$. 

The  vector of frequencies is defined for each word $w_{j}$ from the dictionary. Let us use  the notation  \textbf{$\overrightarrow{w_{j}}$} for it. Coordinates of this vectors are $w_{jk}$, where index $k=1,...,K$ corresponds to  the subject categories. 

Thus, each word  $w_{j}$ in the corpus is represented by a vector of frequencies $w_{jk}$ denoted by
$$\textbf{$\overrightarrow{w_{j}}$} = (w_{j1},w_{j2},...,w_{jK}),$$
where \textit{K} is the number of categories in the corpus. The collection of vectors, with all words and categories in the entire corpus, can be shown in a table. Each entry $w_{jk}$ of the Table \ref{table:tabrepr} corresponds to a word and a category.

\begin{table} [h]
\centering
\caption{The structure of  dictionary representation by frequencies $w_{jk}$}
  \label{table:tabrepr}
\begin{tabular}{|c|cccc|}
\hline
\backslashbox{\tabular{@{}l@{}}Word\endtabular}{Category}&$c_{1}$ &$c_{2}$ & $\cdots$ & $c_{K}$  \\\hline
$w_{1}$ & $w_{11}$  & $w_{12}$  &$\cdots$ &$w_{1K}$    \\
$w_{2}$ &  $w_{21}$  & $w_{22}$  &$\cdots$  &$w_{2K}$    \\
&    &   &  &    \\
$\vdots$ & $\vdots$ &  $\vdots$ &  & $\vdots$  \\
&    &   &  &    \\
$w_{N}$ &  $w_{N1}$  & $w_{N2}$  &$\cdots$  &$w_{NK}$   \\\hline
\end{tabular}
\end{table}

The number of documents in the category $c_k$ is $ \vert D_{k} \vert$. Importantly, 
$$ \vert D_{k} \vert \leq \sum_{j} w_{jk} $$
as each text usually has  more than one word, and several different words can belong to the same text. To simplify the notation for further calculations, we now define the set of texts containing the word $w_{j}$ as $ D^{j} $. We note that 
 $$ \vert D^{j} \vert \leq \sum_{k} w_{jk} $$ and equality holds in the case when categories are mutually exclusive.

The number of texts in the categories varies widely, so $w_{jk}$ is expected to increase as the number of texts in a category increases. This does not necessarily mean that a word rarely appearing in a category is less important for this category than for other categories in which the word appears more frequently (see the definition of information gain in the next section). Therefore, direct usage of frequencies may result inappropriate findings in quantification of words' meanings.

Given the collection of vectors, various schemes for normalisation can be performed to adjust the vectors $\textbf{$\overrightarrow{w_{j}}$}$ to a common scale. The simplest and the most popular approach for normalisation is transformation to a vector where the sum of the elements is 1, that is normalisation to unite $l_1$ norm. For the mutually exclusive categories, this normalisation is related to the law of total probability.  The objective of this normalisation scheme is to make vectors comparable by rescaling them to the same length in the $l_1$ norm. For a given vector $\textbf{$\overrightarrow{w_{j}}$}$, the normalisation can be performed as 
$$  P_{jk}=\dfrac{w_{jk}}{\sum_{i} w_{ji}}$$
where $ \sum_{k} P_{jk}=1$. It should be stressed that when categories are not exclusive, $ \sum_{k} w_{jk} $ is not the total number of texts containing the word $w_{j}$. In other words, texts containing the word could be counted more than once in the sum. 

In similar way, the column vectors can be normalised as 
$$  Q_{jk}=\dfrac{w_{jk}}{\sum_{i} w_{ik}}.$$
However, this representation does not indicate the proportion of exact number of texts in the category.

A reasonable normalisation can also be obtained in two-steps:
\begin{enumerate}
\item Normalize each frequency:
$$w_{jk}\mapsto \frac{w_{jk}}{\vert D_{k} \vert};$$
\item Normalize the matrix to the unite sum in rows.
\end{enumerate}
As a result, $w_{jk}$ will be transformed into
$$\dfrac{w_{jk}}{\vert D_{k} \vert \sum_{i} \dfrac{w_{ji}}{\vert D_{i} \vert}}\, .$$

In calculation of RIGs below, the estimation of probabilities are used based on the table  of frequencies. For ranking of words in categories, the raw frequencies were also used and compared to RIG-based ranking.   

\subsection{Word Meaning as a Vector of RIGs Extracted for Categories\nopunct}\hspace*{\fill} \\\\

Having a collection of frequency vectors, it is easy to calculate the vectors of information gains (from observing the word in the text to categories which the text belongs to). These vectors will quantify the meaning the words. The hypothesis here is that the informational content of a word about each category can be measured by comparing the appearance of a word in texts of a given category and its appearance in texts not related to this category (i.e, how the presence/absence of the word in texts can help to separate the category from its set-theoretical complement). 

A general concept for computing information is the “Shannon entropy” introduced by Shannon  \cite{shannon}. \textit{Information Gain (IG)} is a  common feature selection criterion in machine learning used, in particular, for evaluation of word goodness \cite{largeron,yangpederson}. The information gain is the measure of the information extracted about one random variable if the value of another random variable is known. It is closely related to the \textit{mutual information}, that measures the statistical dependence between two random variables. The larger value of the gain means the stronger relationship between the variables. The information gain of random variable $A$ with values  (or states) $a_1,\ldots,a_n$  from the random variable $B$ with values (or states) $b_1,\ldots,b_m$ is defined as:

\begin{equation}\label{eq:IGe}
\begin{split}
IG(A, B)= &-\sum_{i=1}^n  P(A=a_i)\log_2 P(A=a_i)\\
&+\sum_{j=1}^m P(B=b_j)\sum_{i=1}^n  P(A=a_i \vert B=b_j)\log_2 P(A=a_i \vert B=b_j)
\end{split}
\end{equation}
where $P(A=a_i)$ is probability of observing   the value $a_i$ of the random variable $A$, $P(B=b_j)$ is probability of observing  the value $b_j$ of the random variable $B$, $ P(A=a_i \vert B=b_j)$ is conditional probability of observing  the value $a_i$ of the random variable $A$ given the value $b_j$ of the random variable $B$. $ IG(A\vert B)$ measures the number of bits of information obtained for prediction of a value of the variable $A$ by knowing the value of the variable $B$. 

In the concept of text categorisation, the information gain measures how important a given word is for category prediction. A larger gain indicates that the  probability to find the  word in the texts {\em inside} the category differs considerably from the probability to find it in the text {\em outside} this category. If the categories are mutually exclusive then we can consider them as values of a categoric feature $C$ of the text with values $c_i$ and define the  information gain $IG(C,w)$ from observing a word $w$ in the text about the value of $C$ \cite{yangpederson} by the textbook formula:
\begin{equation}\label{eq:IG}
\begin{split}
IG(C,w)= - \sum_{i=1}^{m} P(c_{i})\log_2 P(c_{i})+P(w)\sum_{i=1}^{m} P(c_{i}|w)\log_2 P(c_{i}|w) \\ 
+P(\overline{w})\sum_{i=1}^{m} P(c_{i}|\overline{w})\log_2 P(c_{i}|\overline{w})
\end{split}
\end{equation}
where \{$c_{i}$\} is the set of classes in the target space, $P(c_{i})$ is the probability of observing  the $i^{th}$ class, $P(w)$ is the probability that the term $w$ appears, $P(\overline{w})$ is the probability that $w$ does not appear, $ P(c_{i}|w) $  is the conditional probability of observing the $i^{th}$ class given that the term $w$ appears, and $P(c_{i}|\overline{w})$ is the conditional probability of observing the $i^{th}$ class given that the term $w$ does not appear. 

$IG(C,w)$ measures the number of bits of information obtained for prediction of classes $c_{i}$ by knowing the presence and absence of a term $w$ in documents of classes. 

The quantity $IG(C,w)$ measures the amount of information provided by a word when splitting the documents into classes but only in the case of mutually exclusive classes, that is, each text is assigned to a single class only. On the contrary, the scientific texts belong very often to several categories. The research subject categories are not mutually exclusive and this approach cannot be used directly.

 Unlike this approach, we start from measuring how a word is informative for a category in terms of its ability to separate the corresponding category from its set-theoretical complement. We hypothesize that the topic-specific words in categories have larger information gain than other words and such words are expected to have less gain in most other categories. Therefore, we approach to this problem by defining for each subject category $c_k$ a random Boolean  variable, or a classification  with two states: the text belongs to the category $c_k$ and the text does not belong to the category $c_k$ (this class is denoted as $\overline{c_k}$). The frequencies  of words in classes of texts $c_k$ and $\overline{c_k}$ is demonstrated in Table~\ref{table:tablepair}.

\begin{table} [ht]
\centering
\caption{Representation of the word by a pair of frequencies: the number of texts containing the word $w_{j}$ that belong and does not belong to the category $c_{k}$}
  \label{table:tablepair}
\begin{tabular}{|c|c|c|}
\hline
\backslashbox{\tabular{@{}l@{}}Word\endtabular}{Category}&$c_{k}$ &$\overline{c_{k}}$  \\\hline
$w_{1}$ & $w_{1k}$  & $|D^{1}|-w_{1k}$    \\
$w_{2}$ & $w_{2k}$  & $|D^{2}|-w_{2k}$    \\
 &    &       \\
$\vdots$ & $\vdots$ &  $\vdots$  \\
 &    &       \\
$w_{N}$ &  $w_{Nk}$  & $|D^{N}|-w_{Nk}$   \\ \hline
\end{tabular}
\end{table}

Since words are obviously not mutually exclusive (one text usually contains several different words) we cannot consider the occurrence of different words as values of a random variable to use \eqref{eq:IGe} directly. To evaluate the information gain of the category $c_{k}$ from the word $w_{j}$ it is necessary to introduce for each word $w_{j}$ a random Boolean variable with two states: $w_{j}$ denotes the presence of the word in texts of the category $c_{k}$ and $\overline{w_{j}}$ denotes the absence of the word $w_{j}$ in texts of the category $c_{k}$. Contingency $2\times 2$ table to calculate information gain of the category $c_{k}$ from the word $w_{j}$ is presented in Table \ref{table:cont}. It  used the raw frequencies $w_{jk}$ introduced in previous Subsection.

\begin{table}
\caption{Contingency table for the category $c_{k}$ and the word $w_{j}$}
\label{table:cont}
\begin{tabular}{|c|c|c|c|}
\hline
\backslashbox{\tabular{@{}l@{}}Word\endtabular}{Category}&$c_{k}$ &$\overline{c_{k}}$& Total  \\\hline
$w_{j}$ & $w_{jk}$  & $|D^{j}|-w_{jk}$&$|D^{j}|$  \\\hline
$\overline{w_{j}}$ & $|D_{k}|-w_{jk}$  & $M-|D_{k}|-(|D^{j}|-w_{jk})$ &$ M-|D^{j}|$ \\\hline
 Total &$|D_{k}|$    &  $M-|D_{k}|$ & $M$  \\ \hline
\end{tabular}
\end{table}

Table \ref{table:cont} can be used to calculate two information gains: the word $w_{j}$ from the category $c_{k}$ and the category $c_{k}$ from the word $w_{j}$. Both information gains have a meaning for different problems. The goal of this research is to evaluate informativeness of words for category identification and use this informativeness for word ranking and text representations. Therefore, we will consider information gain of the category $c_{k}$ from the word $w_{j}$: $IG(c_{k},w_{j} )$. This information gain evaluates the number of bits extracted from presence/absence of the word $w_{j}$ in the text for prediction of belonging of this text to the category $c_{k}$. One may expect that if a word is a very topic-specific for a category, it appears in texts belonging to this category more frequently than in texts which do not belong to this category; and the major part of texts belonging to this category contains the word.  

For each category, $c_{k}$, a function is defined on texts that takes the value 1, if the text belongs to the category $c_{k}$, and 0 otherwise. For each word, $w_{j}$, a function is defined on texts that takes the value 1 if the word $w_{j}$ belongs to the text, and 0 otherwise. We use for these functions the same notations $c_{k}$and $w_{j}$. Consider the corpus as a probabilistic sample space (the space of equally probable elementary outcomes). For the Boolean random variables, $c_{k}$ and $w_{j}$, the joint probability distribution is defined according to  Table~\ref{table:cont}, the entropy and information gains can be defined as follows.

The information gain about the category $c_{k}$ from the word $w_{j}$, $IG(c_{k},w_{j} )$,  is the amount of information on belonging of a text from the corpus to the category $c_{k}$ from observing the word $w_{j}$ in the text. It can be calculated as \cite{shannon}:

\begin{equation}\label{eq:IGs}
\begin{split}
IG(c_{k},w_{j} )= H(c_{k})-H(c_{k}|w_{j}),
\end{split}
\end{equation}
where $H(c_{k})$ is the Shannon entropy of $c_{k}$ and $H(c_{k}|w_{j})$ is the conditional entropy of $c_{k}$ given the observing   the word $w_{j}$. Entropies $H(c_{k})$  and $H(c_{k}|w_{j})$ are 

\begin{equation}\label{eq:ent1}
\begin{split}
H(c_{k})=&-P(c_{k})\log_{2}P(c_{k})-P(\overline{c_k})\log_{2}P(\overline{c_k}),
\end{split}
\end{equation}
where $ P(c_{k} )$ is the probability that the text belongs to the category $c_{k}$, $P(\overline{c_{k}})$ is the probability that the text does not belong to the category $c_{k}$ and

\begin{equation}\label{eq:ent2}
\begin{split}
H(c_{k}|w_{j})=&P(w_j)\big(-P(c_{k}\vert w_j)\log_{2}P(c_{k}\vert w_j)-P(\overline{c_k}\vert w_j)\log_{2}P(\overline{c_k}\vert w_j) \big)\\
&+P(\overline{w_j})\big(-P(c_{k}\vert \overline{w_j})\log_{2}P(c_{k}\vert \overline{w_j})-P(\overline{c_k}\vert \overline{w_j})\log_{2}P(\overline{c_k}\vert \overline{w_j}) \big),
\end{split}
\end{equation}
where 
\begin{itemize}
\item$P(w_{j} )$ is the probability that the word $ w_{j} $ appears in a text from the corpus; 
\item$P(\overline{w_{j}})$ is the probability that the word $ w_{j} $ does not appear in a text from the corpus; 
\item$P(c_{k}|w_{j} )$ is the probability that a text  belongs to the category $c_{k}$ under the condition that it contains the word $ w_{j} $;
\item$P(\overline{c_{k}}|w_{j})$ is the probability that a text does not belong to the category $c_{k}$ under the  condition that it contains the word $ w_{j} $; 
\item$P(c_{k}| \overline{w_{j}})$ is the probability that a text belongs to the category  $c_{k}$ under  the condition that it does not contain the word $ w_{j} $; 
\item$P(\overline{c_{k}}| \overline{w_{j}})$ is the probability that a text  does not belong to the category $c_{k}$ under the  condition that it does not contain the word $ w_{j} $.
\end{itemize} 

All the required probabilities, entropies and relative entropies are evaluated using 
 the contingency table \ref{table:cont} as: 

\begin{equation}\label{eq:ent3}
\begin{split}
H(c_{k})=&-\frac{\vert D_k\vert}{M}\log_{2}\frac{\vert D_k\vert}{M}-\frac{M-\vert D_k\vert}{M}\log_{2}\frac{M-\vert D_k\vert}{M},
\end{split}
\end{equation}
  
and

\begin{equation}\label{eq:ent4}
\begin{split}
H(c_{k}|w_{j})= \dfrac{|D^{j}|}{M}\Big(-\dfrac{w_{jk}}{|D^{j}|}  \log_{2}\dfrac{w_{jk}}{|D^{j}|} -\dfrac{|D^{j}|-w_{jk}}{|D^{j}|}\log_{2}\dfrac{|D^{j}|-w_{jk}}{|D^{j}|}\Big) \\
+\dfrac{M-|D^{j}|}{M}\Big(-\dfrac{|D_{k} |-w_{jk}}{M-|D^{j}|}\log_{2} \dfrac{|D_{k} |-w_{jk}}{M-|D^{j}|}\\
-\dfrac{M-|D_{k}|-(|D^{j}|-w_{jk})}{M-|D^{j}|}\log_{2}  \dfrac{M-|D_{k}|-(|D^{j}|-w_{jk})}{M-|D^{j}|} \Big).
\end{split}
\end{equation}

information about whether an element belongs to a set.

High value of the  informational gain   $ IG (c_ {k}, w_ {j}) $ (\ref{eq:IGs}) does not mean, in general, that the large proportion of information about whether a text belongs to the category $ c_ {k} $ can be extracted from observing the word $ w_ {j} $ in this text. This proportion depends on the value of the entropy $H(c_{k})$ (\ref{eq:ent3}).  The \textit{Relative Information Gain} (RIG) measures this proportion directly. It provides a normalised measure of the Information Gain with regard to the entropy of $c_{k}$. RIG is defined as

\begin{equation}\label{eq:RIG}
\begin{split}
RIG(c_{k} \vert w_{j} )=\dfrac{IG(c_{k},w_{j} )}{H(c_{k})}.
\end{split}
\end{equation}

The value of $RIG(c_{k}\vert w_{j} )$ will be 0 when $H(c_{k})=H(c_{k}|w_{j})$ and 1 when $H(c_{k}|w_{j})=0$. In the first case, the presence/absence of the given word $w_{j}$ does not contain information for the category $c_{k}$. So, this word is uninformative. In the second case, using the word in the category provides exactly  $H(c_{k})$ bits of information. That is, presence or absence of a word resolves exactly the question of belonging the text to the category. $RIG(c_{k}\vert w_{j} )$ can be equal to 1 in two cases:

\begin{itemize}
  \item All texts with the word $w_j$ belong to the category $c_k$ and all texts without the word $w_j$ do not belong to the category $c_k$;
  \item All texts with the word $w_j$ do not belong to the category $c_k$ and all texts without the word $w_j$ belong to the category $c_k$;
\end{itemize}

We expect higher  $RIG(c_{k}\vert w_{j})$ for the topic-specific words of the category $c_k$. 

For simplicity, we denote $RIG(c_{k}\vert w_{j} )$ by $ RIG_{jk} $. Given the word $w_{j}$, $ RIG_{jk} $ is used to form vector $\overrightarrow{RIG_{j}}$, where each component of the vector corresponds to a category. Therefore, each word is represented by a vector of RIGs. It is obvious that the dimension of vector for each word is the number of categories $K$ (for the WoS subject categories $K=252$). For the word $w_{j}$, this vector is
$$\textbf{$\overrightarrow{RIG_{j}}$} = (RIG_{j1},RIG_{j2},...,RIG_{jK}).$$

The set of  vectors $\overrightarrow{RIG_{j}}$ can be used to form the \textit{Word-Category RIG Matrix}, in which each column corresponds to a category  $ c_{k} $ and each row corresponds to a word $w_j$. Each component $ RIG_{jk} $ corresponds to a pair $(c_{k},w_j)$ and its value is the RIG from the word $w_{j}$ to the category  $ c_{k} $. The structure of the Word-Category RIG Matrix is demonstrated in Table \ref{table:RIGs}. 

\begin{table}[ht]
\centering
\caption{The structure of the Word-Category RIG Matrix}\label{table:RIGs}
\begin{tabular}{|c|cccc|}
\hline
\backslashbox{\tabular{@{}l@{}}Word\endtabular}{Category}&$c_{1}$ &$c_{2}$ & $\cdots$ & $c_{k}$  \\\hline
$w_1$ & $ RIG_{11} $  & $ RIG_{12} $  &$\cdots$  &$ RIG_{1K} $    \\
$w_2$ & $ RIG_{21} $  &$ RIG_{22} $ &$\cdots$  &$ RIG_{2K}$  \\
 &    &   &  &   \\
$\vdots$ & $\vdots$ &  $\vdots$ &  & $\vdots$  \\
 &    &   &  &   \\
$w_N$ &  $ RIG_{N1} $ & $ RIG_{N2} $  &$\cdots$  &$ RIG_{NK} $ \\\hline
\end{tabular}
\end{table}

In the Word-Category RIG Matrix, a row vector represents the corresponding word as a vector of RIGs for categories.  We defined the \textit{Meaning Space} as the vector space of such vectors $\overrightarrow{RIG_{j}}$. The dimension of this space is the number of categories and each coordinate is the RIG from a word to this category.  

Note that in the Word-Category RIG Matrix, a column vector represents RIGs of all words in an individual category. If we choose an arbitrary category, the words can be ordered by their RIGs from the most informative word to the least informative one. We expect that the topic-specific words will appear at the top of the list. 

The words can be ordered by their informativeness in the whole corpus of scientific texts as well as they are ordered in each category.  A norm or a more general proximity measure in the Meaning Space is needed to compare the meaningfulness of words across all categories.  Two criteria were tested for measuring informativeness of words in the corpus of scientific texts: the sum ($l_1$ norm)  and the maximum ($l_{\infty}$ norm) of RIGs in categories. For a given word $w_{j}$, the sum $ S_{j} $ and the maximum $M_{j}$ of RIGs are calculated from the Word-Category RIG Matrix as:
\begin{equation}\label{eq:sum}
S_{j}=\sum_{k=1}^K RIG_{jk}
\end{equation}
and
\begin{equation}\label{eq:max} 
 M_{j}=\max_{k=1,\ldots,K} (RIG_{jk}).
\end{equation}

The sum $S_{j}$ is a measure of the average informativeness of a word (this word has the informativeness $S_{j}/K$ on average), whereas the maximum $M_{j}$ is a measure of the maximal informativeness of the word across the categories (this word is not more informative than $M_{j}$ in any category).

Now, the words in the dictionary can be ordered by their $S_{j}$ or $M_{j}$. For each of these ordered lists of words, the most informative (meaningful) $n$ words for scientific texts can be selected based on one of these two criteria. The higher the value of the criterion ($S_{j}$ or $M_{j}$), the more informative the word is.

\section{Experimental Results}\label{exp}
This section describes the experimental details and the analysis done to show the performance of the vector representation method described in Section \ref{rep}. The dataset used in this study is the Leicester Scientific Corpus (LSC) \cite{LSCn}. The LSC contains a collection of abstracts of research articles and proceeding papers with metadata such as authors, title, categories, research areas and times cited. Each record (text) in the dataset is assigned to at least one of the WoS categories. The Leicester Scientific Dictionary-Core (LScDC) is the collection of unique words appearing in 10 or more documents in the LSC \cite{LScDCn}.

For each word $w_{j}$ and category $c_{k}$, $RIG_{jk}$ is calculated and the Word-Category RIG Matrix for the LSC was formed as described in Section \ref{rep}. In each category, a list of words where words are sorted in descending order by their RIGs can be created. The higher the relative information a word gained in a category, the more important the word is in terms of being topic-specific for the category. Therefore, one could look at the top $n$ words in categories in order to get a good grasp of the representation method. The visualisation of the top words in each category is carried out with the word clouds. Having calculated the frequencies of words in the categories (Table \ref{table:tabrepr}), we compare the purposed method with the commonly-used approach based on raw frequency.

\subsection{Discovering Anomalies in the Data by Information Gain-Based Representation Technique \nopunct}\hspace*{\fill} \\\\

\begin{figure}[h]
\centering
 \includegraphics[width=1\linewidth]{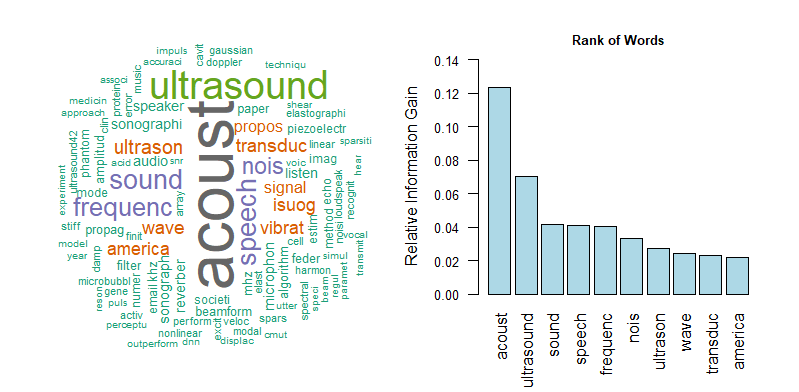}
   \caption{The most informative 100 words in the category `Acoustics'. The font size and colour of words indicate different RIGs of the words. The histogram shows RIGs for the top 10 most informative words in the category.}
  \label{fig:acoustic_old} 
\end{figure}

At first, the procedure of word representation was applied to the LSC version 1 \cite{LSC} with the dictionary \cite{LScDC}. To visualise top words in each categories in a convenient way, we looked at word clouds. The font size of each word in a word cloud is proportional to its RIG in the category. Intuitively, the more informative the word is, the bigger size the word appears in word clouds. For example, from Figure \ref{fig:acoustic_old}, it can be seen that the most informative 6 words for the category `Acoustics' are `acoust', `ultrasound', `speech', `nois', `sound' and `frequenc'. The majority of papers in Acoustics is expected to include these words which are absent or at least less frequent in many other categories. These words are inferred to be informative for the category `Acoustics'. 
 
\begin{figure}[tb]
\centering
 \includegraphics[width=1\linewidth]{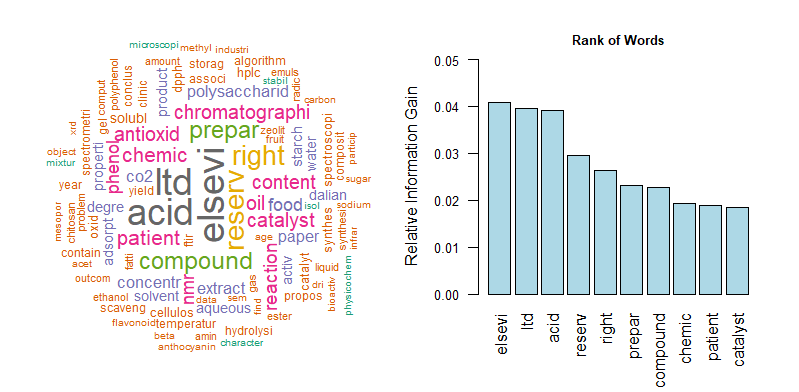}
   \caption{The most informative 100 words in the category `Chemistry, Applied' before additional cleaning. The font size and colour of words indicate different RIGs of words. The histogram shows RIGs for the top 10 most informative words in the category.}
  \label{fig:Chemistry_app_old}
\end{figure}

However, this method detected anomalies in some categories. Anomalies here refers to words that do not conform to the expected set of words to be appearing in a subject category. Such words can appear in any category frequently regardless of being a topic-specific word. These words are likely to be potential anomalies generated by inappropriate joints of words, phrases or sentences to the texts of abstracts. As shown in Figure \ref{fig:Chemistry_app_old}, for the category `Chemistry, Applied', words `elsevi', `ltd', `acid', `reserv' and `right' stand out in the word cloud. We see that trends in majority of words in the word cloud agree with each other as being related to the subject. However, `elsevi', `ltd', `reserv' and `right' seem like more prominent and unusual (non-specific) for Chemistry. In fact, the experiments were preliminary, but we discovered alarms indicating anomalies by our representation technique. 

To understand why these words arose and how they can be avoided, we checked the abstracts containing such words. Our review showed that these words appeared in copyright notices such as “Published by Elsevier ltd.” or `All rights reserved', and they were added at the footer of abstracts. In order to have a comprehensive understanding of their appearance as being informative for only some categories, for instance in Chemistry, we compared distributions of `elsevier', `right' and `reserve'  in categories. For each word, categories are ordered by the number of documents containing the word, and the first 20 categories are presented in Figure \ref{fig:wordstogether}. When we consider the list of categories ordered by the number of documents in the entire corpus, we conclude that not all categories in the list of top categories appear in the charts. This is because usage of copyright notices is much more noticeable in some categories such as Chemistry. For instance, the rank of the category `Engineering, Electrical \& Electronic' is 1 in the corpus; however, one can see that this category has rank 15 for the word 'Elsevier'. 

\begin{figure}[tb]
\centering
 \includegraphics[width=1\linewidth]{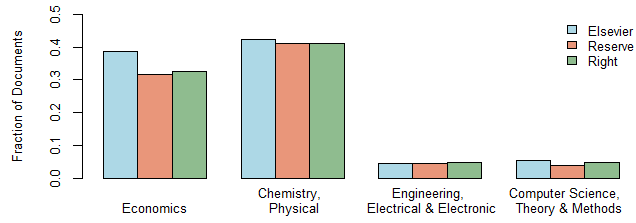}
   \caption{Fractions of texts containing the words `Elsevier', `Right' and `Reserve' in four   categories before additional cleaning.}
  \label{fig:threewordsall}
\end{figure}

\begin{figure}[h]
\centering
\includegraphics[width=1\linewidth]{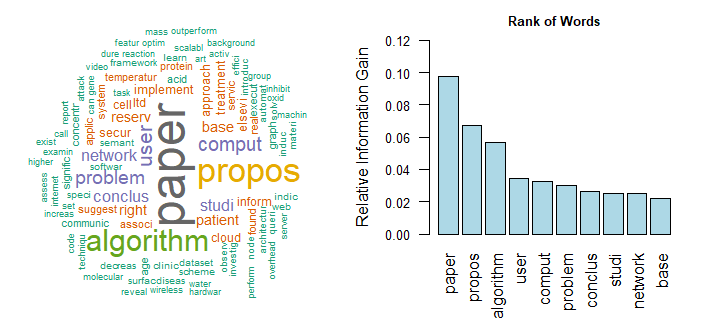}
  \caption{The most informative 100 words in the category `Computer Science, Theory \& Methods'. The font size and colour of words indicate different RIGs of words. The histogram shows RIGs for the top 10 most informative words in the category. (Before additional cleaning. However, the additional cleaning does not change  noticeably the  diagrams for this category.}
  \label{fig:comptheoryold}
\end{figure}

To show that not all categories have the same/similar distribution of use of copyright notices, we presented Figure \ref{fig:threewordsall} where fractions of documents containing words `elsevier', `right' and `reserve' in four of categories are demonstrated. We can see from the figure that  about 40\% of texts in `Chemistry, Physical' (rank is 4 in the corpus) contain  these three words, while only   small fragments of `Engineering, Electrical \& Electronic' and `Computer Science, Theory \& Methods' collections (rank is 6 in the corpus) include these words. `Economics' (rank is 34 in the corpus) papers include these words in considerable portion. As expected, these words did not appear as the most important words (10 words) for `Computer Science, Theory \& Methods' (see Figure  \ref{fig:comptheoryold}). 

\begin{landscape}\centering
\vspace*{\fill}
\begin{figure}[htpb]
  \centering
  \includegraphics[height=0.6\textheight, width=1.5\textwidth]{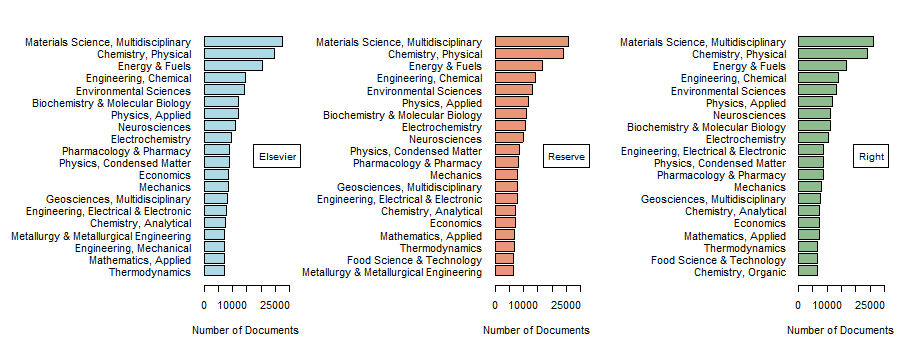}
  \caption{Top 20 categories that words `Elsevier', `Reserve' and `Right' appear in}
  \label{fig:wordstogether}
\end{figure}
\vfill
\end{landscape}

\subsection{The Data \nopunct}\hspace*{\fill} 

This subsection provides the description of procedure of additional cleaning and correction for the LSC and LScDC.

\subsubsection{\textbf{Further Cleaning of the Leicester Scientific Corpus (LSC)} \nopunct}\hspace*{\fill} \\\\

Many conferences and journals put copyright notices, permission policies or conference names below abstract of papers. Such footers were added to abstracts in many records in Web of Science database and so in the LSC during processing and storage of the original data (see Table \ref{table:copyrightex}). 

It is really a huge and practically impossible task to find out with the help of human inspection which notifications were added in the texts of 1,673,824 abstracts in \cite{LSC}. Once a sample of abstracts containing publishing houses names was browsed, we found that there are much more scenarios to consider. Some examples of these scenarios are presented in Table \ref{table:scenarios}. As such expressions are more frequent in some categories than in others, a cleaning procedure is needed to avoid possible abnormal appearances of words in categories. A quick look at the scenarios is sufficient to conclude that clearing such sentences or phrases cannot be fully automated. Human intervention is needed to identify and list them to avoid deleting useful information from the data.

 % Table of sample copyright mistake
	\begin{table}[b]
		\centering
		\caption{An example of abstract with a copyright notice}
		\renewcommand\arraystretch{1.3}
  			\begin{tabular}{|m{2cm}|m{10cm}|}
	  	 		\hline
	  	 		\rr \textbf{Title} & \rr Neonicotinoid concentrations in arable soils after seed treatment applications in preceding years  \tn \hline
	  	\rr \textbf{Authors} &\rr Jones, A; Harrington, P; Turnbull, G  \tn \hline
	 
	  	\rr \textbf{Abstract} & \rr Concentrations of the neonicotinoid insecticides clothianidin, thiamethoxam and imidacloprid were determined in arable soils from a variety of locations in England. \textbf{...[Truncated]}. As clothianidin and thiamethoxam have largely superseded imidacloprid in the United Kingdom, neonicotinoid levels were lower than suggested by predictions based on imidacloprid alone. (c) 2014 Crown copyright. Pest Management Science (c) 2014 Society of Chemical Industry \tn \hline
			\end{tabular}

\label{table:copyrightex}
   \end{table}

 % Table of different scenarios
	\begin{table}[tb]
		\centering
		\caption{Some examples of notices attached to the abstract}
		\renewcommand\arraystretch{1.3}
  			\begin{tabular}{ | m{12cm} |}
	  	 		\hline
	  	 		\rc \textbf{Copyright Notice, Name of Conference, Journal or Publishing House} 					 \tn \hline
	  	 		\rr (c) 2014 Elsevier ltd. All rights reserved.													 \tn \hline
	  			\rr (c) 2014 Published by Elsevier B.V.			 												 \tn \hline
	 		  	\rr (c) The Authors. Published by Elsevier Inc. All rights reserved.							 \tn \hline
			    \rr Crown Copyright (c) 2014 Published by Elsevier B.V. All rights reserved.	 				 \tn \hline							
			    \rr (c) 2014 The British Infection Association. Published by Elsevier Ltd. All rights reserved.	 \tn \hline	
			    \rr (c) Wolters Kluwer Health | Lippincott Williams \& Wilkins								 	 \tn \hline	
			    \rr (c) 2014 Wiley Periodicals Inc.																 \tn \hline	
			    \rr (c) Springer-verlag Berlin Heidelberg 2014													 \tn \hline	
			    \rr (c) The Authors. Published by SPIE under a Creative Commons Attribution 3.0 Unported License.\tn \hline
			    \rr (c) RSNA.																					 \tn \hline
			    \rr 2014 American Cancer Society.																 \tn \hline
			    \rr Pediatr Blood Cancer.																 		 \tn \hline
			    \rr J. med. virol																  				 \tn \hline
			\end{tabular}

\label{table:scenarios}
   \end{table}   

Individual notices with different appearances were identified by sampling of abstracts based on keyword search. A keyword search refers to browsing words, phrases or sentences to list different appearances of them in order to delete all identified appearances from abstracts. The position of notices was also taken into account since they appeared either at the beginning (by mistake) or at the end of the text. We used several specially developed procedures successively to clean them. For instance, when removing notices in the form of `(c) Published by Elsevier', we first checked the appearance of `Crown Copyright (c) Published by Elsevier'. It can also appear in the form of `Published by Elsevier', thus we consider all cases based on empirical study. During cleaning, we removed copyright notices, names of conferences, names of journals, authors' rights, licenses and permission policies identified. To give an insight, Table \ref{table:scenarios2} presents the number of document containing some notices before cleaning. These notices were completely removed after cleaning. We note that names of publishing houses could appear inside the text, in this case we did not remove them. More examples of notices that were removed from abstracts can be found in Appendix \ref{table:addsch}. 

\begin{figure}[tb]
\centering
 \includegraphics[width=1\linewidth]{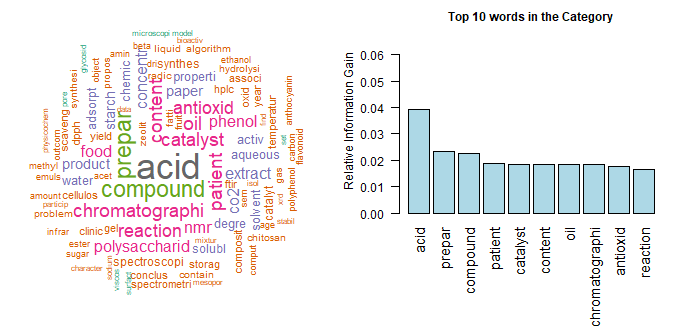}
   \caption{The most informative 100 words in the category `Chemistry, Applied' after cleaning. The font size and colour of words indicate different RIGs of words. The histogram shows RIGs for the top 10 most informative words in the category.}
  \label{fig:Chemistry_app_new}
\end{figure}

To display the initial result of the cleaning, we present the word cloud and histogram of RIGs for the category `Chemistry, Applied' in Figure \ref{fig:Chemistry_app_new}. One can see that words `elsevi', `ltd', `acid', `reserv' and `right' do not appear in the list of top words as was in the word cloud before cleaning (see Figure \ref{fig:Chemistry_app_old}). Instead, the cloud gives greater prominence to words that are related to specific topics and likely to be more informative for the category. The word `acid' has been preserved in the list of the most informative words.

 % Table of before-afetr cleaning
	\begin{table}[h]
		\centering
		\caption{The number of abstracts containing some attached notices before cleaning (These notices were completely removed after cleaning)}
		\renewcommand\arraystretch{1.3}
  			\begin{tabular}{ | m{7cm} | R{4cm} | }
	  	 		\hline
	  	 \rc \textbf{Notice}				&  \rc  \textbf{Number of Notices Before Cleaning }     			 \tn \hline
	  	 \rr Elsevier ltd. All rights reserved											 	&	101,994       						 \tn \hline
	     \rr (c) The Authors			  													& 	561									 \tn \hline				
		 \rr All rights reserved	 														&	283,041								 \tn \hline	
		 \rr (c) Springer-verlag Berlin Heidelberg 2014										&	20						 			 \tn \hline	 
			\end{tabular}
\label{table:scenarios2}
   \end{table}

\subsubsection{\textbf{The Latest Version of LSC and Dictionaries} \nopunct}\hspace*{\fill} \\\\
After detecting and cleaning copyright notices, permission policies and conference names from abstracts, a new version of the LSC was created and made accessible in \cite{LSCn}. The cleaning procedure described in previous section leaded to some abstracts having less than our minimum length criteria (30 words). Such abstracts were not contained in the new version; therefore, the remaining 1,673,350 texts were used in this study (474 texts were removed). As was the case for the LSC before cleaning, the latest version of the LSC involved text of abstracts, list of authors, title, list of research areas, list of categories and times cited. 

\begin{figure}[h]
\centering
 \includegraphics[width=0.55\linewidth]{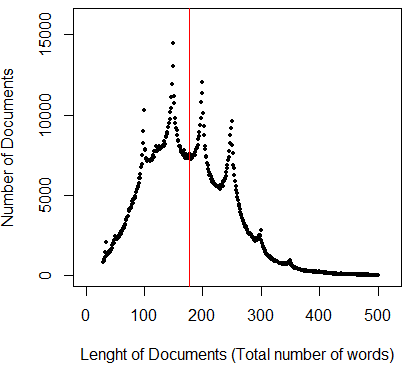}
   \caption{The number of abstracts with specified length against lengths of abstracts in the latest version of the LSC. The minimum length is 30 and the maximum length is 500 with an average of 176 words.}
  \label{fig:lenght}
\end{figure}

It is noteworthy that, in both versions of the LSC, the number of subject categories is 252. All categories and the number of documents assigned to the corresponding category are presented in Table \ref{table:categories}. Same information for research areas was provided in Table \ref{table:ra}. The distribution of length of abstracts is displayed in Figure \ref{fig:lenght}. There is no noticeable difference between distributions for two versions and the average length of texts is 176 words. 
	
The latest version of the Leicester Scientific Dictionary (LScD) was developed by extracting words from the new version of the LSC \cite{LScDn}. The procedure applied to process the LSC in creation the LScD was the same as described in \cite{our}. The new version of the LScD contains 972,060 unique words with the number of texts that a word appears in. A new version of the core list, LScDC, was created from  the LScD by removing words appearing in no more than 10 texts of the LSC \cite{LScDCn}. All steps applied were the same as for the previous version of the LScDC and can be found in \cite{our}.

Based on the decision to clean copyright notices, we expect that words such as `Elsevier', `Reserved', `Ltd', `Right' and `Springer' will not appear frequently in the LSC as they did before. In fact, the number of appearance of these words decreased after cleaning (see Table \ref{table:beforeafter}). The results indicated that some words, for instance `Right' and `Reserve', are still relatively frequent in the corpus. This is because these words are specific for some categories. To give an insight, we compared top categories for three words `Elsevi', `Reserv' and `Right' (see Figure \ref{fig:wordstogethernew}). The results for the word `Right' indicate that it is frequently used in medicine related categories such as `Neuroscience' and `Surgery', and in social science categories such as `Law' and `Political Science'.  This is an expected result as it can appear to determine the side of organs such as `right hippocampus' or `right hemisphere' in medicine; and the normative rules in such disciplines as law and ethics. `Elsevier' and `Reserv' are much more uniformly distributed to the categories when the rank of categories is taken into account. For `Reserv', one can identify categories related to Biosciences such as `Ecology', `Zoology' and `Environmental Studies'. Specifically, this word occurs to indicate `nature reserves'.

	\begin{table}[h]
		\centering
		\caption{The number of occurrence of some words appearing in copyright notices before and after cleaning}
		\renewcommand\arraystretch{1.3}
  			\begin{tabular}{ | m{2cm} | R{4cm} | R{4cm}|}
	  	 		\hline
	  	 \rc \textbf{Word}				&  \rc  \textbf{Number of Documents Containing the Word Before Cleaning }     & \rc \textbf{Number of Documents Containing the Word After Cleaning}			 \tn \hline
	  	 \rr Elsevi 						& 314,204       & 80					 \tn \hline
	     \rr Right 						  	& 306,075		& 27,279				 \tn \hline
		 \rr Reserv 	 					& 288,193		& 5,761					 \tn \hline	
		 \rr Ltd							& 147,466		& 830		 			 \tn \hline	
		 \rr Springer 		 				& 296			& 187					 \tn \hline
		 \rr Copyright 		 				& 21,160		& 396					 \tn \hline
			   
			\end{tabular}
\label{table:beforeafter}
   \end{table}

\begin{landscape}\centering
\vspace*{\fill}
\begin{figure}[htpb]
  \centering
  \includegraphics[height=0.6\textheight, width=1.5\textwidth]{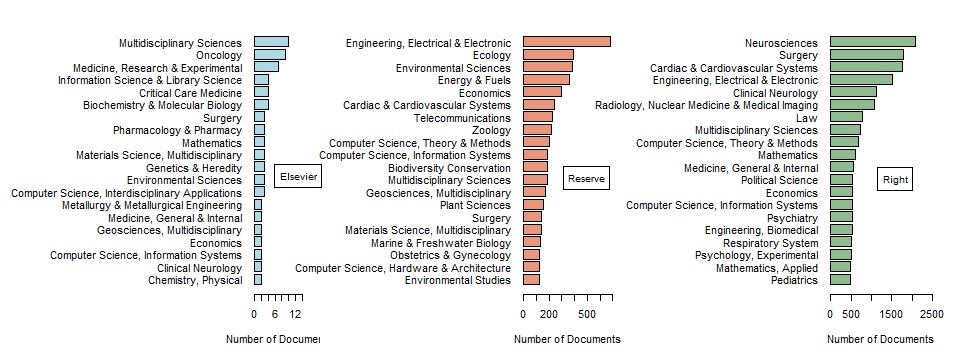}
  \caption{Top 20 categories that words `Elsevier', `Reserve' and `Right' appear in (after cleaning)}
  \label{fig:wordstogethernew}
\end{figure}
\vfill
\end{landscape}

\subsection{Words Represented by Vectors of Frequencies in Categories for the LSC \nopunct}\hspace*{\fill} \\\\

Recall that a representation technique for words was introduced in Section \ref{rep}. The vectors of frequencies in subject categories were obtained for each word. The frequency associated to a category was computed by counting texts containing the word in this category.

\begin{figure}[tb]
\centering
 \includegraphics[width=1\linewidth]{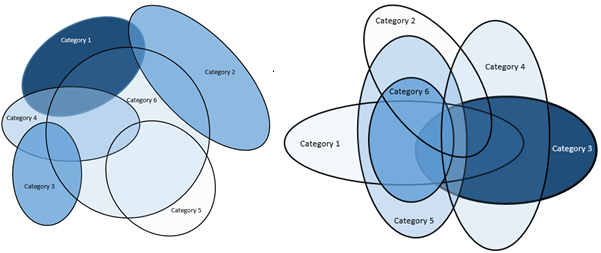}
   \caption{Two examples of intersection of categories in the LSC. The maximal number of categories that a document belongs to is 6 in the LSC.}
  \label{fig:LSCcategories}
\end{figure}

Subject categories are used to categorise papers in the WoS collection; however, documents do not necessarily belong to a unique category due to interdisciplinary studies. In other words, categories are not exclusive in WoS and so in the LSC. In the LSC, texts belong to at least 1 and a maximum of 6 categories out of a total of 252 subject categories (see Figure \ref{fig:LSCcategories}). It is noteworthy that our consideration is to count the number of times a word appears in texts of a category rather than analysing exclusivity of categories. Therefore, in this stage, we just looked at the frequency of texts with these words in categories. 

The vectors of frequencies in categories are built for 103,998 LScDC words and 252 subject categories. Each row represents a word of the LScDC in 252-dimensional space, that is, each word is represented by a vector of frequencies in 252 categories. For each category, a frequency distribution can be obtained for the set of words. The distribution indicates words used in texts of each category and the most frequently used words can be sorted in categories. To illustrate this, the most frequent 10 words for categories `Astronomy \& Astrophysics', `Mathematics' and `Asian Studies' with frequencies are displayed in Table \ref{table:3categories}. A table containing all words and categories are included in \cite{wordcat}. One can expect that not all words in the table indicate a topic in the related subject. As an example, words `use', `also', `studi'  and `paper' are frequent words in the LScDC and so in categories. These non-topic specific words occur many times in abstracts without indicating subject specificity. Therefore, using the frequencies of words in categories may not reflect how specific a word is to a category.

\begin{table}[h]
		\centering
		\caption{The most frequent 10 words for categories Astronomy \& Astrophysics, Mathematics and Asian Studies}
		\renewcommand\arraystretch{1.3}
\begin{tabular}{ | m{1.5cm} | R{1.5cm} | m{1.5cm}| R{1.5cm} | m{1.5cm} | R{1.5cm}|}
	  	 		\hline
 \multicolumn{2}{|C{3cm}|}{\textbf{Astronomy \& Astrophysics}}	&  \multicolumn{2}{C{3cm}|}{\textbf{Mathematics}}  &   \multicolumn{2}{C{3cm}|}{\textbf{Asian Studies}} \\ \hline

\rr use 	& 11,100 	&\rr paper		&	9,408	&\rr articl	&415 	\tn \hline
\rr observ 	& 10,237	&\rr result		&	9,074	&\rr examin	&236	\tn \hline
\rr model	& 9,295		&\rr prove		&	7,743	&\rr studi	&230	\tn \hline
\rr result	& 9,213		&\rr show		&	6,705	&\rr one	&226	\tn \hline
\rr present	& 7,810		&\rr space		&	6,224	&\rr argu	&226	\tn \hline
\rr can		& 7,598		&\rr also		&	6,194	&\rr also	&216	\tn \hline
\rr studi	& 7,350		&\rr studi		&	6,187	&\rr paper	&211	\tn \hline
\rr also	& 7,314		&\rr use		&	6,062	&\rr polit	&196	\tn \hline
\rr show	& 7,191		&\rr function	&	5,904	&\rr use	&195	\tn \hline
\rr similar	& 6,622		&\rr general	&	5,766	&\rr two	&192	\tn \hline

\end{tabular}
\label{table:3categories}
   \end{table}

\subsection{Word-Category RIG Matrix for the LSC \nopunct}\hspace*{\fill} \\\\

On the basis of exploratory work by the frequency table, we concluded that the use of word frequencies in categories does not provide much information about the category. To be specific, we expected that `use' is not a topic-specific word as it appears in all 252 categories and it is likely to be used in almost all texts. This means that the meaning of a word in the text cannot be directly extracted from the frequency.

Aiming at this result, we must now apply a different perspective to measure the importance of words for categories, with a special attention given to the hypothesis that each word in the LScDC has scientifically specific meaning in categories and the meaning can be extracted from the information of words for 252 subject categories in the LSC. Thus, as described in Section \ref{rep}, words were represented in a 252-dimensional Meaning Space. RIGs for each word in 252 categories were calculated and vectors of words were formed. We then represented these vectors in the Word-Category RIG Matrix. 

For each word in the Word-Category RIG Matrix, the sum $S_{j}$ and maximum $M_{j}$ of RIGs in categories were calculated and added at the end of the matrix. The Word-Category RIG Matrix can be found in \cite{wordcat}. One can extract the most informative $n$ words for scientific texts by ordering/sorting the column of words based on their $ S_{j} $ or $M_{j}$.   

\subsection{Results\nopunct}\hspace*{\fill} \\\\

The experimental results presented in this section were obtained using abstracts of academic research papers in the LSC \cite{LSCn}. We used words from the core dictionary LScDC \cite{LScDCn}.

Having calculated RIGs for each word and created the Word-Category RIG Matrix, we evaluate the representation model by checking words in each category. That is, we consider the list of words with their RIGs in the corresponding category. Those words that have larger RIG are more informative in the category.  Being `more informative' here allows for the interpretation of being `more specific' to the category's topic.

For each category, words are sorted by their RIGs and the top 100 words are shown in the word clouds. The bigger font size the word in word clouds, the more informative it is. Word clouds for the top 100 most informative words and histograms of RIGs for the top 10 most informative words for each of 252 categories can be found in \cite{fig_tabl}. The most informative 100 words with their RIGs for each of categories are presented in Appendix \ref{InfCat} and \cite{fig_tabl}.

\begin{figure}[bt]
\centering
 \includegraphics[width=.8\linewidth]{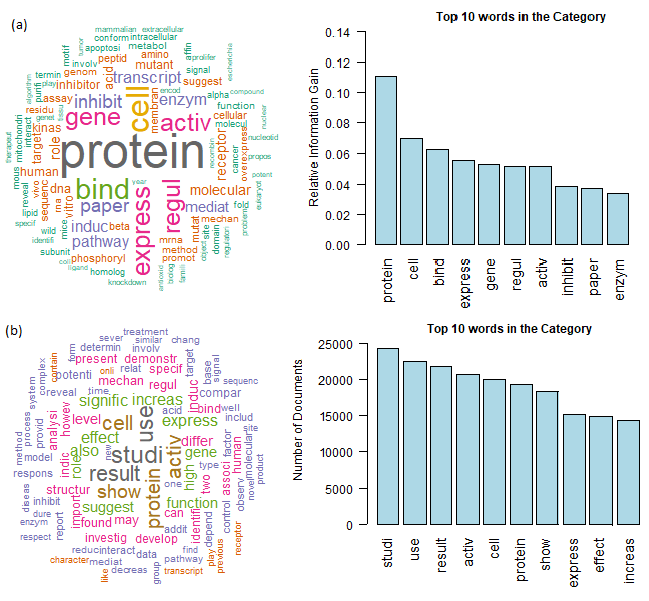}
   \caption{Category `Biochemistry \& Molecular Biology': word cloud of the top 100 most informative words and the histogram of the the top 10 most informative words. The informativeness is defined by (a) RIG (b) frequency.}
  \label{RIG1_BiochemistryMolecular Biologyfreq}
\end{figure}

\begin{figure}[bt]
\centering
 \includegraphics[width=.8\linewidth]{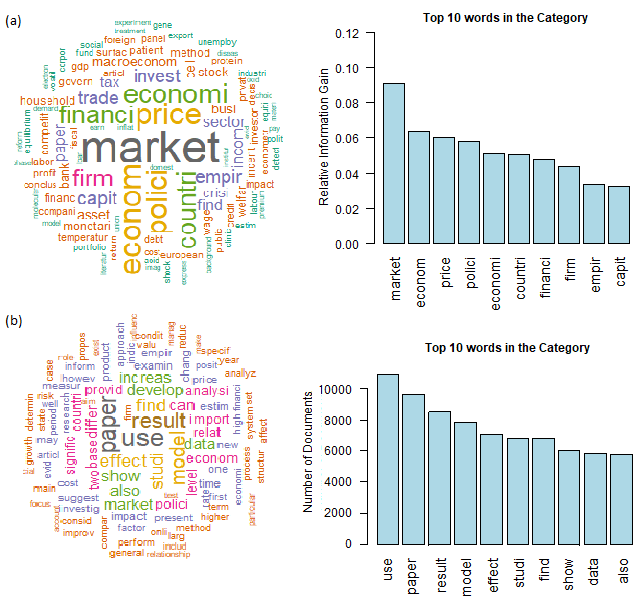}
   \caption{Category `Economics': word cloud of the top 100 most informative words and the histogram of the top 10 most informative words. The informativeness is defined by (a) RIG (b) frequency.}
  \label{RIG1_Economicsfreq}
\end{figure}

\begin{figure}[bt]
\centering
 \includegraphics[width=.8\linewidth]{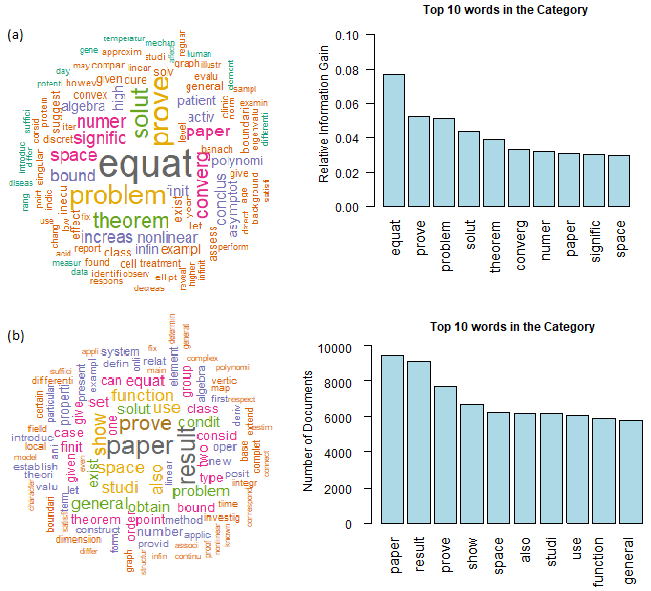}
   \caption{Category `Mathematics': word cloud of the top 100 most informative words and the histogram of the top 10 most informative words. The informativeness is defined by (a) RIG (b) frequency.}
  \label{RIG1_Mathematicsfreq}
\end{figure}

In general, the RIG based method proves to be more sensitive than the frequency based method in identifying topic-specific words of a category. This means that representing words in Meaning Space has the advantage of transforming words to efficient vectors with a benefit of considerably lower dimension than the standard word representation schemes. To illustrate this result, we choose categories `Biochemistry \& Molecular Biology', `Economics' and `Mathematics' and compare two word clouds that are formed by using raw frequencies and RIGs in categories (see Figures \ref{RIG1_BiochemistryMolecular Biologyfreq}, \ref{RIG1_Economicsfreq} and \ref{RIG1_Mathematicsfreq}). It can be seen from the figures that the majority of the most frequent words in all three categories are frequent words in the entire corpus. These words are not topic-specific for categories as they appear in almost all abstracts. The frequent but non-informative words can be considered as {\it generalised service words of Science} and deserve special analysis. 

This proves that raw frequency is not much important to identify scientifically specific meanings of words. Therefore, by representing words as vector of RIGs, we can avoid such frequency bias. The most informative words in categories for RIG representation are topic-related in the corresponding category. We interpret these results as evidence for the usefulness of the RIG based representation.

Words that are expected to be used together have very close values of RIGs. In `Health Care Sciences \& Services', `health' and `care' are top words and RIGs for these words are so close (see Figure \ref{fig:RIG1_105}). Another example is `xrd' and `difract' in `Material Science, Ceramics'. `XRD' is actually abbreviation of `X-ray diffraction'; therefore, they appear together as `X-ray diffraction (XRD)' for most of cases in the category (see Figure \ref{fig:RIG1_139}).

We can extract some stylistic properties in texts of categories. For instance, in computer science related categories the word `paper' has the highest RIGs (see Figures \ref{fig:RIG1_48}, \ref{fig:RIG1_45}, \ref{fig:RIG1_42}, \ref{fig:RIG1_46}, \ref{fig:RIG1_44}, \ref{fig:RIG1_47} and \ref{fig:RIG1_43}). This may be result of the stylistic features in papers for such categories. 

\begin{figure}[bt]
\centering
 \includegraphics[width=.8\linewidth]{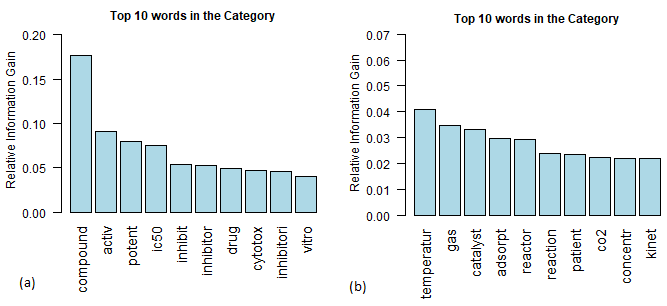}
   \caption{Histograms of the most informative 10 words in categories (a) Chemistry, Medicinal and (b) Engineering, Chemical}
  \label{chems2}
\end{figure}

A casual observation indicates that while the most informative words in some categories have similar RIGs, differences in values of RIGs are much more noticeable for the most informative words in some other categories. To give an insight, we present the categories `Chemistry, Medicinal' and `Engineering, Chemical' in Figure \ref{chems2}. In `Chemistry, Medicinal', the word `compound' can be easily separated from the other words, while in `Engineering, Chemical', there is a slight decrease in RIGs for the top 10 most informative words. However, in general we did not observe any explicit rule for this property.

\begin{figure}[bt]
\centering
 \includegraphics[width=.8\linewidth]{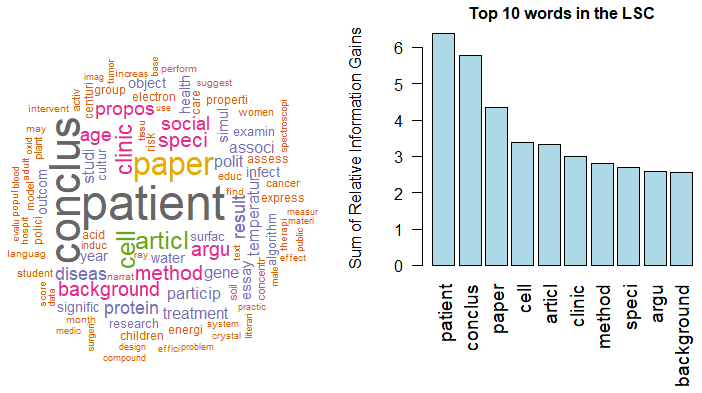}
   \caption{The most informative 100 words in the LSC. Words are arranged and selected by the sum of their RIGs in 252 categories. The font size and colour of words indicate different sum of RIGs. The histogram shows the sum of RIGs for the top 10 informative words.}
  \label{fig:wordcloudsSum}
\end{figure}

\begin{figure}[bt]
\centering
 \includegraphics[width=.8\linewidth]{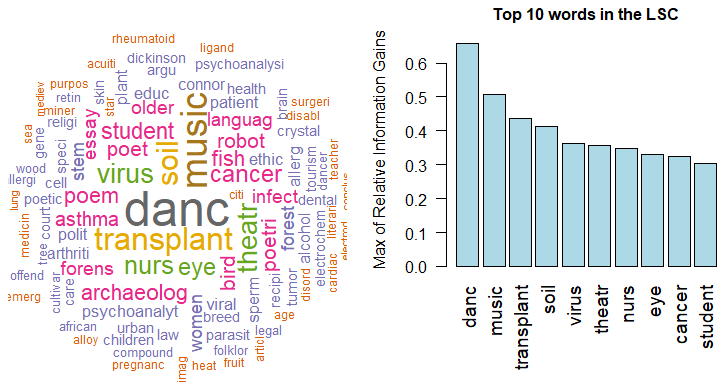}
   \caption{The most informative 100 words in the LSC. Words are arranged and selected by the maximal RIG over 252 categories. The font size and colour of words indicate different maximal RIG. The histogram shows the maximal RIG for the top 10 informative words.}
  \label{fig:wordcloudsMax}
\end{figure}

Finally, we formed two lists of words that arranged in descending order based on the sum and maximum of their RIGs in 252 categories. The top 100 words in two lists are displayed by word clouds in Figure \ref{fig:wordcloudsSum} and Figure \ref{fig:wordcloudsMax}. Histograms in the figures show the most informative 10 words in the lists. We found that the most informative 10 words in two lists are completely different, as shown in the figures. From words clouds, one can see that the majority of the first 100 words do not match. We then compared two lists by counting the number of matches in the top $n$ words, where $n$ ranges from 100 to 50,000. The numbers of matched words for different $n$ are presented in Table \ref{summax}.  As can be seen, 18\% of words match for the top 50 most informative words. This proportion increases to approximately 50\% for the top 1,000 words and to 58\% for the top 2,000 words. The intersection of lists reaches to approximately 99\% for the top 50,000 words. From these results, one can conclude that two lists are different in the top words. When higher number of words is taken into account, lists become more similar in terms of words included. However, the rank of words are not similar. Any of these criteria for selecting the most informative words can be used depending on the task and the information required. 

\begin{table}
\centering
\caption{Comparison of words ordered by the maximum $M_{j}$ RIG and sum $ S_{j} $ of RIGs in categories}
\renewcommand{\arraystretch}{1.5}

\begin{tabular}{| R{3cm} |  R{3cm} |R{3cm} | }
\hline
  	  \multicolumn{1}{|C{3cm}|}{\textbf{Top (n) Words in Two Lists}} & \multicolumn{1}{C{3cm}|}{\textbf{Number of Matches}} & \multicolumn{1}{C{3cm}|}{\textbf{Fraction of Matches}}  \\\hline
10&0&0.000 	  \\\hline
50 & 9 &0.180 \\\hline
100 & 28&0.280  \\\hline
500 & 189&0.378 \\\hline
1,000 & 498 &0.498 \\\hline
2,000 & 1,168&0.584\\\hline
5,000 & 3,412& 0.682 \\\hline
10,000 & 7,492&0.749  \\\hline
50,000 & 49,542&0.991  \\\hline
103,998 & 103,998&1.000  \\\hline

\end{tabular}
 \label{summax}
\end{table}

\begin{figure}[p]
\centering
 \includegraphics[width=.8\linewidth]{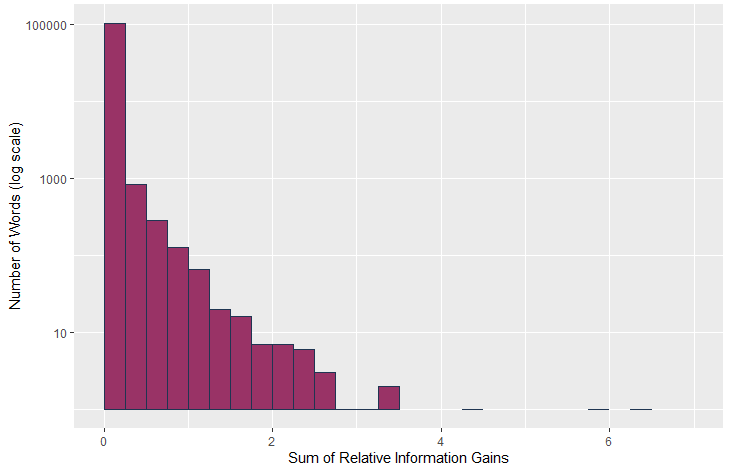}
   \caption{Histogram of the sum of RIGs for words of the LScDC (logarithmic scale for the y-axis)}
  \label{fig:sumlog3}
\end{figure}

\begin{figure}[bt]
\centering
 \includegraphics[width=.8\linewidth]{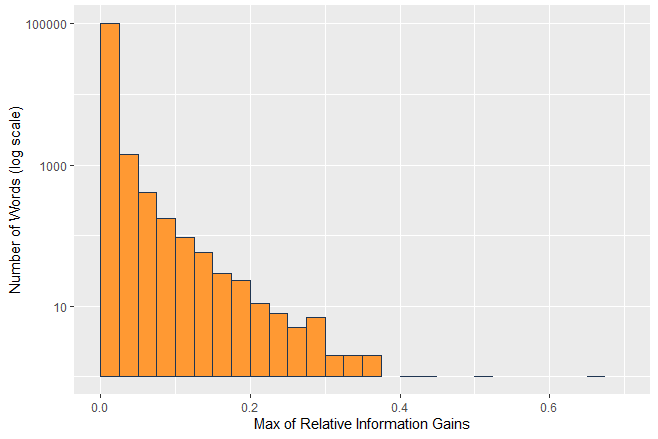}
   \caption{Histogram of the maximum of RIGs for words of the LScDC (logarithmic scale for the y-axis)}
  \label{fig:maxlog2}
\end{figure}

The numbers $ S_{j} $ and $M_{j}$ are differently distributed for words. We observed from the lists that many words have low $ S_{j} $ and $M_{j}$. Figure \ref{fig:sumlog3} and Figure \ref{fig:maxlog2} show the distribution of $ S_{j} $ and $M_{j}$ for words in the logarithmic scale. Supper-exponential picks near zero RIGs are noticeable for both criteria. We can see that the trend is going down almost linearly beyond the picks. The bottom 10 least informative words in two lists are presented in Table \ref{table:bottom}. One may consider words having almost zero $ S_{j} $ or $M_{j}$ as less meaningful words for scientific texts.   

\begin{table}[bt]
	\centering
	\caption{The least informative 10 words that are arranged in ascending order based on the sum $ S_{j} $ and the maximum $ M_{j} $ of RIGs in 252 categories. Words are in stemmed form.}
	\renewcommand\arraystretch{1.3}
  	\begin{tabular}{|m{2cm}|m{2cm}|m{2cm}|m{2cm}|}
	\hline
	  	\multicolumn{2}{|C{4cm}}{\textbf{Words in the list where the sum of RIGs is calculated}} &  \multicolumn{2}{|C{4cm}|}{\textbf{Words in the list where the maximum of RIGs is calculated}} \\ \hline
	  	\rc \textbf{Word} &\rc \textbf{$ S_{j} $} & \rc \textbf{ Word} & \rc \textbf{$ M_{j} $}  \tn \hline
	  	\rr tgvs & \rl 0.000301 &  \rr vhp &\rl 0.0000149  \tn \hline
	 	\rr nonisland &\rl 0.000302& \rr msmc &\rl 0.0000154    \tn \hline
	  	\rr antipad &\rl 0.000308 &\rr scandat  &\rl  0.0000179 \tn \hline
	  	\rr aigan &\rl 0.000323 &\rr metaloxid  &\rl 0.0000195  \tn \hline
	  	\rr ultrabook &\rl  0.000324 &\rr ntfs &\rl 0.0000195  \tn \hline
	  	\rr inzno &\rl 0.000324 &\rr interg  &\rl 0.0000196  \tn \hline
	  	\rr semiparallel &\rl 0.000328 &\rr nonfeas  &\rl 0.0000196  \tn \hline
	  	\rr 22dbm &\rl 0.000328 &\rr biperiod &\rl 0.0000206 \tn \hline
	  	\rr biperiod &\rl 0.000329 &\rr microl  &\rl 0.0000206  \tn \hline
	  	\rr 150mhz &\rl 0.000330 & \rr multiparallel &\rl 0.0000206   \tn \hline
	\end{tabular}

\label{table:bottom}
   \end{table}

\section{{Thesaurus for Science: Leicester Scientific Thesaurus (LScT)}}\label{thes}
In this section, we introduce a scientific thesaurus of English: Leicester Scientific Thesaurus (LScT). LScT is a list of 5,000 words which are created by arranging words of LScDC in their informativeness in the scientific corpus. The procedure for creation of the thesaurus is described in detail.

Under the assumption that not all words having very low RIGs are informative in categories, we search a cut-off point for RIG to create a list of words that can be considered as relatively meaningful in scientific texts. In other words, we extract meaningful words for science from the LScDC to build a scientific thesaurus. Before moving on the decision taken to determine the number of words for the thesaurus, we recall the notion `informativeness' and investigate further the criteria of $ S_{j} $ and $M_{j}$ to arrange words of LScDC in their informativeness.

Having the top 100 words in two lists where words are descending ordered by their $ S_{j} $ and $M_{j}$, we see that the criteria of maximum is more likely to stand out some words that are frequently used in specific categories such as categories `Dance', `Music', `Soil Science' and `Theatre' (see Table \ref{table:2lists}) and are relatively rarely used outside them. Indeed, we expect drastic differences in RIGs of such words for these categories. For instance, one of the most informative word `dance' is used in 154 categories, but the RIG from this word to the category `Dance' is very distinguishable from all others (see Table \ref{table:danc}).  This is actually an expected result, since the word `danc' is likely to be informative for categories related to the performing arts.
 
\begin{table}[bt]
	\centering
	\caption{Top 10 most informative words that are arranged in descending order based on the sum $ S_{j} $ and the maximum $ M_{j} $ of RIGs in 252 categories. These words are in the stemmed form.}
	\renewcommand\arraystretch{1.3}
  	\begin{tabular}{|m{2cm}|m{2cm}|m{2cm}|m{2cm}|}
	\hline
	  	\multicolumn{2}{|C{4cm}}{\textbf{List ordered by the sum $ S_{j} $ of RIGs}} &  \multicolumn{2}{|C{4cm}|}{\textbf{List ordered by the maximum $ M_{j} $ of RIGs}} \\ \hline
	  	\rc \textbf{Word} &\rc \textbf{$ S_{j} $} & \rc \textbf{ Word} & \rc \textbf{$ M_{j} $}  \tn \hline
	  	\rr patient & \rl 6.382 &  \rr danc &\rl 0.657  \tn \hline
	 	\rr conclus &\rl 5.766 & \rr music &\rl  0.508  \tn \hline
	  	\rr paper &\rl 4.345 &\rr transplant  &\rl  0.435 \tn \hline
	  	\rr cell &\rl 3.394 &\rr soil  &\rl 0.413  \tn \hline
	  	\rr articl &\rl  3.336 &\rr virus &\rl 0.362  \tn \hline
	  	\rr clinic &\rl 3.003 &\rr theatr  &\rl 0.358  \tn \hline
	  	\rr method &\rl 2.797 &\rr nurs  &\rl 0.348  \tn \hline
	  	\rr speci &\rl 2.686 &\rr eye &\rl 0.329  \tn \hline
	  	\rr argu &\rl 2.581 &\rr cancer  &\rl 0.324  \tn \hline
	  	\rr background &\rl 2.562 & \rr student &\rl 0.302   \tn \hline
	\end{tabular}

\label{table:2lists}
   \end{table}

\begin{table}[bt]
	\centering
	\caption{Five categories with the highest RIGs of the word `danc'}
	\renewcommand\arraystretch{1.3}
  	\begin{tabular}{|m{3cm}|m{1.5cm}|}
	\hline
	  	\rc \textbf{Category} & \rc \textbf{RIG}  \tn \hline
	  	\rr Dance &\rl 0.657  \tn \hline
	 	\rr Theatre &\rl 0.042  \tn \hline
	  	\rr Music &\rl 0.024  \tn \hline
	  	\rr Folklore &\rl 0.009  \tn \hline
	  	\rr Literary Reviews &\rl 0.008  \tn \hline
	\end{tabular}

\label{table:danc}
   \end{table}

To compare the meaningfulness of words across all categories, we tested two norms in the Meaning Space, $l_1$ ($S_j$ or the sum of RIGs) and $l_\infty$ ($M_j$ or the maximal RIG). After a series of trials, we decided to use $l_1$. This choice cannot be proven formally but the   ordering words by $M_{j}$ lead to some words that are very specific in only one category but stand out in the list of the most informative words on average. The sum can be considered as more appropriate measure for general scientific thesaurus. When creating an LScT, we consider ordering the LScDC words by the sum of their RIGs in categories. The meaningfulness of words was evaluated by the average informativeness of words in the categories. Given the dictionary LScDC, the procedure to create the LScT is: 

\begin{itemize}
\item Sort the words of the LScDC by their $ S_{j} $ in descending order.
\item Take the top 5,000 words.
\end{itemize}

To find the number of words to be contained in the thesaurus, we initially follow an empirical procedure: 
\begin{enumerate}
\item Having arranged list of words in descending order by $ S_{j} $, take a sub-list of the top $m$ words, denoted by $T_{m}$
\item Create the histogram of $ S_{j} $ for the words in this sub-list
\item Check the trend in the histogram
\item Take words when the exponential pick is avoided and the histogram follows roughly linear trend.
\end{enumerate}

\begin{figure}[p]
  \centering
  \includegraphics[height=.8\textheight, width=1\textwidth]{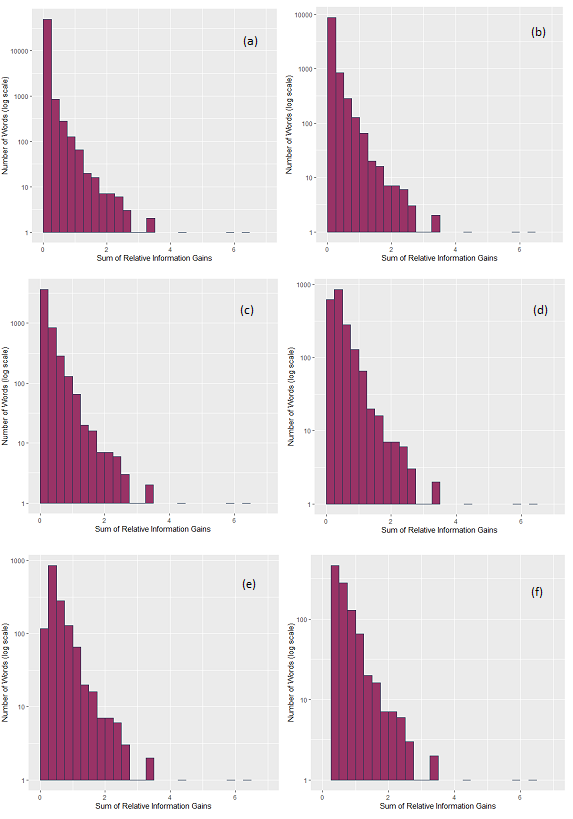}
  \caption{Histograms of the sums of RIGs for words of the LScDC (logarithmic scale for the y-axis): (a) The top 50,000 words (b) The top 10,000 words (c) The top 5,000 words (d) The top 2,000 words (e) The top 1,500 words (f) The top 1,000 words}
  \label{fig:allSUM}
\end{figure}

We begin with investigating the top 50,000 words in arranged list as it is almost the half of the 103,998 words of the LScDC. As the trend in the histogram for 50,000 words was showing the same behaviour with the histogram of 103,998 words (see Figure \ref{fig:sumlog3} and Figure \ref{fig:allSUM} (a)), there was no point to check a number between 50,000 and 103,998. We then decreased the number $m$ to 10,000, 5,000, 2,000, 1,500 and finally 1,000. All histograms are presented in Figure \ref{fig:allSUM}. We see a substantial change in the trend of the histogram when we take the subset of 5,000 words. The trend at that point is almost linear. After that, the first bin in the histogram is slightly becoming smaller and finally it disappears for 1,000 words.

In this step, we also checked the minimum of the sum of RIGs in the lists $T_{m}$ to make sure that the minimal average informativeness in the list (to be selected) is not so close to zero. These values are displayed in the Table \ref{table:minRIG}. We can see from the table that the minimal RIG is decreased less than half from 1,000 to 2,000, while it decreased faster (more than halved) from 5,000 to 10,000. 

\begin{table}[bt]
	\centering
	\caption{The minimal sum of RIGs in the list of the top $m$ words ($T_{m}$)}
	\renewcommand\arraystretch{1.3}
  	\begin{tabular}{|m{2cm}|m{2cm}|}
	\hline
	  	\rc \textbf{List} & \rc \textbf{Minimal $ S_{j}$}  \tn \hline
	  	\rr $T_{1,000}$ &\rl 0.3208  \tn \hline
	 	\rr $T_{1,500}$ &\rl 0.2329  \tn \hline
	  	\rr $T_{2,000}$ &\rl 0.1797  \tn \hline
	  	\rr $T_{5,000}$ &\rl 0.0658  \tn \hline
	  	\rr $T_{10,000}$ &\rl 0.0268  \tn \hline
	  	\rr $T_{50,000}$ &\rl 0.0027  \tn \hline
	  	\rr $T_{103,998}$ &\rl 0.0003  \tn \hline
	\end{tabular}

\label{table:minRIG}
   \end{table}

Finally, to support our selection of the number of words for the LScT and for evaluation of the result, we consider the following heuristic suggestion: the majority of words in the LScT appears in the list of top  informative words in the categories. This does not mean that all informative words in categories should appear in the LScT, but we expect that most of top $n$ words in categories will be included in the LScT. This is a natural result of the statistics (average) used for selection of the LScT. If a word in a specific category is informative with high RIG for only this category, this word may not appear in the LScT as we considered the average informativeness over categories. 
 
We consider the matches of the list $T_{m}$ with the most informative words in categories defined by the sum of RIGs. For collection $ C_{k,n} $ of $n$ most informative words in the category $k$, we define $X_{n}=\bigcup_{k=1}^K C_{k,n}$. Then we test the coverage of the list $T_{m}$ by $X_{n}$. For each category in the Word-Category RIG matrix (a column in Table~\ref{table:RIGs}), order in descending by their RIGs. This gives a list of words sorted from the most informative to the least informative for this specific category. Then, individual collections $ C_{k,n} $ are formed for each category.    

\begin{figure}[htpb]
  \centering
  \includegraphics[width=.6\textwidth]{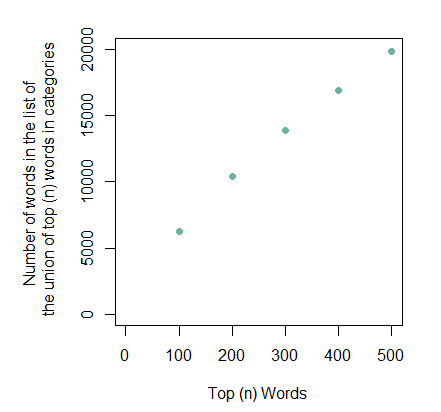}
  \caption{Number of words in the list of union of the top ($n$) words in categories ($X_{n}$)}
  \label{fig:union}
\end{figure}

The set $\bigcup C_{k,n}$ was formed with different numbers of words ($n$). We built the collections containing the most informative 100, 200, 300, 400 and 500 words in each category, and $X_{n}$ is created by uniting them for each $n$. The numbers of words and the minimal RIGs of words in $X_{n}$ are presented in Table~\ref{table:union}. The minimal RIGs are checked to avoid zero/near-zero RIG in lists. One can see from the table that words in categories are not completely different. For instance, if all $C_{k,100}$ do not intersect then there should be 25,200 words in the list $X_{100}$, but there are just 6,254 words in this union, which is almost four times less. For other $n$, the result is similar, and the values of $X_{n}$ follow almost a linear trend (see Figure~\ref{fig:union}). That is, the intersection  $\bigcap_{k=1}^K C_{k,n}$ is not empty. The intersection may be pairwise or $q$-wise for different $q$.  

\begin{table}[bth]
	\centering
	\caption{The number of words in the list $X_{n}$ and the minimal RIGs for $X_{n}$}
	\renewcommand\arraystretch{1.3}
  	\begin{tabular}{|m{1cm}|m{3.5cm}|m{2cm}|}
	\hline
	  	\rc $n$ & \rc \textbf{Number of words in the list $X_{n}$ } & \rc \textbf{Minimal RIG}     \tn \hline
	  	\rl 100 &\rl 6,254 &\rl 0.0025  \tn \hline
	 	\rl 200 &\rl 10,435 &\rl 0.0016  \tn \hline
	  	\rl 300 &\rl 13,850 &\rl 0.0012  \tn \hline
	  	\rl 400 &\rl 16,910 &\rl 0.0009  \tn \hline
	  	\rl 500 &\rl 19,790 &\rl 0.0008  \tn \hline
	  	
	\end{tabular}

\label{table:union}
   \end{table}

The coverage is calculated by counting the number of matches words of the list $T_{m}$ and words of $X_{n}$.  Table \ref{table:matces} illustrates the numbers of matches when $n$ is 100, 200 and 500. Up to top 2,000 words, the words are concordant in the lists $T_{m}$ and $X_{n}$, suggesting that the most informative words are highly consistent. In fact, words in the lists are in agreement for the case where 500 words in each category are considered as informative for categories. Given the list $X_{100}$, the majority of the words (3,992 words) in the $T_{5,000}$ can be covered by words of $X_{100}$. This trend changes and goes down when we consider the percentage of words found in 10,000 and 50,000 words. However, in this stage we have to consider the total number of words in $X_{n}$. For instance, the number of words in $X_{100}$ is 6,254. In this case, the matches cannot be more than this number for the list $T_{10,000}$. Similar conclusions were obtained by comparing the number of matches for 5,000, 10,000 and 50,000 words for $n$=200.

\begin{table}[bth]
	\centering
	\caption{Number of matched words of the list $T_{m}$ and words of $X_{n}$. $n$ is the number of words taken from each category to create $X_{n}$. }
	\renewcommand\arraystretch{1.3}
\begin{tabular}{|c|c|c|c|}
\hline
\backslashbox{\textbf{$T_{m}$}}{ \strut \textbf{$X_{n}$} }&\textbf{$X_{100}$} &\textbf{$X_{200}$ }& \textbf{$X_{500}$}  \\\hline
\textbf{$T_{1,000}$} & 995  & 1,000  &1,000   \\\hline
\textbf{$T_{1,500}$} & 1,469  & 1,499  &1,500   \\\hline
\textbf{$T_{2,000}$} & 1,908  & 1,992  &2,000   \\\hline
\textbf{$T_{5,000}$} & 3,992  & 4,674  &4,987   \\\hline
\textbf{$T_{10,000}$} & 5,631  & 7,745  &9,588   \\\hline
\textbf{$T_{50,000}$} & 6,254  & 10,435  &19,754   \\\hline
\textbf{$T_{103,998}$} & 6,254 & 10,435  &19,790   \\\hline
\end{tabular}
\label{table:matces}
   \end{table}

We examined various heuristic criteria to evaluate how many words are suitable for inclusion in a thesaurus. Since we want to keep the size of thesaurus reasonable, and pay attention not to loose many words in case there might be informative words having not very high RIGs, we decided to include these 5,000 words ($T_{5,000}$) in the scientific thesaurus. This thesaurus is called Leicester Scientific Thesaurus (LScT). It  is published online \cite{wordcat}.

 \section{Conclusion and Discussion}\label{conc}
In this work, we have studied the first stage of `quantifying of meaning' for scientific texts: constructing the space of meaning. We have introduced the \textit{Meaning Space} for scientific texts based on computational analysis of situations of words' use. The situation of use of the word is described by the absence/presence of the word in the text in scientific subject categories. The meaning of the text is hidden in the situations of usage and should be extracted by evaluating the situation related to the text as a whole.

This research is done based on 1,673,350 texts from the LSC and its 103,998 words listed in the LScDC \cite{LSCn,LScDCn}. A text in the LSC belongs to at least one and at most six of 252 Web of Science categories presented in Table \ref{table:categories}. That is, categories can intersect. The situation of use is described by these 252 binary attributes of the text. These attributes have the form: a text is present (or not present) in a category. The meaning of a word is determined by categorising texts that contain the word and texts that do not. It is represented by the 252-dimensional vector of RIG  about the categories that the text belongs to, which can be obtained from observing the word in the text. This representation is demonstrated in Table \ref{table:RIGs}. Each text in the LSC can be considered as a cloud of these RIG vectors. 

We begin with representing each word as a vector of frequencies in categories (Table \ref{table:tabrepr}). Components of a vector are the number of texts containing the word and belonging to the corresponding category. 
Then we moved on to representing the meaning of a word as a RIGs vector about categories.

We consider the corpus (LSC) as a probabilistic sample space (the space of equal probable elementary outcomes). The function is defined on texts that takes the value 1, if the text belongs to the category $ c_{k} $, and 0 otherwise. Similarly, for each word $w_{j}$, a function is defined on texts that takes the value 1 if the word $w_{j}$ belongs to the text, and 0 otherwise. Both functions can be considered as the random Boolean variables.  The information gain $IG(c_{k},w_{j} )$ about the category $ c_{k} $ from the word $ w_{j} $ is calculated by (\ref{eq:IGs}), (\ref{eq:ent1}) and (\ref{eq:ent2}). $IG(c_{k},w_{j} )$ measures the amount of information extracted from observing the word $w_{j}$ in the text on prediction of belonging of this text to the category $c_{k}$. The RIG $RIG(c_{k},w_{j} )$ is calculated by (\ref{eq:RIG}) that provides us a normalised measure of Information Gain giving the ability of comparing information gains for different categories. 

Vectors of RIGs are denoted by $\overrightarrow{RIG_{j}}$ for a word  $ w_{j} $. $\overrightarrow{RIG_{j}}$ vectors for all words are presented in a \textit{Word-Category RIG Matrix} (see the structure in Table \ref{table:RIGs}) (available online \cite{wordcat}). A column vector of the matrix contains RIGs for all words in an individual category and a row vector represents the corresponding word's meaning as a vector of RIGs for categories. The \textit{Meaning Space} has been described as a 252-dimensional vector space, where vectors are $\overrightarrow{RIG_{j}}$. Beyond the representation of words, the Word-Category RIG Matrix can be also used for the ordering words in a category from the most informative to the least informative as well as identifying the most informative words in the science for different subjects and their combinations. Ranking of words in a scientific corpus are performed based on two criteria: sum of RIGs ($S_{j}$) and maximum of RIGs ($M_{j}$) in a row vector. Calculations are done by (\ref{eq:sum}) and (\ref{eq:max}). Given an ordered list of words, the top $ n $ words are considered as the most informative $ n $ words in the scientific corpus. 

The LSC and LScDC were created and available online  \cite{our, LSC,LScDC}. The proposed word representation technique was applied to this version of the corpus. The evaluation of the model is done based on checking the most informative words in each category. Word clouds are generated using words in lists for each category (For example, see Figure \ref{fig:acoustic_old} and Figure \ref{fig:Chemistry_app_old}). The higher RIG a word has, the bigger font size of the word is in the cloud. The clouds demonstrated that our methodology is able to identify topic specific words for categories, and most of the top words are related to the category subjects.

We note, however, that  some words that were not expected to be appearing as the most informative words were prominent for some categories (Figure \ref{fig:Chemistry_app_old}). We concluded that words occurring in copyright notices, permission policies and the names of journals and organisations are added at the footer of abstracts in WoS database (see Table \ref{table:copyrightex}, Table \ref{table:scenarios} and Table \ref{table:addsch}). Such joints result in anomalies in the word clouds and our representation technique was able to detect them. A further cleaning on identified phrases, sentences and paragraphs was performed to avoid possible abnormal appearances of words in the lists. This is done by sampling of texts based on keywords search and then deleting them from the texts. After cleaning procedure, new versions of the LSC and the LScDC are created by the same pre-processing steps as for the previous versions and can be found in \cite{LSCn, LScDCn}.     

Words of LScDC were represented by vectors of RIGs in 252-dimensional Meaning Space as described before. The Word-Category Matrix for the LSC was formed with the collection of all words of the LScDC \cite{wordcat}. The sum $S_{j}$ and the maximum $M_{j}$ of RIGs in categories are calculated and added at the end of the matrix.  Word clouds with the top 100 words and histograms of the most informative 10 words for each category are presented in \cite{fig_tabl}. The most informative 100 words for each category with their RIGs can be found in Appendix \ref{InfCat} and \cite{fig_tabl}. The proposed model of RIG-based word representation is analysed through these top ranked words in each category. 

We have evaluated the Meaning Space by comparing our approach to traditional frequency-based model. Words in each category were also ranked and ordered by their raw frequencies in categories. It is proven that frequencies is not much important and efficient to represent scientifically specific meanings of words as the most frequent words are not topic related words such as `use', `studi' and `result'. Figure \ref{RIG1_BiochemistryMolecular Biologyfreq}, Figure \ref{RIG1_Economicsfreq} and Figure \ref{RIG1_Mathematicsfreq} compare two approaches using word clouds for three categories. The word clouds demonstrated that the information gain-based method is capable of standing topic-specific words out. This proves that the frequency is not much important in identifying such words. By representing words in the Meaning Space, we have shown by the human inspection that the top words in categories are topic-related in the corresponding category. It can therefore be viewed as an evidence of the usefulness of the Meaning Space and the representing words in this space.

$S_{j}$ and $M_{j}$ have been calculated for the LScDC words and two lists of words are created with words that are in descending orders by their $S_{j}$ and $M_{j}$. The lists enable sorting the most important $n$ words in science. We have compared these lists. The number of matches in the top $n$ words in two lists are counted, where $n$ is ranges from 100 to 50,000 (Table \ref{summax}). The top 10 words in two lists are completely different and only 28\% of words match in the first 100 words. This follows approximately 50\% for the first 1,000 and 58\% for the first 2,000 words. This concludes that two lists are not the same for the top words (Figure \ref{fig:wordcloudsSum} and Figure \ref{fig:wordcloudsMax}), however, both criteria can be used for selection of top $n$ words regarding to task and the information required. Many words in the lists have low  $S_{j}$ and $M_{j}$ values. The plot of the number of words for $S_{j}$ and $M_{j}$ indicate a super exponential picks near zero $S_{j}$ and $M_{j}$ (see Figure \ref{fig:sumlog3} and Figure \ref{fig:maxlog2}). The trend beyond the pick is going down almost linearly. Those words with near zero values can be considered as less meaningful words for scientific texts.  

Finally, a scientific thesaurus of English, named Leicester Scientific Thesaurus (LScT), has been introduced. The thesaurus contains of the most informative 5,000 words from the LScDC. Words in the LScT are selected by their average RIGs in categories. That is, the top 5,000 most informative words in the LScDC, where words are arranged by their $S_{j}$ are considered as the most meaningful 5,000 words in scientific texts. The full list of words in the LScT with their $S_{j}$ can be found in \cite{wordcat}.      

The next focus of the research in `quantifying of meaning' will be extraction of the meaning of text in scientific corpus from the clouds of words in the Meaning Space and study of more complex models in which co-occurrence of words and combination of word's meaning will be used. This, we follow the road: Corpus of texts + categories $\to$ Meaning Space for words $\to$  Geometric representation of the meaning of texts. The first two technical steps were done: the Corpus of texts was collected and cleaned, and the meaning of words was represented and analysed in the Meaning Space. The next step will be analysis of the meaning of texts. 

The analysis of dictionaries is not finalised yet. This work was focused on the most informative words. They are the main {\em scientific content words}. But, for example, the frequent but non-informative words (like `use') can be considered as  {\em generalised service words of Science} and deserve special analysis.  

It is also very desirable to extend the set of attributes for representation of the situation behind the text (Figures~\ref{fig:trainingset}, \ref{fig:trainingset2}). The first choice,  the research subject categories, is simple and natural, but it may be useful to enrich this list of attributes.

\section*{Appendices}
\appendix

\section{Some additional examples of notices.}
\raggedbottom
\begin{table}[t]
		\centering
		\caption{Some additional examples of notices attached to the abstract}
		\renewcommand\arraystretch{1.3}
  			\begin{tabular}{ | m{12cm} |}
	  	 		\hline
	  	 		\rc \textbf{Copyright Notice; Name of Conference, Journal or Publishing House} 													 \tn \hline
	  	 		\rr (C) 2014 Elsevier B.V. This is an open access article under the CC BY-NC-ND license (https://creativecommons.org/licenses/by-nc-nd/3.0/).
													 \tn \hline
	  			\rr Copyright (C) 2014, Taiwan Association of Obstetrics \& Gynecology. Published by Elsevier Taiwan LLC. All rights reserved.			 												 \tn \hline
	 		  	\rr (c) 2014 AIP Publishing LLC.							 \tn \hline
			    \rr (C) The Author(s) 2014. Published by ECS. This is an open access article distributed under the terms of the Creative Commons Attribution 4.0 License (CC BY, http://creativecommons.org/licenses/by/4.0/), which permits unrestricted reuse of the work in any medium, provided the original work is properly cited. All rights reserved.	 				 \tn \hline							
			    \rr (C) 2014 Mosby, Inc. All rights reserved.	 \tn \hline	
			    \rr (C) 2014 Society of Photo-Optical Instrumentation Engineers (SPIE) \& Wilkins								 \tn \hline	
			    \rr (C) 2014 by American Society for Reproductive Medicine.																 \tn \hline	
			    \rr (C) 2014 Elsevier Inc. All rights reserved.													 \tn \hline	
			    \rr Copyright (C) 2014, Hydrogen Energy Publications, LLC. Published by Elsevier Ltd. All rights reserved 		 \tn \hline
			    \rr (C) 2014 Optical Society of America																 \tn \hline
			    \rr	(C) 2014 AACR.																 \tn \hline
			    \rr	(C) 2014 S. Karger AG, Basel																 \tn \hline
			    \rr (C) 2014 American Society of Civil Engineers.																 \tn \hline
			    \rr (C) 2014 Wiley Periodicals, Inc. and the American Pharmacists Association.																 \tn \hline
			    \rr (C) 2014 Elsevier Ltd and Techna Group S.r.l. All rights reserved.																 \tn \hline
			    \rr J. surg. oncol		 			\tn \hline						
			    \rr J. med.virol						\tn \hline
			    \rr Polym. compos					\tn \hline
			    \rr Developmental Dynamics			\tn \hline
			    \rr J. exp. zool					 	\tn \hline
			    \rr Proteins 2014					\tn \hline
			    \rr Bioelectromagnetics				\tn \hline
			    \rr J. Cell. Biochemistry			\tn \hline
			    \rr am. J. Hematol					\tn\hline
			    \rr am. J. Primatol					\tn\hline
			\end{tabular}

\label{table:addsch}
   \end{table}

\section{List of Categories.}
\raggedbottom	

\begin{longtable}[h]{| L{0.8cm} | L{8cm} |  R{2cm} | }
	
		\caption{The list of 252 WoS categories with the number of LSC documents assigned to the corresponding category \label{table:categories}}\\ 
		\hline
  	  \multicolumn{1}{|C{0.8cm}|}{\textbf{No.}} & \multicolumn{1}{C{8cm}|}{\textbf{Category}} & \multicolumn{1}{C{2cm}|}{\textbf{Number of Documents}}\\ \hline
\endhead
\hline
\endfoot

1	&	Engineering, Electrical \& Electronic	   			&       174,272  \\ \hline
2	&	Materials Science, Multidisciplinary	   				&       112,912  \\ \hline
3	&	Physics, Applied	                       				&        78,796  \\ \hline
4	&	Chemistry, Physical	 					  			&        58,065  \\ \hline
5	&	Chemistry, Multidisciplinary	           				&        55,907  \\ \hline
6	&	Computer Science, Theory \& Methods	       			&        55,591  \\ \hline
7	&	Multidisciplinary Sciences	               			&        53,140  \\ \hline
8	&	Engineering, Mechanical	                   			&        50,972  \\ \hline
9	&	Optics	             					   			&  		 47,737  \\ \hline
10	&	Biochemistry \& Molecular Biology	       			&	     47,490  \\ \hline
11	&	Computer Science, Information Systems	   			&        45,865  \\ \hline
12	&	Energy \& Fuels	                           			&        44,202  \\ \hline
13	&	Environmental Sciences	                   			&        42,082  \\ \hline
14	&	Computer Science, Artificial Intelligence  			&        41,210  \\ \hline    
15	&	Telecommunications	                       			&        40,550  \\ \hline
16	&	Nanoscience \& Nanotechnology	           			&        35,050  \\ \hline
17	&	Oncology	      						  				&        34,339  \\ \hline
18	&	Mechanics	              				   			&		 33,545  \\ \hline
19	&	Neurosciences	             			   			&  		 32,972  \\ \hline
20	&	Surgery	                                   			& 		 30,805  \\ \hline
21	&	Pharmacology \& Pharmacy	               				& 		 30,713  \\ \hline
22	&	Automation \& Control Systems	           			&  		 29,427  \\ \hline
23	&	Engineering, Chemical	                    			&        29,171  \\ \hline
24	&	Computer Science, Interdisciplinary Applications		&     	 29,153  \\ \hline
25	&	Mathematics, Applied	             					& 		 27,982  \\ \hline
26	&	Physics, Condensed Matter	                        & 		 27,316  \\ \hline
27	&	Biotechnology \& Applied Microbiology	            & 		 26,286  \\ \hline
28	&	Public, Environmental \& Occupational Health	    		&        25,493  \\ \hline
29	&	Mathematics	              							& 		 25,450  \\ \hline
30	&	Geosciences, Multidisciplinary	                    &		 24,644  \\ \hline
31	&	Cell Biology	             							&        23,108  \\ \hline
32	&	Physics, Multidisciplinary	                        &        22,930  \\ \hline
33	&	Astronomy \& Astrophysics	                        &        22,825  \\ \hline
34	&	Economics	                                        &		 	22,338  \\ \hline
35	&	Clinical Neurology	             					&		 22,127	 \\ \hline
36	&	Engineering, Civil	             					&		 22,127	 \\ \hline
37	&	Chemistry, Analytical	        						&		 21,490  \\ \hline
38	&	Plant Sciences	             	 					&	 	 21,321  \\ \hline
39	&	Engineering, Multidisciplinary	 					&		 21,144  \\ \hline
40	&	Radiology, Nuclear Medicine \& Medical Imaging	    &        21,014  \\ \hline
41	&	Food Science \& Technology	                        &        20,414  \\ \hline
42	&	Education \& Educational Research	                &        20,087  \\ \hline
43	&	Medicine, Research \& Experimental	                &        19,744  \\ \hline
44	&	Genetics \& Heredity	                           		&        19,512  \\ \hline
45	&	Computer Science, Hardware \& Architecture	        &        18,489  \\ \hline
46	&	Immunology	                                        &        18,270  \\ \hline
47	&	Polymer Science	                                    &        18,017  \\ \hline
48	&	Chemistry, Organic	                                &        17,941  \\ \hline
49	&	Engineering, Biomedical	              				&		 17,786  \\ \hline
50	&	Microbiology	                     					&		 17,252  \\ \hline
51	&	Computer Science, Software Engineering	            &        17,104  \\ \hline
52  &	Instruments \& Instrumentation	             		&        17,090  \\ \hline
53	&	Physics, Atomic, Molecular \& Chemical	            & 		 17,010  \\ \hline
54	&	Metallurgy \& Metallurgical Engineering	            &        16,898  \\ \hline
55	&	Ecology	             								&		 16,760  \\ \hline
56	&	Cardiac \& Cardiovascular Systems	             	&		 16,369  \\ \hline
57	&	Medicine, General \& Internal	             		&		 16,179  \\ \hline
58	&	Psychiatry	             							&		 16,055  \\ \hline
59	&	Electrochemistry	             						&		 15,663  \\ \hline
60	&	Biochemical Research Methods	             			&		 15,050  \\ \hline
61	&	Endocrinology \& Metabolism	              			&		 14,622  \\ \hline
62	&	Engineering, Environmental	             			&		 14,614  \\ \hline
63	&	Management	             							&		 14,339  \\ \hline
64	&	Chemistry, Applied	             					&		 14,058  \\ \hline
65	&	Water Resources	              						& 		 13,997  \\ \hline
66	&	Thermodynamics	             						&		 13,852  \\ \hline
67	&	Pediatrics	               							&		 13,364  \\ \hline
68	&	Physics, Particles \& Fields	             			&		 13,203  \\ \hline
69	&	Engineering, Manufacturing	             			&		 13,102  \\ \hline
70	&	Biophysics	             							&		 12,630  \\ \hline
71	&	Chemistry, Inorganic \& Nuclear	                 	&		 12,591  \\ \hline
72	&	Infectious Diseases	             					&		 12,521  \\ \hline
73	&	Chemistry, Medicinal	             					&		 12,456  \\ \hline
74	&	Meteorology \& Atmospheric Sciences	             	&		 12,318  \\ \hline
75	&	Construction \& Building Technology	             	&		 12,078  \\ \hline
76	&	Operations Research \& Management Science	        		&     	 11,879  \\ \hline
77	&	Veterinary Sciences	             					&		 11,502  \\ \hline
78	&	Remote Sensing	             						&		 11,388  \\ \hline
79	&	Nuclear Science \& Technology	             		&		 11,359  \\ \hline
80	&	Zoology	             								&		 11,218  \\ \hline
81	&	Social Sciences, Interdisciplinary	             	&		 11,035  \\ \hline
82	&	Gastroenterology \& Hepatology	             		&		 10,943  \\ \hline
83	&	Orthopedics	             							&		 10,538  \\ \hline
84	&	Physics, Mathematical	             				&		 10,426  \\ \hline
85	&	Engineering, Industrial	             				&		 10,362  \\ \hline
86	&	Marine \& Freshwater Biology	             			&		 10,124  \\ \hline
87	&	Mathematics, Interdisciplinary Applications	        &    	 10,072  \\ \hline
88	&	Geochemistry \& Geophysics	             			&		 10,023  \\ \hline
89	&	Biology	                								&		  9,917  \\ \hline
90	&	Obstetrics \& Gynecology	                				&		  9,883  \\ \hline
91	&	Physics, Fluids \& Plasmas	                			&		  9,704  \\ \hline
92	&	Toxicology	                							&		  9,613  \\ \hline
93	&	Statistics \& Probability	                			&		  9,532  \\ \hline
94	&	Nutrition \& Dietetics	                     		&		  9,416  \\ \hline
95	&	Business	                								&		  9,394  \\ \hline
96	&	Imaging Science \& Photographic Technology	        &     	  9,353  \\ \hline
97	&	Hematology	               							&		  9,096  \\ \hline
98	&	Physiology	                							&		  9,009  \\ \hline
99	&	Peripheral Vascular Disease	                			&		  8,700  \\ \hline
100	&	Agronomy	                    							&		  8,651  \\ \hline
101	&	Dentistry, Oral Surgery \& Medicine	                &		  8,502  \\ \hline
102	&	Robotics	                								&		  8,491  \\ \hline
103	&	Transportation Science \& Technology	           		&    	  8,411  \\ \hline
104	&	Sport Sciences	                					 	&		  8,368  \\ \hline
105	&	Psychology, Multidisciplinary	                		&		  8,332  \\ \hline
106	&	Urology \& Nephrology	                				&		  8,264  \\ \hline
107	&	Materials Science, Biomaterials	                		&		  8,040  \\ \hline
108	&	Mathematical \& Computational Biology	            &         8,015  \\ \hline
109 &	Health Care Sciences \& Services	                		&         7,999  \\ \hline
110	&	Physics, Nuclear	                                		&         7,876  \\ \hline
111	&	Ophthalmology	                                    &         7,830  \\ \hline
112	&	Environmental Studies	                            &         7,811  \\ \hline
113	&	Rehabilitation	                                    &         7,791  \\ \hline
114	&	Respiratory System	                                &         7,666  \\ \hline
115	&	Oceanography	                						&		  7,417  \\ \hline
116	&	Spectroscopy	                     				&         7,388  \\ \hline
117	&	Materials Science, Coatings \& Films	            &         7,226  \\ \hline
118	&	Pathology	                						&         7,217  \\ \hline
119	&	Business, Finance	                				&         7,214  \\ \hline
120	&	Psychology	                						&         6,989  \\ \hline
121	&	Acoustics	                						&         6,935  \\ \hline
122	&	Crystallography	               					&         6,932  \\ \hline
123	&	Psychology, Clinical	                           	&		  6,860  \\ \hline
124	&	Geography, Physical	                             &		  6,806  \\ \hline
125	&	Psychology, Experimental	                			&		  6,784  \\ \hline
126	&	Nursing	                							&		  6,637  \\ \hline
127	&	Green \& Sustainable Science \& Technology	    &         6,412  \\ \hline
128	&	Agriculture, Multidisciplinary	                	&	      6,406  \\ \hline
129	&	Education, Scientific Disciplines	            &		  6,308  \\ \hline
130	&	Virology	                							&		  6,270  \\ \hline
131	&	Materials Science, Ceramics	               		&		  6,222  \\ \hline
132	&	Agriculture, Dairy \& Animal Science	            & 		  6,163  \\ \hline
133	&	Behavioral Sciences	                				&	      5,922  \\ \hline
134	&	Linguistics	                						&	      5,921  \\ \hline
135	&	Dermatology	               						&	      5,793  \\ \hline
136	&	Evolutionary Biology	                				&	      5,742  \\ \hline
137	&	Entomology	                						&	      5,704  \\ \hline
138	&	Parasitology	               						&		  5,683  \\ \hline
139	&	Horticulture	                						&		  5,338  \\ \hline
140	&	Health Policy \& Services	                		&		  5,318  \\ \hline
141	&	Language \& Linguistics	               				&		  5,174  \\ \hline
142	&	Political Science	                				&		  5,106  \\ \hline
143	&	Soil Science	               						&	      4,800  \\ \hline
144	&	Otorhinolaryngology	              					&	      4,797  \\ \hline
145	&	Geriatrics \& Gerontology	            			&	      4,742  \\ \hline
146	&	Sociology	                						&		  4,725  \\ \hline
147	&	Biodiversity Conservation	              			&		  4,705  \\ \hline
148	&	Fisheries	                						&		  4,702  \\ \hline
149	&	Engineering, Geological	                			&		  4,573  \\ \hline
150	&	Information Science \& Library Science	  			&         4,565  \\ \hline
151	&	Forestry	                						&		  4,472  \\ \hline
152	&	Engineering, Aerospace	               				&		  4,435  \\ \hline
153	&	Psychology, Developmental	               			&		  4,390  \\ \hline
154	&	Materials Science, Composites	               		&		  4,277  \\ \hline
155	&	Planning \& Development	               				&		  4,115  \\ \hline
156	&	Transplantation	                					&		  4,105  \\ \hline
157	&	Transportation	               						&		  4,035  \\ \hline
158	&	Medical Informatics	               					&		  3,991  \\ \hline
159	&	Reproductive Biology	                			&		  3,984  \\ \hline
160	&	Critical Care Medicine	               				&		  3,982  \\ \hline
161	&	Rheumatology	             						&	      3,942  \\ \hline
162	&	Geography	                						&		  3,908  \\ \hline
163	&	Materials Science, Characterization \& Testing	    &         3,878  \\ \hline
164	&	Agricultural Engineering	               			&		  3,727  \\ \hline
165	&	Tropical Medicine	              					&		  3,696  \\ \hline
166	&	Philosophy	               							&		  3,657  \\ \hline
167	&	Computer Science, Cybernetics	                	&	      3,652  \\ \hline
168	&	Developmental Biology	                			&	      3,593  \\ \hline
169	&	Law	               									&		  3,574  \\ \hline
170	&	Psychology, Social	              					&	      3,548  \\ \hline
171	&	Psychology, Applied	                				&		  3,523  \\ \hline 
172	&	Social Sciences, Mathematical Methods	            &         3,496  \\ \hline
173	&	History	               								&		  3,487  \\ \hline
174	&	Integrative \& Complementary Medicine	            &         3,453  \\ \hline
175	&	Substance Abuse	               						&	      3,433  \\ \hline
176	&	Communication	                					&		  3,200  \\ \hline
177	&	Anthropology	                					&		  3,149  \\ \hline
178	&	Social Sciences, Biomedical	              			&		  3,003  \\ \hline
179	&	Hospitality, Leisure, Sport \& Tourism	            &   	  2,998  \\ \hline
180	&	Anesthesiology	                					&		  2,943  \\ \hline
181	&	International Relations	                			&		  2,941  \\ \hline
182	&	Neuroimaging	               						&		  2,702  \\ \hline
183	&	Mining \& Mineral Processing	               		&		  2,687  \\ \hline
184	&	Emergency Medicine	                				&		  2,627  \\ \hline
185	&	Medical Laboratory Technology	               		&		  2,598  \\ \hline
186	&	Humanities, Multidisciplinary	                	&		  2,559  \\ \hline
187	&	Mineralogy	               							&	      2,550  \\ \hline
188	&	Materials Science, Textiles	                		&		  2,548  \\ \hline
189	&	Gerontology	                						&		  2,531  \\ \hline
190	&	Paleontology	               						&	      2,503  \\ \hline
191	&	Cell \& Tissue Engineering	                		&		  2,455  \\ \hline
192	&	Engineering, Ocean	                				&		  2,352  \\ \hline
193	&	Religion	               							&		  2,335  \\ \hline
194	&	Urban Studies	               						&		  2,309  \\ \hline
195	&	Family Studies	               						&		  2,229  \\ \hline
196	&	Public Administration	               				&		  2,204  \\ \hline
197	&	History \& Philosophy Of Science	               	&		  2,199  \\ \hline
198	&	Geology	                							&		  2,153  \\ \hline
199	&	Archaeology	               							&		  2,118  \\ \hline
200	&	Social Work	                						&		  2,114  \\ \hline
201	&	Psychology, Educational	                			&		  2,112  \\ \hline
202	&	Engineering, Marine	                				&		  2,110  \\ \hline
203	&	Audiology \& Speech-Language Pathology	            &         2,052  \\ \hline
204	&	Area Studies	                					&		  2,046  \\ \hline
205	&	Criminology \& Penology	                			&		  2,015  \\ \hline
206	&	Materials Science, Paper \& Wood	                &		  1,963  \\ \hline
207	&	Limnology	               							&		  1,941  \\ \hline
208	&	Engineering, Petroleum	               				&         1,930  \\ \hline
209	&	Ethics	                							&		  1,928  \\ \hline
210	&	Anatomy \& Morphology	                			&		  1,889  \\ \hline
211	&	Mycology	                						&		  1,829  \\ \hline
212	&	Logic	               								&	      1,786  \\ \hline
213	&	Allergy	                							&		  1,765  \\ \hline
214	&	Medicine, Legal	                					&		  1,711  \\ \hline
215	&	Education, Special	                				&		  1,666  \\ \hline
216	&	Literature	                						&		  1,608  \\ \hline
217	&	Psychology, Biological	                			&		  1,527  \\ \hline
218	&	Ergonomics	                						&		  1,431  \\ \hline
219	&	Architecture	                					&		  1,376  \\ \hline
220	&	Women's Studies	                					&		  1,341  \\ \hline
221	&	Microscopy	                						&		  1,319  \\ \hline
222	&	Social Issues	                					&		  1,296  \\ \hline
223	&	Primary Health Care	                				&		  1,269  \\ \hline
224	&	Ornithology	                						&		  1,008  \\ \hline
225	&	Demography	                   						&		    948  \\ \hline
226	&	Cultural Studies	                   				&		    945  \\ \hline
227	&	Music	                   							&			888  \\ \hline
228	&	Agricultural Economics \& Policy	                &  			880  \\ \hline
229	&	History Of Social Sciences	                   		&			879  \\ \hline
230	&	Industrial Relations \& Labor	                   	&			879  \\ \hline
231	&	Asian Studies	                   					&			877  \\ \hline
232	&	Art	                   								&			725  \\ \hline
233	&	Ethnic Studies	                   					&			675  \\ \hline
234	&	Medical Ethics	                   					&			674  \\ \hline
235	&	Psychology, Mathematical	                   		&			538  \\ \hline
236	&	Literary Theory \& Criticism	                   	&			498  \\ \hline
237	&	Medieval \& Renaissance Studies	                  	&			485  \\ \hline
238	&	Film, Radio, Television	                   			&			398  \\ \hline
239	&	Andrology	                   						&			391  \\ \hline
240	&	Psychology, Psychoanalysis	                   		&			345  \\ \hline
241	&	Classics	                   						&			325  \\ \hline
242	&	Theater	                   							&			300  \\ \hline
243	&	Literature, Romance	                   				&			269  \\ \hline
244	&	Literature, British Isles	                   		&			220  \\ \hline
245	&	Folklore	                   						&			134  \\ \hline
246	&	Literature, German, Dutch, Scandinavian	            &     		128  \\ \hline
247 &	Literature, American	                      		&			 75  \\ \hline
248 &	Dance	                      						&			 74  \\ \hline
249 &	Literature, African, Australian, Canadian	        &            59  \\ \hline
250 &	Poetry	                      						&			 42  \\ \hline
251 &	Literary Reviews	                      			&			 35  \\ \hline
252 &	Literature, Slavic	                                &            35  \\ \hline
 
\label{table:cate_tabl} 
 
	 \end{longtable}

\section{List of Research Areas.}
\raggedbottom	

\begin{longtable}[h]{| L{1cm}| L{8cm} |  R{2cm} | }
	
		\caption{The list of 151 WoS research areas with the number of LSC documents assigned to the corresponding research area \label{table:ra}}\\ 
		\hline
  	  \multicolumn{1}{|C{1cm}|}{\textbf{No.}} & \multicolumn{1}{C{8cm}|}{\textbf{Research Area}} & \multicolumn{1}{C{2cm}|}{\textbf{Number of Documents}}\\ \hline
\endhead
\hline
\endfoot

1	&	Engineering	   										&       328,136   \\ \hline
2	&	Chemistry	   										&       162,934   \\ \hline
3	&	Physics        										&       158,438   \\ \hline
4	&	Computer Science									&       142,633   \\ \hline
5	&	Materials Science	               					&       141,754   \\ \hline
6	&	Science \& Technology - Other Topics	            &        96,388   \\ \hline
7	&	Environmental Sciences \& Ecology	             	&  		 60,657   \\ \hline
8	&	Biochemistry \& Molecular Biology	       			&	     60,027   \\ \hline
9	&	Mathematics	   										&        59,525   \\ \hline
10	&	Neurosciences \& Neurology	                        &        48,680   \\ \hline
11	&	Optics	                   							&        47,737   \\ \hline
12	&	Energy \& Fuels  									&        44,202   \\ \hline    
13	&	Business \& Economics	                       		&        40,743   \\ \hline
14	&	Telecommunications	           						&        40,550   \\ \hline
15	&	Pharmacology \& Pharmacy	      					&        38,837   \\ \hline
16	&	Psychology	              				   			&		 36,282   \\ \hline
17	&	Oncology	             			   				&  		 34,339   \\ \hline
18	&	Mechanics	                                   		& 		 33,545   \\ \hline
19	&	Agriculture	               							& 		 31,191   \\ \hline
20	&	Surgery												& 		 30,805   \\ \hline
21	&	Automation \& Control Systems	           			&  		 29,427   \\ \hline
22	&	Geology	                    						&        26,632   \\ \hline
23	&	Biotechnology \& Applied Microbiology				&     	 26,286   \\ \hline
24	&	Education \& Educational Research	             	& 		 25,924   \\ \hline
25	&	Public, Environmental \& Occupational Health	    &        25,493   \\ \hline
26	&	Cell Biology	             						&        24,145   \\ \hline
27	&	Cardiovascular System \& Cardiology	                &        23,396   \\ \hline
28	&	Astronomy \& Astrophysics	                        &        22,825   \\ \hline
29	&	Plant Sciences	                                    &		 21,321   \\ \hline
30	&	Radiology, Nuclear Medicine \& Medical Imaging	    &		 21,014   \\ \hline
31	&	Food Science \& Technology	             			&		 20,414   \\ \hline
32	&	General \& Internal Medicine	        			&		 20,409   \\ \hline
33	&	Research \& Experimental Medicine	             	&	 	 19,744   \\ \hline
34	&	Genetics \& Heredity	 							&		 19,512   \\ \hline
35	&	Immunology	    									&        18,270   \\ \hline
36	&	Polymer Science	                        			&        18,017   \\ \hline
37	&	Microbiology	                					&        17,252   \\ \hline
38	&	Instruments \& Instrumentation	               		&        17,090   \\ \hline
39	&	Metallurgy \& Metallurgical Engineering	            &        16,898   \\ \hline
40	&	Social Sciences - Other Topics	       				&        16,666   \\ \hline
41	&	Psychiatry	                                        &        16,055   \\ \hline
42	&	Electrochemistry	                                &        15,663   \\ \hline
43	&	Endocrinology \& Metabolism	                        &        15,013   \\ \hline
44	&	Water Resources	              						&		 13,997   \\ \hline
45	&	Thermodynamics	                     				&		 13,852   \\ \hline
46	&	Pediatrics	            							&        13,365   \\ \hline
47	&	Biophysics	             							&        12,630   \\ \hline
48	&	Infectious Diseases	            					& 		 12,521   \\ \hline
49	&	Meteorology \& Atmospheric Sciences	            	&        12,318   \\ \hline
50	&	Zoology	             								&		 12,200   \\ \hline
51	&	Construction \& Building Technology	             	&		 12,078   \\ \hline
52	&	Operations Research \& Management Science	        &		 11,879   \\ \hline
53	&	Marine \& Freshwater Biology	             		&		 11,562   \\ \hline
54	&	Veterinary Sciences	             					&		 11,502   \\ \hline
55	&	Remote Sensing	             						&		 11,388   \\ \hline
56	&	Nuclear Science \& Technology	              		&		 11,359   \\ \hline
57	&	Gastroenterology \& Hepatology	             		&		 10,943   \\ \hline
58	&	Orthopedics	             							&		 10,538   \\ \hline
59	&	Transportation	             						&		 10,280   \\ \hline
60	&	Health Care Sciences \& Services              		& 		 10,243   \\ \hline
61	&	Geochemistry \& Geophysics	             			&		 10,023   \\ \hline
62	&	Life Sciences \& Biomedicine - Other Topics	        &		  9,917   \\ \hline
63	&	Obstetrics \& Gynecology	             			&		  9,883   \\ \hline
64	&	Toxicology	             							&		  9,613   \\ \hline
65	&	Nutrition \& Dietetics	             				&		  9,416   \\ \hline
66	&	Imaging Science \& Photographic Technology	        &		  9,353   \\ \hline
67	&	Hematology	             							&		  9,096   \\ \hline
68	&	Physiology	             							&		  9,009   \\ \hline
69	&	Dentistry, Oral Surgery \& Medicine	             	&		  8,502   \\ \hline
70	&	Government \& Law             						&		  8,492   \\ \hline
71	&	Robotics	        								&     	  8,491   \\ \hline
72	&	Sport Sciences	             						&		  8,368   \\ \hline
73	&	Urology \& Nephrology	             				&		  8,264   \\ \hline
74	&	Mathematical \& Computational Biology	            &		  8,015   \\ \hline
75	&	Ophthalmology	             						&		  7,830   \\ \hline
76	&	Rehabilitation	             						&		  7,791   \\ \hline
77	&	Respiratory System	             					&		  7,666   \\ \hline
78	&	Oceanography	             						&		  7,417   \\ \hline
79	&	Spectroscopy	             						&		  7,388   \\ \hline
80	&	Pathology	             							&		  7,217   \\ \hline
81	&	Linguistics	             							&		  7,077   \\ \hline
82	&	Acoustics	        								&    	  6,935   \\ \hline
83	&	Crystallography	             						&		  6,932   \\ \hline
84	&	Physical Geography	                				&		  6,806   \\ \hline
85	&	Nursing	                							&		  6,637   \\ \hline
86	&	Virology	                						&		  6,270   \\ \hline
87	&	Public Administration	                			&		  6,120   \\ \hline
88	&	Behavioral Sciences	                				&		  5,922   \\ \hline
89	&	Dermatology	               							&	      5,793   \\ \hline
90	&	Evolutionary Biology	                			&	      5,742   \\ \hline
91	&	Entomology	                						&	      5,704   \\ \hline
92	&	Parasitology	               						&		  5,683   \\ \hline
93	&	Geriatrics \& Gerontology	               			&	      5,505   \\ \hline
94	&	Otorhinolaryngology	              					&	      4,797   \\ \hline
95	&	Sociology	                						&		  4,725   \\ \hline
96	&	Biodiversity \& Conservation	              		&		  4,705   \\ \hline
97	&	Fisheries	                						&		  4,702   \\ \hline
98	&	Information Science \& Library Science	  			&         4,565   \\ \hline
99	&	Forestry	                						&		  4,472   \\ \hline
100	&	Transplantation	                					&		  4,105   \\ \hline
101	&	Medical Informatics	               					&		  3,991   \\ \hline
102	&	Reproductive Biology	                			&		  3,984   \\ \hline
103	&	Rheumatology	             						&	      3,942   \\ \hline
104	&	Geography	                						&		  3,908   \\ \hline
105	&	Tropical Medicine	              					&		  3,696   \\ \hline
106	&	Philosophy	               							&		  3,657   \\ \hline
107	&	Developmental Biology	                			&	      3,593   \\ \hline
108	&	Mathematical Methods In Social Sciences	            &         3,496   \\ \hline
109	&	History	               								&		  3,487   \\ \hline
110	&	Integrative \& Complementary Medicine	            &         3,453   \\ \hline
111	&	Substance Abuse	               						&	      3,433   \\ \hline
112	&	Communication	                					&		  3,200   \\ \hline
113	&	Arts \& Humanities - Other Topics					&		  3,178   \\ \hline
114	&	Anthropology	                					&		  3,149   \\ \hline
115	&	Biomedical Social Sciences	              			&		  3,003   \\ \hline
116	&	Anesthesiology	                					&		  2,943   \\ \hline
117	&	International Relations	                			&		  2,941   \\ \hline
118	&	Literature		               						&		  2,735   \\ \hline
119	&	Mining \& Mineral Processing	               		&		  2,687   \\ \hline
120	&	Emergency Medicine	                				&		  2,627   \\ \hline
121	&	Medical Laboratory Technology	               		&		  2,598   \\ \hline
122	&	Mineralogy	               							&	      2,550   \\ \hline
123	&	Paleontology	               						&	      2,503   \\ \hline
124	&	Religion	               							&		  2,335   \\ \hline
125	&	Urban Studies	               					&		  2,309   \\ \hline
126	&	Family Studies	               					&		  2,229   \\ \hline
127	&	History \& Philosophy Of Science	               	&		  2,199   \\ \hline
128	&	Archaeology	               						&		  2,118   \\ \hline
129	&	Social Work	                						&		  2,114   \\ \hline
130	&	Audiology \& Speech-Language Pathology	        &         2,052   \\ \hline
131	&	Area Studies	                						&		  2,046   \\ \hline
132	&	Criminology \& Penology	                			&		  2,015   \\ \hline
133	&	Anatomy \& Morphology	                			&		  1,889   \\ \hline
134	&	Mycology	                							&		  1,829   \\ \hline
135	&	Allergy	                							&		  1,765   \\ \hline
136	&	Legal Medicine	                					&		  1,711   \\ \hline
137	&	Architecture	                						&		  1,376   \\ \hline
138	&	Women's Studies	                					&		  1,341   \\ \hline
139	&	Microscopy	                						&		  1,319   \\ \hline
140	&	Social Issues	                					&		  1,296   \\ \hline
141	&	Demography	                   						&		    948   \\ \hline
142	&	Cultural Studies	                   				&		    945   \\ \hline
143	&	Music	                   							&			888   \\ \hline
144	&	Asian Studies	                   					&			877   \\ \hline
145	&	Art	                   								&			725   \\ \hline
146	&	Ethnic Studies	                   					&			675   \\ \hline
147	&	Medical Ethics	                   					&			674   \\ \hline
148	&	Film, Radio, Television	                   			&			398   \\ \hline
149	&	Classics	                   						&			325   \\ \hline
150	&	Theater	                   							&			300   \\ \hline
151	&	Dance	                      						&			 74   \\ \hline
	
		\end{longtable}	

\section{Word Clouds and Histograms for Categories.} \label{wordclouds}
Word clouds presenting the top 100 words ordered by their RIGs and histograms of RIGs for the first 10 words in the word clouds for 252 categories in LSC.

\begin{figure}[p]
\centering
 \includegraphics[width=.8\linewidth]{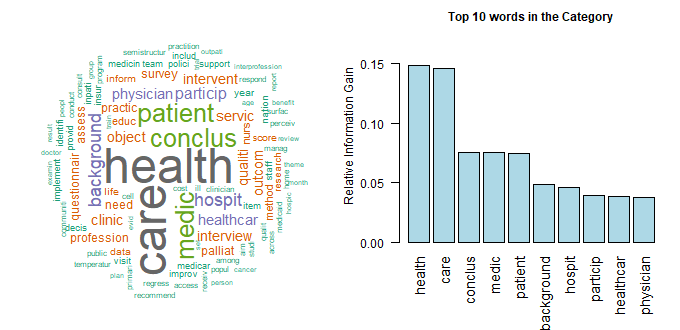}
   \caption{Health Care Sciences \& Services}
  \label{fig:RIG1_105}
\end{figure}

\begin{figure}[p]
\centering
 \includegraphics[width=.8\linewidth]{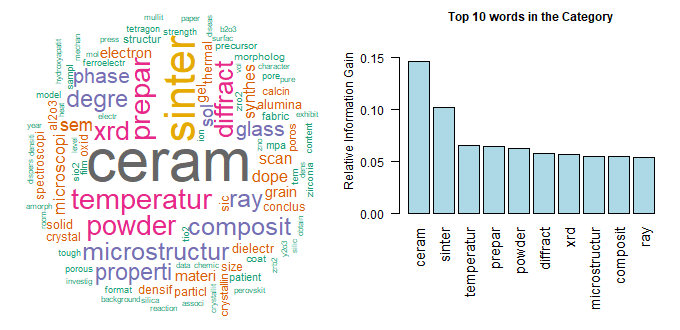}
   \caption{Material Science, Ceramics}
  \label{fig:RIG1_139}
\end{figure}

\begin{figure}[p]
\centering
 \includegraphics[width=.8\linewidth]{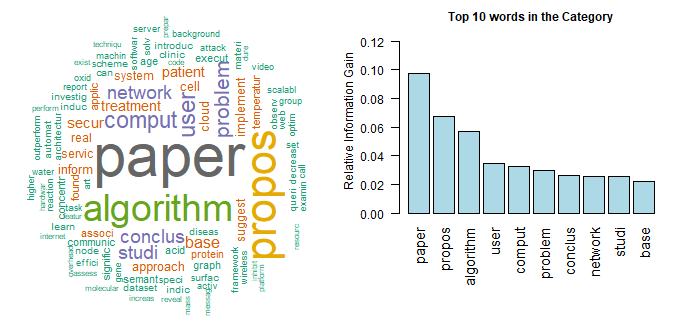}
   \caption{Computer Science, Theory \& Methods}
  \label{fig:RIG1_48}
\end{figure}

\begin{figure}[p]
\centering
 \includegraphics[width=.8\linewidth]{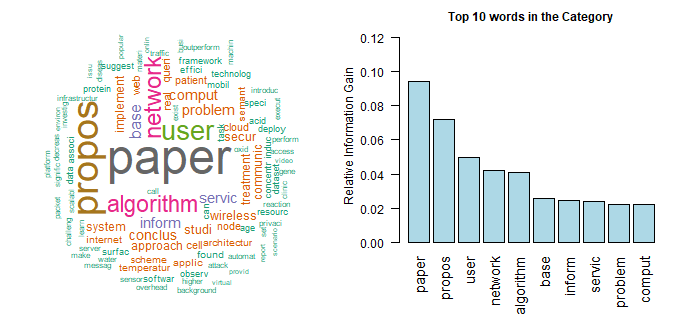}
   \caption{Computer Science, Information Systems}
  \label{fig:RIG1_45}
\end{figure}

\begin{figure}[p]
\centering
 \includegraphics[width=.8\linewidth]{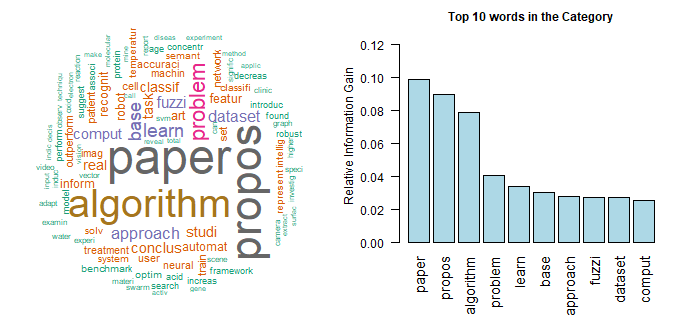}
   \caption{Computer Science, Artificial Intelligence}
  \label{fig:RIG1_42}
\end{figure}

\begin{figure}[p]
\centering
 \includegraphics[width=.8\linewidth]{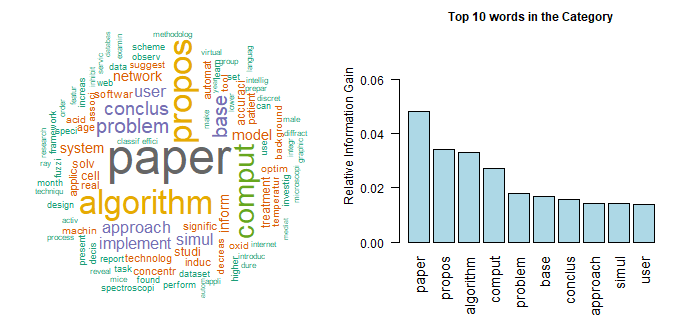}
   \caption{Computer Science, Interdisciplinary Applications}
  \label{fig:RIG1_46}
\end{figure}

\begin{figure}[p]
\centering
 \includegraphics[width=.8\linewidth]{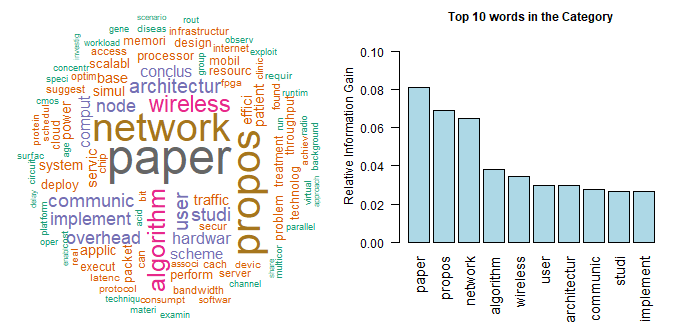}
   \caption{Computer Science, Hardware \& Architecture}
  \label{fig:RIG1_44}
\end{figure}

\begin{figure}[p]
\centering
 \includegraphics[width=.8\linewidth]{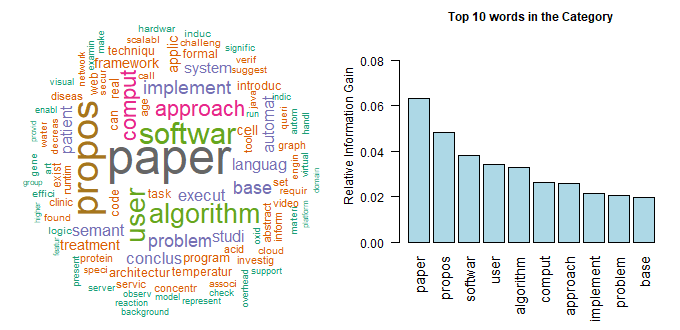}
   \caption{Computer Science, Software Engineering}
  \label{fig:RIG1_47}
\end{figure}

\begin{figure}[p]
\centering
 \includegraphics[width=.8\linewidth]{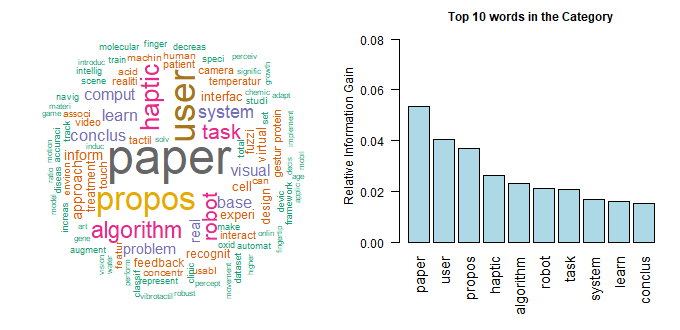}
   \caption{Computer Science, Cybernetics}
  \label{fig:RIG1_43}
\end{figure}

\section{The most Informative 100 words in categories with their RIGs.} \label{InfCat}

\clearpage 

\newpage 

\thispagestyle{plain} 

\vspace*{-1.6in}

\setlength{\LTleft}{-20cm plus -1fill}

\setlength{\LTright}{\LTleft} 

\begin{longtable}[h]{| C{.6cm}| L{5cm} |  R{1.6cm}  |  R{.3cm} |  C{.6cm}| L{5cm} |  R{1.6cm} | }

\caption{Acoustics}\\ 

\hhline{|-|-|-|~|-|-|-|}

\multicolumn{1}{|C{.6cm}|}{\textbf{No.}} & \multicolumn{1}{C{5cm}|}{\textbf{Word}} & \multicolumn{1}{C{1.6cm}|}{\textbf{RIG}} &   &\multicolumn{1}{C{.6cm}|}{\textbf{No.}} & \multicolumn{1}{C{5cm}|}{\textbf{Word}} & \multicolumn{1}{C{1.6cm}|}{\textbf{RIG}} \\ \hhline{|-|-|-|~|-|-|-|} 

\endhead 

\hhline{|-|-|-|~|-|-|-|}

\endfoot
 1 & acoust & $ 9.4\times10^{-2} $ &   & 51 & finit & $ 6.6\times10^{-3} $ \\ \hhline{|-|-|-|~|-|-|-|} 
 
 2 & ultrasound & $ 6.2\times10^{-2} $ &   & 52 & veloc & $ 6.5\times10^{-3} $ \\ \hhline{|-|-|-|~|-|-|-|} 
 
 3 & sound & $ 4.1\times10^{-2} $ &   & 53 & spars & $ 6.5\times10^{-3} $ \\ \hhline{|-|-|-|~|-|-|-|} 
 
 4 & speech & $ 4.1\times10^{-2} $ &   & 54 & music & $ 6.1\times10^{-3} $ \\ \hhline{|-|-|-|~|-|-|-|} 
 
 5 & frequenc & $ 4\times10^{-2} $ &   & 55 & gene & $ 6.1\times10^{-3} $ \\ \hhline{|-|-|-|~|-|-|-|} 
 
 6 & nois & $ 3.3\times10^{-2} $ &   & 56 & linear & $ 6.1\times10^{-3} $ \\ \hhline{|-|-|-|~|-|-|-|} 
 
 7 & ultrason & $ 2.7\times10^{-2} $ &   & 57 & spectral & $ 6.1\times10^{-3} $ \\ \hhline{|-|-|-|~|-|-|-|} 
 
 8 & wave & $ 2.4\times10^{-2} $ &   & 58 & noisi & $ 6\times10^{-3} $ \\
\end{longtable}
 \newpage 
 
 \clearpage 

\newpage 

\thispagestyle{plain} 

\vspace*{-1.6in}

\setlength{\LTleft}{-20cm plus -1fill}

\setlength{\LTright}{\LTleft} 

\begin{longtable}[h]{| C{.6cm}| L{5cm} |  R{1.6cm}  |  R{.3cm} |  C{.6cm}| L{5cm} |  R{1.6cm} | }

\caption{Agricultural Economics and Policy}\\ 

\hhline{|-|-|-|~|-|-|-|}

\multicolumn{1}{|C{.6cm}|}{\textbf{No.}} & \multicolumn{1}{C{5cm}|}{\textbf{Word}} & \multicolumn{1}{C{1.6cm}|}{\textbf{RIG}} &   &\multicolumn{1}{C{.6cm}|}{\textbf{No.}} & \multicolumn{1}{C{5cm}|}{\textbf{Word}} & \multicolumn{1}{C{1.6cm}|}{\textbf{RIG}} \\ \hhline{|-|-|-|~|-|-|-|} 

\endhead 

\hhline{|-|-|-|~|-|-|-|}

\endfoot
 1 & agricultur & $ 1.1\times10^{-1} $ &   & 51 & consumpt & $ 9.9\times10^{-3} $ \\ \hhline{|-|-|-|~|-|-|-|} 
 
 2 & farmer & $ 8.7\times10^{-2} $ &   & 52 & payment & $ 9.9\times10^{-3} $ \\ \hhline{|-|-|-|~|-|-|-|} 
 
 3 & market & $ 8.6\times10^{-2} $ &   & 53 & agribusi & $ 9.8\times10^{-3} $ \\ \hhline{|-|-|-|~|-|-|-|} 
 
 4 & farm & $ 8.3\times10^{-2} $ &   & 54 & economi & $ 9.6\times10^{-3} $ \\ \hhline{|-|-|-|~|-|-|-|} 
 
 5 & price & $ 8.1\times10^{-2} $ &   & 55 & wtp & $ 9.6\times10^{-3} $ \\ \hhline{|-|-|-|~|-|-|-|} 
 
 6 & food & $ 7.2\times10^{-2} $ &   & 56 & method & $ 9.4\times10^{-3} $ \\ \hhline{|-|-|-|~|-|-|-|} 
 
 7 & polici & $ 6\times10^{-2} $ &   & 57 & choic & $ 9.3\times10^{-3} $ \\ \hhline{|-|-|-|~|-|-|-|} 
 
 8 & product & $ 4.8\times10^{-2} $ &   & 58 & competit & $ 9.3\times10^{-3} $ \\
\end{longtable}
 \newpage 
 
 \clearpage 

\newpage 

\thispagestyle{plain} 

\vspace*{-1.6in}

\setlength{\LTleft}{-20cm plus -1fill}

\setlength{\LTright}{\LTleft} 

\begin{longtable}[h]{| C{.6cm}| L{5cm} |  R{1.6cm}  |  R{.3cm} |  C{.6cm}| L{5cm} |  R{1.6cm} | }

\caption{Agricultural Engineering}\\ 

\hhline{|-|-|-|~|-|-|-|}

\multicolumn{1}{|C{.6cm}|}{\textbf{No.}} & \multicolumn{1}{C{5cm}|}{\textbf{Word}} & \multicolumn{1}{C{1.6cm}|}{\textbf{RIG}} &   &\multicolumn{1}{C{.6cm}|}{\textbf{No.}} & \multicolumn{1}{C{5cm}|}{\textbf{Word}} & \multicolumn{1}{C{1.6cm}|}{\textbf{RIG}} \\ \hhline{|-|-|-|~|-|-|-|} 

\endhead 

\hhline{|-|-|-|~|-|-|-|}

\endfoot
 1 & biomass & $ 6.8\times10^{-2} $ &   & 51 & soil & $ 1.2\times10^{-2} $ \\ \hhline{|-|-|-|~|-|-|-|} 
 
 2 & product & $ 6\times10^{-2} $ &   & 52 & fatti & $ 1.1\times10^{-2} $ \\ \hhline{|-|-|-|~|-|-|-|} 
 
 3 & yield & $ 4.1\times10^{-2} $ &   & 53 & saccharif & $ 1.1\times10^{-2} $ \\ \hhline{|-|-|-|~|-|-|-|} 
 
 4 & anaerob & $ 3.1\times10^{-2} $ &   & 54 & solid & $ 1.1\times10^{-2} $ \\ \hhline{|-|-|-|~|-|-|-|} 
 
 5 & content & $ 2.9\times10^{-2} $ &   & 55 & corn & $ 1.1\times10^{-2} $ \\ \hhline{|-|-|-|~|-|-|-|} 
 
 6 & ferment & $ 2.9\times10^{-2} $ &   & 56 & fuel & $ 1.1\times10^{-2} $ \\ \hhline{|-|-|-|~|-|-|-|} 
 
 7 & crop & $ 2.9\times10^{-2} $ &   & 57 & enzymat & $ 1.1\times10^{-2} $ \\ \hhline{|-|-|-|~|-|-|-|} 
 
 8 & sludg & $ 2.6\times10^{-2} $ &   & 58 & manur & $ 1\times10^{-2} $ \\
\end{longtable}
 \newpage 
 
 \clearpage 

\newpage 

\thispagestyle{plain} 

\vspace*{-1.6in}

\setlength{\LTleft}{-20cm plus -1fill}

\setlength{\LTright}{\LTleft} 

\begin{longtable}[h]{| C{.6cm}| L{5cm} |  R{1.6cm}  |  R{.3cm} |  C{.6cm}| L{5cm} |  R{1.6cm} | }

\caption{Agriculture, Dairy and Animal Science}\\ 

\hhline{|-|-|-|~|-|-|-|}

\multicolumn{1}{|C{.6cm}|}{\textbf{No.}} & \multicolumn{1}{C{5cm}|}{\textbf{Word}} & \multicolumn{1}{C{1.6cm}|}{\textbf{RIG}} &   &\multicolumn{1}{C{.6cm}|}{\textbf{No.}} & \multicolumn{1}{C{5cm}|}{\textbf{Word}} & \multicolumn{1}{C{1.6cm}|}{\textbf{RIG}} \\ \hhline{|-|-|-|~|-|-|-|} 

\endhead 

\hhline{|-|-|-|~|-|-|-|}

\endfoot
 1 & feed & $ 1.1\times10^{-1} $ &   & 51 & silag & $ 2.5\times10^{-2} $ \\ \hhline{|-|-|-|~|-|-|-|} 
 
 2 & diet & $ 1.1\times10^{-1} $ &   & 52 & libitum & $ 2.4\times10^{-2} $ \\ \hhline{|-|-|-|~|-|-|-|} 
 
 3 & cow & $ 1\times10^{-1} $ &   & 53 & dmi & $ 2.3\times10^{-2} $ \\ \hhline{|-|-|-|~|-|-|-|} 
 
 4 & milk & $ 9.4\times10^{-2} $ &   & 54 & heifer & $ 2.3\times10^{-2} $ \\ \hhline{|-|-|-|~|-|-|-|} 
 
 5 & fed & $ 9.1\times10^{-2} $ &   & 55 & chick & $ 2.3\times10^{-2} $ \\ \hhline{|-|-|-|~|-|-|-|} 
 
 6 & dairi & $ 8.9\times10^{-2} $ &   & 56 & matter & $ 2.3\times10^{-2} $ \\ \hhline{|-|-|-|~|-|-|-|} 
 
 7 & anim & $ 7.1\times10^{-2} $ &   & 57 & paper & $ 2.2\times10^{-2} $ \\ \hhline{|-|-|-|~|-|-|-|} 
 
 8 & intak & $ 6.9\times10^{-2} $ &   & 58 & period & $ 2.2\times10^{-2} $ \\
\end{longtable}
 \newpage 
 
 \clearpage 

\newpage 

\thispagestyle{plain} 

\vspace*{-1.6in}

\setlength{\LTleft}{-20cm plus -1fill}

\setlength{\LTright}{\LTleft} 

\begin{longtable}[h]{| C{.6cm}| L{5cm} |  R{1.6cm}  |  R{.3cm} |  C{.6cm}| L{5cm} |  R{1.6cm} | }

\caption{Agriculture, Multidisciplinary}\\ 

\hhline{|-|-|-|~|-|-|-|}

\multicolumn{1}{|C{.6cm}|}{\textbf{No.}} & \multicolumn{1}{C{5cm}|}{\textbf{Word}} & \multicolumn{1}{C{1.6cm}|}{\textbf{RIG}} &   &\multicolumn{1}{C{.6cm}|}{\textbf{No.}} & \multicolumn{1}{C{5cm}|}{\textbf{Word}} & \multicolumn{1}{C{1.6cm}|}{\textbf{RIG}} \\ \hhline{|-|-|-|~|-|-|-|} 

\endhead 

\hhline{|-|-|-|~|-|-|-|}

\endfoot
 1 & crop & $ 6\times10^{-2} $ &   & 51 & chromatographi & $ 8.4\times10^{-3} $ \\ \hhline{|-|-|-|~|-|-|-|} 
 
 2 & plant & $ 4.4\times10^{-2} $ &   & 52 & agronom & $ 8.4\times10^{-3} $ \\ \hhline{|-|-|-|~|-|-|-|} 
 
 3 & agricultur & $ 4\times10^{-2} $ &   & 53 & paper & $ 8.3\times10^{-3} $ \\ \hhline{|-|-|-|~|-|-|-|} 
 
 4 & soil & $ 3.5\times10^{-2} $ &   & 54 & day & $ 8.3\times10^{-3} $ \\ \hhline{|-|-|-|~|-|-|-|} 
 
 5 & product & $ 3.1\times10^{-2} $ &   & 55 & weed & $ 8.3\times10^{-3} $ \\ \hhline{|-|-|-|~|-|-|-|} 
 
 6 & farm & $ 2.9\times10^{-2} $ &   & 56 & matter & $ 8.2\times10^{-3} $ \\ \hhline{|-|-|-|~|-|-|-|} 
 
 7 & wheat & $ 2.4\times10^{-2} $ &   & 57 & diet & $ 8.2\times10^{-3} $ \\ \hhline{|-|-|-|~|-|-|-|} 
 
 8 & cultivar & $ 2.3\times10^{-2} $ &   & 58 & digest & $ 8.2\times10^{-3} $ \\
\end{longtable}
 \newpage 
 
 \clearpage 

\newpage 

\thispagestyle{plain} 

\vspace*{-1.6in}

\setlength{\LTleft}{-20cm plus -1fill}

\setlength{\LTright}{\LTleft} 

\begin{longtable}[h]{| C{.6cm}| L{5cm} |  R{1.6cm}  |  R{.3cm} |  C{.6cm}| L{5cm} |  R{1.6cm} | }

\caption{Agronomy}\\ 

\hhline{|-|-|-|~|-|-|-|}

\multicolumn{1}{|C{.6cm}|}{\textbf{No.}} & \multicolumn{1}{C{5cm}|}{\textbf{Word}} & \multicolumn{1}{C{1.6cm}|}{\textbf{RIG}} &   &\multicolumn{1}{C{.6cm}|}{\textbf{No.}} & \multicolumn{1}{C{5cm}|}{\textbf{Word}} & \multicolumn{1}{C{1.6cm}|}{\textbf{RIG}} \\ \hhline{|-|-|-|~|-|-|-|} 

\endhead 

\hhline{|-|-|-|~|-|-|-|}

\endfoot
 1 & plant & $ 1.3\times10^{-1} $ &   & 51 & aestivum & $ 1.4\times10^{-2} $ \\ \hhline{|-|-|-|~|-|-|-|} 
 
 2 & crop & $ 1.2\times10^{-1} $ &   & 52 & highest & $ 1.4\times10^{-2} $ \\ \hhline{|-|-|-|~|-|-|-|} 
 
 3 & cultivar & $ 9.3\times10^{-2} $ &   & 53 & winter & $ 1.4\times10^{-2} $ \\ \hhline{|-|-|-|~|-|-|-|} 
 
 4 & soil & $ 7.9\times10^{-2} $ &   & 54 & plot & $ 1.4\times10^{-2} $ \\ \hhline{|-|-|-|~|-|-|-|} 
 
 5 & yield & $ 6\times10^{-2} $ &   & 55 & total & $ 1.4\times10^{-2} $ \\ \hhline{|-|-|-|~|-|-|-|} 
 
 6 & seed & $ 4.5\times10^{-2} $ &   & 56 & zea & $ 1.4\times10^{-2} $ \\ \hhline{|-|-|-|~|-|-|-|} 
 
 7 & leaf & $ 4.4\times10^{-2} $ &   & 57 & farmer & $ 1.4\times10^{-2} $ \\ \hhline{|-|-|-|~|-|-|-|} 
 
 8 & breed & $ 4.2\times10^{-2} $ &   & 58 & pest & $ 1.4\times10^{-2} $ \\
\end{longtable}
 \newpage 
 
 \clearpage 

\newpage 

\thispagestyle{plain} 

\vspace*{-1.6in}

\setlength{\LTleft}{-20cm plus -1fill}

\setlength{\LTright}{\LTleft} 

\begin{longtable}[h]{| C{.6cm}| L{5cm} |  R{1.6cm}  |  R{.3cm} |  C{.6cm}| L{5cm} |  R{1.6cm} | }

\caption{Allergy}\\ 

\hhline{|-|-|-|~|-|-|-|}

\multicolumn{1}{|C{.6cm}|}{\textbf{No.}} & \multicolumn{1}{C{5cm}|}{\textbf{Word}} & \multicolumn{1}{C{1.6cm}|}{\textbf{RIG}} &   &\multicolumn{1}{C{.6cm}|}{\textbf{No.}} & \multicolumn{1}{C{5cm}|}{\textbf{Word}} & \multicolumn{1}{C{1.6cm}|}{\textbf{RIG}} \\ \hhline{|-|-|-|~|-|-|-|} 

\endhead 

\hhline{|-|-|-|~|-|-|-|}

\endfoot
 1 & asthma & $ 2.6\times10^{-1} $ &   & 51 & lung & $ 2\times10^{-2} $ \\ \hhline{|-|-|-|~|-|-|-|} 
 
 2 & allerg & $ 2.2\times10^{-1} $ &   & 52 & hyperrespons & $ 2\times10^{-2} $ \\ \hhline{|-|-|-|~|-|-|-|} 
 
 3 & allergi & $ 1.7\times10^{-1} $ &   & 53 & nasal & $ 1.9\times10^{-2} $ \\ \hhline{|-|-|-|~|-|-|-|} 
 
 4 & allergen & $ 1.4\times10^{-1} $ &   & 54 & th2 & $ 1.9\times10^{-2} $ \\ \hhline{|-|-|-|~|-|-|-|} 
 
 5 & conclus & $ 1.3\times10^{-1} $ &   & 55 & immunolog & $ 1.9\times10^{-2} $ \\ \hhline{|-|-|-|~|-|-|-|} 
 
 6 & background & $ 1.3\times10^{-1} $ &   & 56 & food & $ 1.9\times10^{-2} $ \\ \hhline{|-|-|-|~|-|-|-|} 
 
 7 & ige & $ 1.3\times10^{-1} $ &   & 57 & sever & $ 1.9\times10^{-2} $ \\ \hhline{|-|-|-|~|-|-|-|} 
 
 8 & patient & $ 8\times10^{-2} $ &   & 58 & eosinophilia & $ 1.8\times10^{-2} $ \\
\end{longtable}
 \newpage 
 
 \clearpage 

\newpage 

\thispagestyle{plain} 

\vspace*{-1.6in}

\setlength{\LTleft}{-20cm plus -1fill}

\setlength{\LTright}{\LTleft} 

\begin{longtable}[h]{| C{.6cm}| L{5cm} |  R{1.6cm}  |  R{.3cm} |  C{.6cm}| L{5cm} |  R{1.6cm} | }

\caption{Anatomy and Morphology}\\ 

\hhline{|-|-|-|~|-|-|-|}

\multicolumn{1}{|C{.6cm}|}{\textbf{No.}} & \multicolumn{1}{C{5cm}|}{\textbf{Word}} & \multicolumn{1}{C{1.6cm}|}{\textbf{RIG}} &   &\multicolumn{1}{C{.6cm}|}{\textbf{No.}} & \multicolumn{1}{C{5cm}|}{\textbf{Word}} & \multicolumn{1}{C{1.6cm}|}{\textbf{RIG}} \\ \hhline{|-|-|-|~|-|-|-|} 

\endhead 

\hhline{|-|-|-|~|-|-|-|}

\endfoot
 1 & anatom & $ 5.8\times10^{-2} $ &   & 51 & rat & $ 9.9\times10^{-3} $ \\ \hhline{|-|-|-|~|-|-|-|} 
 
 2 & anatomi & $ 4.8\times10^{-2} $ &   & 52 & microscopi & $ 9.9\times10^{-3} $ \\ \hhline{|-|-|-|~|-|-|-|} 
 
 3 & morpholog & $ 4\times10^{-2} $ &   & 53 & section & $ 9.8\times10^{-3} $ \\ \hhline{|-|-|-|~|-|-|-|} 
 
 4 & cadav & $ 3.8\times10^{-2} $ &   & 54 & anatomist & $ 9.6\times10^{-3} $ \\ \hhline{|-|-|-|~|-|-|-|} 
 
 5 & dissect & $ 3.2\times10^{-2} $ &   & 55 & bilater & $ 9.6\times10^{-3} $ \\ \hhline{|-|-|-|~|-|-|-|} 
 
 6 & posterior & $ 2.6\times10^{-2} $ &   & 56 & cartilag & $ 9.3\times10^{-3} $ \\ \hhline{|-|-|-|~|-|-|-|} 
 
 7 & neuron & $ 2.5\times10^{-2} $ &   & 57 & trunk & $ 9.2\times10^{-3} $ \\ \hhline{|-|-|-|~|-|-|-|} 
 
 8 & anterior & $ 2.5\times10^{-2} $ &   & 58 & fossa & $ 8.7\times10^{-3} $ \\
\end{longtable}
 \newpage 
 
 \clearpage 

\newpage 

\thispagestyle{plain} 

\vspace*{-1.6in}

\setlength{\LTleft}{-20cm plus -1fill}

\setlength{\LTright}{\LTleft} 

\begin{longtable}[h]{| C{.6cm}| L{5cm} |  R{1.6cm}  |  R{.3cm} |  C{.6cm}| L{5cm} |  R{1.6cm} | }

\caption{Andrology}\\ 

\hhline{|-|-|-|~|-|-|-|}

\multicolumn{1}{|C{.6cm}|}{\textbf{No.}} & \multicolumn{1}{C{5cm}|}{\textbf{Word}} & \multicolumn{1}{C{1.6cm}|}{\textbf{RIG}} &   &\multicolumn{1}{C{.6cm}|}{\textbf{No.}} & \multicolumn{1}{C{5cm}|}{\textbf{Word}} & \multicolumn{1}{C{1.6cm}|}{\textbf{RIG}} \\ \hhline{|-|-|-|~|-|-|-|} 

\endhead 

\hhline{|-|-|-|~|-|-|-|}

\endfoot
 1 & sperm & $ 2.2\times10^{-1} $ &   & 51 & antioxid & $ 1.6\times10^{-2} $ \\ \hhline{|-|-|-|~|-|-|-|} 
 
 2 & infertil & $ 1.2\times10^{-1} $ &   & 52 & treatment & $ 1.6\times10^{-2} $ \\ \hhline{|-|-|-|~|-|-|-|} 
 
 3 & testicular & $ 1.2\times10^{-1} $ &   & 53 & follicl & $ 1.5\times10^{-2} $ \\ \hhline{|-|-|-|~|-|-|-|} 
 
 4 & semen & $ 1.1\times10^{-1} $ &   & 54 & count & $ 1.5\times10^{-2} $ \\ \hhline{|-|-|-|~|-|-|-|} 
 
 5 & spermatozoa & $ 1.1\times10^{-1} $ &   & 55 & fsh & $ 1.5\times10^{-2} $ \\ \hhline{|-|-|-|~|-|-|-|} 
 
 6 & male & $ 9.9\times10^{-2} $ &   & 56 & cell & $ 1.5\times10^{-2} $ \\ \hhline{|-|-|-|~|-|-|-|} 
 
 7 & testi & $ 9.7\times10^{-2} $ &   & 57 & treat & $ 1.5\times10^{-2} $ \\ \hhline{|-|-|-|~|-|-|-|} 
 
 8 & motil & $ 8\times10^{-2} $ &   & 58 & dna & $ 1.5\times10^{-2} $ \\
\end{longtable}
 \newpage 
 
 \clearpage 

\newpage 

\thispagestyle{plain} 

\vspace*{-1.6in}

\setlength{\LTleft}{-20cm plus -1fill}

\setlength{\LTright}{\LTleft} 

\begin{longtable}[h]{| C{.6cm}| L{5cm} |  R{1.6cm}  |  R{.3cm} |  C{.6cm}| L{5cm} |  R{1.6cm} | }

\caption{Anesthesiology}\\ 

\hhline{|-|-|-|~|-|-|-|}

\multicolumn{1}{|C{.6cm}|}{\textbf{No.}} & \multicolumn{1}{C{5cm}|}{\textbf{Word}} & \multicolumn{1}{C{1.6cm}|}{\textbf{RIG}} &   &\multicolumn{1}{C{.6cm}|}{\textbf{No.}} & \multicolumn{1}{C{5cm}|}{\textbf{Word}} & \multicolumn{1}{C{1.6cm}|}{\textbf{RIG}} \\ \hhline{|-|-|-|~|-|-|-|} 

\endhead 

\hhline{|-|-|-|~|-|-|-|}

\endfoot
 1 & pain & $ 1.4\times10^{-1} $ &   & 51 & asa & $ 2\times10^{-2} $ \\ \hhline{|-|-|-|~|-|-|-|} 
 
 2 & patient & $ 1.2\times10^{-1} $ &   & 52 & befor & $ 2\times10^{-2} $ \\ \hhline{|-|-|-|~|-|-|-|} 
 
 3 & anesthesia & $ 9.7\times10^{-2} $ &   & 53 & intervent & $ 1.9\times10^{-2} $ \\ \hhline{|-|-|-|~|-|-|-|} 
 
 4 & conclus & $ 9\times10^{-2} $ &   & 54 & dure & $ 1.9\times10^{-2} $ \\ \hhline{|-|-|-|~|-|-|-|} 
 
 5 & surgeri & $ 8.4\times10^{-2} $ &   & 55 & nocicept & $ 1.9\times10^{-2} $ \\ \hhline{|-|-|-|~|-|-|-|} 
 
 6 & background & $ 8\times10^{-2} $ &   & 56 & airway & $ 1.9\times10^{-2} $ \\ \hhline{|-|-|-|~|-|-|-|} 
 
 7 & anesthet & $ 6\times10^{-2} $ &   & 57 & morphin & $ 1.8\times10^{-2} $ \\ \hhline{|-|-|-|~|-|-|-|} 
 
 8 & postop & $ 5.6\times10^{-2} $ &   & 58 & confid & $ 1.8\times10^{-2} $ \\
\end{longtable}
 \newpage 
 
 \clearpage 

\newpage 

\thispagestyle{plain} 

\vspace*{-1.6in}

\setlength{\LTleft}{-20cm plus -1fill}

\setlength{\LTright}{\LTleft} 

\begin{longtable}[h]{| C{.6cm}| L{5cm} |  R{1.6cm}  |  R{.3cm} |  C{.6cm}| L{5cm} |  R{1.6cm} | }

\caption{Anthropology}\\ 

\hhline{|-|-|-|~|-|-|-|}

\multicolumn{1}{|C{.6cm}|}{\textbf{No.}} & \multicolumn{1}{C{5cm}|}{\textbf{Word}} & \multicolumn{1}{C{1.6cm}|}{\textbf{RIG}} &   &\multicolumn{1}{C{.6cm}|}{\textbf{No.}} & \multicolumn{1}{C{5cm}|}{\textbf{Word}} & \multicolumn{1}{C{1.6cm}|}{\textbf{RIG}} \\ \hhline{|-|-|-|~|-|-|-|} 

\endhead 

\hhline{|-|-|-|~|-|-|-|}

\endfoot
 1 & archaeolog & $ 8.7\times10^{-2} $ &   & 51 & landscap & $ 1\times10^{-2} $ \\ \hhline{|-|-|-|~|-|-|-|} 
 
 2 & anthropolog & $ 4.7\times10^{-2} $ &   & 52 & radiocarbon & $ 9.8\times10^{-3} $ \\ \hhline{|-|-|-|~|-|-|-|} 
 
 3 & ethnograph & $ 4.5\times10^{-2} $ &   & 53 & stone & $ 9.8\times10^{-3} $ \\ \hhline{|-|-|-|~|-|-|-|} 
 
 4 & social & $ 4.3\times10^{-2} $ &   & 54 & live & $ 9.7\times10^{-3} $ \\ \hhline{|-|-|-|~|-|-|-|} 
 
 5 & polit & $ 3.1\times10^{-2} $ &   & 55 & chronolog & $ 9.6\times10^{-3} $ \\ \hhline{|-|-|-|~|-|-|-|} 
 
 6 & cultur & $ 3.1\times10^{-2} $ &   & 56 & individu & $ 9.5\times10^{-3} $ \\ \hhline{|-|-|-|~|-|-|-|} 
 
 7 & argu & $ 2.7\times10^{-2} $ &   & 57 & world & $ 9.1\times10^{-3} $ \\ \hhline{|-|-|-|~|-|-|-|} 
 
 8 & articl & $ 2.4\times10^{-2} $ &   & 58 & earli & $ 9.1\times10^{-3} $ \\
\end{longtable}
 \newpage 
 
 \clearpage 

\newpage 

\thispagestyle{plain} 

\vspace*{-1.6in}

\setlength{\LTleft}{-20cm plus -1fill}

\setlength{\LTright}{\LTleft} 

\begin{longtable}[h]{| C{.6cm}| L{5cm} |  R{1.6cm}  |  R{.3cm} |  C{.6cm}| L{5cm} |  R{1.6cm} | }

\caption{Archaeology}\\ 

\hhline{|-|-|-|~|-|-|-|}

\multicolumn{1}{|C{.6cm}|}{\textbf{No.}} & \multicolumn{1}{C{5cm}|}{\textbf{Word}} & \multicolumn{1}{C{1.6cm}|}{\textbf{RIG}} &   &\multicolumn{1}{C{.6cm}|}{\textbf{No.}} & \multicolumn{1}{C{5cm}|}{\textbf{Word}} & \multicolumn{1}{C{1.6cm}|}{\textbf{RIG}} \\ \hhline{|-|-|-|~|-|-|-|} 

\endhead 

\hhline{|-|-|-|~|-|-|-|}

\endfoot
 1 & archaeolog & $ 2.9\times10^{-1} $ &   & 51 & holocen & $ 1.4\times10^{-2} $ \\ \hhline{|-|-|-|~|-|-|-|} 
 
 2 & excav & $ 8.9\times10^{-2} $ &   & 52 & funerari & $ 1.4\times10^{-2} $ \\ \hhline{|-|-|-|~|-|-|-|} 
 
 3 & site & $ 7.6\times10^{-2} $ &   & 53 & cell & $ 1.4\times10^{-2} $ \\ \hhline{|-|-|-|~|-|-|-|} 
 
 4 & centuri & $ 7.2\times10^{-2} $ &   & 54 & region & $ 1.3\times10^{-2} $ \\ \hhline{|-|-|-|~|-|-|-|} 
 
 5 & neolith & $ 5.1\times10^{-2} $ &   & 55 & museum & $ 1.3\times10^{-2} $ \\ \hhline{|-|-|-|~|-|-|-|} 
 
 6 & archaeologist & $ 4.9\times10^{-2} $ &   & 56 & antiqu & $ 1.3\times10^{-2} $ \\ \hhline{|-|-|-|~|-|-|-|} 
 
 7 & date & $ 4.9\times10^{-2} $ &   & 57 & faunal & $ 1.3\times10^{-2} $ \\ \hhline{|-|-|-|~|-|-|-|} 
 
 8 & bronz & $ 4.8\times10^{-2} $ &   & 58 & south & $ 1.3\times10^{-2} $ \\
\end{longtable}
 \newpage 
 
 \clearpage 

\newpage 

\thispagestyle{plain} 

\vspace*{-1.6in}

\setlength{\LTleft}{-20cm plus -1fill}

\setlength{\LTright}{\LTleft} 

\begin{longtable}[h]{| C{.6cm}| L{5cm} |  R{1.6cm}  |  R{.3cm} |  C{.6cm}| L{5cm} |  R{1.6cm} | }

\caption{Architecture}\\ 

\hhline{|-|-|-|~|-|-|-|}

\multicolumn{1}{|C{.6cm}|}{\textbf{No.}} & \multicolumn{1}{C{5cm}|}{\textbf{Word}} & \multicolumn{1}{C{1.6cm}|}{\textbf{RIG}} &   &\multicolumn{1}{C{.6cm}|}{\textbf{No.}} & \multicolumn{1}{C{5cm}|}{\textbf{Word}} & \multicolumn{1}{C{1.6cm}|}{\textbf{RIG}} \\ \hhline{|-|-|-|~|-|-|-|} 

\endhead 

\hhline{|-|-|-|~|-|-|-|}

\endfoot
 1 & architectur & $ 1.3\times10^{-1} $ &   & 51 & rate & $ 8.6\times10^{-3} $ \\ \hhline{|-|-|-|~|-|-|-|} 
 
 2 & build & $ 9.6\times10^{-2} $ &   & 52 & timber & $ 8.6\times10^{-3} $ \\ \hhline{|-|-|-|~|-|-|-|} 
 
 3 & architect & $ 6.5\times10^{-2} $ &   & 53 & explor & $ 8.6\times10^{-3} $ \\ \hhline{|-|-|-|~|-|-|-|} 
 
 4 & heritag & $ 6.2\times10^{-2} $ &   & 54 & church & $ 8.5\times10^{-3} $ \\ \hhline{|-|-|-|~|-|-|-|} 
 
 5 & urban & $ 4.8\times10^{-2} $ &   & 55 & studio & $ 8.3\times10^{-3} $ \\ \hhline{|-|-|-|~|-|-|-|} 
 
 6 & vernacular & $ 4.5\times10^{-2} $ &   & 56 & protein & $ 8.3\times10^{-3} $ \\ \hhline{|-|-|-|~|-|-|-|} 
 
 7 & hous & $ 4.4\times10^{-2} $ &   & 57 & context & $ 8.2\times10^{-3} $ \\ \hhline{|-|-|-|~|-|-|-|} 
 
 8 & centuri & $ 4.2\times10^{-2} $ &   & 58 & place & $ 8.2\times10^{-3} $ \\
\end{longtable}
 \newpage 
 
 \clearpage 

\newpage 

\thispagestyle{plain} 

\vspace*{-1.6in}

\setlength{\LTleft}{-20cm plus -1fill}

\setlength{\LTright}{\LTleft} 

\begin{longtable}[h]{| C{.6cm}| L{5cm} |  R{1.6cm}  |  R{.3cm} |  C{.6cm}| L{5cm} |  R{1.6cm} | }

\caption{Area Studies}\\ 

\hhline{|-|-|-|~|-|-|-|}

\multicolumn{1}{|C{.6cm}|}{\textbf{No.}} & \multicolumn{1}{C{5cm}|}{\textbf{Word}} & \multicolumn{1}{C{1.6cm}|}{\textbf{RIG}} &   &\multicolumn{1}{C{.6cm}|}{\textbf{No.}} & \multicolumn{1}{C{5cm}|}{\textbf{Word}} & \multicolumn{1}{C{1.6cm}|}{\textbf{RIG}} \\ \hhline{|-|-|-|~|-|-|-|} 

\endhead 

\hhline{|-|-|-|~|-|-|-|}

\endfoot
 1 & polit & $ 1.5\times10^{-1} $ &   & 51 & elect & $ 1.5\times10^{-2} $ \\ \hhline{|-|-|-|~|-|-|-|} 
 
 2 & articl & $ 1.1\times10^{-1} $ &   & 52 & peopl & $ 1.5\times10^{-2} $ \\ \hhline{|-|-|-|~|-|-|-|} 
 
 3 & argu & $ 6.5\times10^{-2} $ &   & 53 & patient & $ 1.4\times10^{-2} $ \\ \hhline{|-|-|-|~|-|-|-|} 
 
 4 & govern & $ 5.1\times10^{-2} $ &   & 54 & militari & $ 1.4\times10^{-2} $ \\ \hhline{|-|-|-|~|-|-|-|} 
 
 5 & polici & $ 4.4\times10^{-2} $ &   & 55 & south & $ 1.4\times10^{-2} $ \\ \hhline{|-|-|-|~|-|-|-|} 
 
 6 & result & $ 3.7\times10^{-2} $ &   & 56 & protest & $ 1.4\times10^{-2} $ \\ \hhline{|-|-|-|~|-|-|-|} 
 
 7 & social & $ 3.7\times10^{-2} $ &   & 57 & religi & $ 1.4\times10^{-2} $ \\ \hhline{|-|-|-|~|-|-|-|} 
 
 8 & nation & $ 3.5\times10^{-2} $ &   & 58 & union & $ 1.4\times10^{-2} $ \\
\end{longtable}
 \newpage 
 
 \clearpage 

\newpage 

\thispagestyle{plain} 

\vspace*{-1.6in}

\setlength{\LTleft}{-20cm plus -1fill}

\setlength{\LTright}{\LTleft} 

\begin{longtable}[h]{| C{.6cm}| L{5cm} |  R{1.6cm}  |  R{.3cm} |  C{.6cm}| L{5cm} |  R{1.6cm} | }

\caption{Art}\\ 

\hhline{|-|-|-|~|-|-|-|}

\multicolumn{1}{|C{.6cm}|}{\textbf{No.}} & \multicolumn{1}{C{5cm}|}{\textbf{Word}} & \multicolumn{1}{C{1.6cm}|}{\textbf{RIG}} &   &\multicolumn{1}{C{.6cm}|}{\textbf{No.}} & \multicolumn{1}{C{5cm}|}{\textbf{Word}} & \multicolumn{1}{C{1.6cm}|}{\textbf{RIG}} \\ \hhline{|-|-|-|~|-|-|-|} 

\endhead 

\hhline{|-|-|-|~|-|-|-|}

\endfoot
 1 & artist & $ 1.2\times10^{-1} $ &   & 51 & compar & $ 1\times10^{-2} $ \\ \hhline{|-|-|-|~|-|-|-|} 
 
 2 & art & $ 1.1\times10^{-1} $ &   & 52 & creat & $ 1\times10^{-2} $ \\ \hhline{|-|-|-|~|-|-|-|} 
 
 3 & centuri & $ 6.1\times10^{-2} $ &   & 53 & interpret & $ 9.4\times10^{-3} $ \\ \hhline{|-|-|-|~|-|-|-|} 
 
 4 & paint & $ 6\times10^{-2} $ &   & 54 & question & $ 9.4\times10^{-3} $ \\ \hhline{|-|-|-|~|-|-|-|} 
 
 5 & articl & $ 5.4\times10^{-2} $ &   & 55 & control & $ 9.3\times10^{-3} $ \\ \hhline{|-|-|-|~|-|-|-|} 
 
 6 & aesthet & $ 4.4\times10^{-2} $ &   & 56 & rate & $ 9.3\times10^{-3} $ \\ \hhline{|-|-|-|~|-|-|-|} 
 
 7 & artwork & $ 4.3\times10^{-2} $ &   & 57 & decreas & $ 8.8\times10^{-3} $ \\ \hhline{|-|-|-|~|-|-|-|} 
 
 8 & cultur & $ 3.4\times10^{-2} $ &   & 58 & embodi & $ 8.5\times10^{-3} $ \\
\end{longtable}
 \newpage 
 
 \clearpage 

\newpage 

\thispagestyle{plain} 

\vspace*{-1.6in}

\setlength{\LTleft}{-20cm plus -1fill}

\setlength{\LTright}{\LTleft} 

\begin{longtable}[h]{| C{.6cm}| L{5cm} |  R{1.6cm}  |  R{.3cm} |  C{.6cm}| L{5cm} |  R{1.6cm} | }

\caption{Asian Studies}\\ 

\hhline{|-|-|-|~|-|-|-|}

\multicolumn{1}{|C{.6cm}|}{\textbf{No.}} & \multicolumn{1}{C{5cm}|}{\textbf{Word}} & \multicolumn{1}{C{1.6cm}|}{\textbf{RIG}} &   &\multicolumn{1}{C{.6cm}|}{\textbf{No.}} & \multicolumn{1}{C{5cm}|}{\textbf{Word}} & \multicolumn{1}{C{1.6cm}|}{\textbf{RIG}} \\ \hhline{|-|-|-|~|-|-|-|} 

\endhead 

\hhline{|-|-|-|~|-|-|-|}

\endfoot
 1 & articl & $ 9.8\times10^{-2} $ &   & 51 & inscript & $ 1.5\times10^{-2} $ \\ \hhline{|-|-|-|~|-|-|-|} 
 
 2 & argu & $ 6.5\times10^{-2} $ &   & 52 & model & $ 1.4\times10^{-2} $ \\ \hhline{|-|-|-|~|-|-|-|} 
 
 3 & polit & $ 6.3\times10^{-2} $ &   & 53 & religion & $ 1.4\times10^{-2} $ \\ \hhline{|-|-|-|~|-|-|-|} 
 
 4 & centuri & $ 6.2\times10^{-2} $ &   & 54 & examin & $ 1.4\times10^{-2} $ \\ \hhline{|-|-|-|~|-|-|-|} 
 
 5 & buddhist & $ 5.2\times10^{-2} $ &   & 55 & china & $ 1.4\times10^{-2} $ \\ \hhline{|-|-|-|~|-|-|-|} 
 
 6 & result & $ 4.5\times10^{-2} $ &   & 56 & india & $ 1.4\times10^{-2} $ \\ \hhline{|-|-|-|~|-|-|-|} 
 
 7 & confucian & $ 4.1\times10^{-2} $ &   & 57 & offici & $ 1.4\times10^{-2} $ \\ \hhline{|-|-|-|~|-|-|-|} 
 
 8 & religi & $ 4\times10^{-2} $ &   & 58 & tangut & $ 1.4\times10^{-2} $ \\
\end{longtable}
 \newpage 
 
 \clearpage 

\newpage 

\thispagestyle{plain} 

\vspace*{-1.6in}

\setlength{\LTleft}{-20cm plus -1fill}

\setlength{\LTright}{\LTleft} 

\begin{longtable}[h]{| C{.6cm}| L{5cm} |  R{1.6cm}  |  R{.3cm} |  C{.6cm}| L{5cm} |  R{1.6cm} | }

\caption{Astronomy and Astrophysics}\\ 

\hhline{|-|-|-|~|-|-|-|}

\multicolumn{1}{|C{.6cm}|}{\textbf{No.}} & \multicolumn{1}{C{5cm}|}{\textbf{Word}} & \multicolumn{1}{C{1.6cm}|}{\textbf{RIG}} &   &\multicolumn{1}{C{.6cm}|}{\textbf{No.}} & \multicolumn{1}{C{5cm}|}{\textbf{Word}} & \multicolumn{1}{C{1.6cm}|}{\textbf{RIG}} \\ \hhline{|-|-|-|~|-|-|-|} 

\endhead 

\hhline{|-|-|-|~|-|-|-|}

\endfoot
 1 & star & $ 1.5\times10^{-1} $ &   & 51 & ngc & $ 1.8\times10^{-2} $ \\ \hhline{|-|-|-|~|-|-|-|} 
 
 2 & galaxi & $ 1.1\times10^{-1} $ &   & 52 & black & $ 1.8\times10^{-2} $ \\ \hhline{|-|-|-|~|-|-|-|} 
 
 3 & telescop & $ 9.9\times10^{-2} $ &   & 53 & gev & $ 1.8\times10^{-2} $ \\ \hhline{|-|-|-|~|-|-|-|} 
 
 4 & stellar & $ 9.8\times10^{-2} $ &   & 54 & baryon & $ 1.8\times10^{-2} $ \\ \hhline{|-|-|-|~|-|-|-|} 
 
 5 & galact & $ 6.6\times10^{-2} $ &   & 55 & spectra & $ 1.8\times10^{-2} $ \\ \hhline{|-|-|-|~|-|-|-|} 
 
 6 & mass & $ 5.9\times10^{-2} $ &   & 56 & near & $ 1.7\times10^{-2} $ \\ \hhline{|-|-|-|~|-|-|-|} 
 
 7 & cosmolog & $ 5.4\times10^{-2} $ &   & 57 & relativist & $ 1.6\times10^{-2} $ \\ \hhline{|-|-|-|~|-|-|-|} 
 
 8 & luminos & $ 5.1\times10^{-2} $ &   & 58 & space & $ 1.6\times10^{-2} $ \\
\end{longtable}
 \newpage 
 
 \clearpage 

\newpage 

\thispagestyle{plain} 

\vspace*{-1.6in}

\setlength{\LTleft}{-20cm plus -1fill}

\setlength{\LTright}{\LTleft} 

\begin{longtable}[h]{| C{.6cm}| L{5cm} |  R{1.6cm}  |  R{.3cm} |  C{.6cm}| L{5cm} |  R{1.6cm} | }

\caption{Audiology and Speech-Language Pathology}\\ 

\hhline{|-|-|-|~|-|-|-|}

\multicolumn{1}{|C{.6cm}|}{\textbf{No.}} & \multicolumn{1}{C{5cm}|}{\textbf{Word}} & \multicolumn{1}{C{1.6cm}|}{\textbf{RIG}} &   &\multicolumn{1}{C{.6cm}|}{\textbf{No.}} & \multicolumn{1}{C{5cm}|}{\textbf{Word}} & \multicolumn{1}{C{1.6cm}|}{\textbf{RIG}} \\ \hhline{|-|-|-|~|-|-|-|} 

\endhead 

\hhline{|-|-|-|~|-|-|-|}

\endfoot
 1 & speech & $ 1.4\times10^{-1} $ &   & 51 & utter & $ 1.2\times10^{-2} $ \\ \hhline{|-|-|-|~|-|-|-|} 
 
 2 & hear & $ 1.3\times10^{-1} $ &   & 52 & across & $ 1.2\times10^{-2} $ \\ \hhline{|-|-|-|~|-|-|-|} 
 
 3 & acoust & $ 8.9\times10^{-2} $ &   & 53 & pathologist & $ 1.2\times10^{-2} $ \\ \hhline{|-|-|-|~|-|-|-|} 
 
 4 & listen & $ 7.9\times10^{-2} $ &   & 54 & aphasia & $ 1.2\times10^{-2} $ \\ \hhline{|-|-|-|~|-|-|-|} 
 
 5 & sound & $ 7.2\times10^{-2} $ &   & 55 & implant & $ 1.2\times10^{-2} $ \\ \hhline{|-|-|-|~|-|-|-|} 
 
 6 & auditori & $ 6\times10^{-2} $ &   & 56 & loud & $ 1.2\times10^{-2} $ \\ \hhline{|-|-|-|~|-|-|-|} 
 
 7 & languag & $ 5.8\times10^{-2} $ &   & 57 & binaur & $ 1.2\times10^{-2} $ \\ \hhline{|-|-|-|~|-|-|-|} 
 
 8 & cochlear & $ 5.2\times10^{-2} $ &   & 58 & otoacoust & $ 1.1\times10^{-2} $ \\
\end{longtable}
 \newpage 
 
 \clearpage 

\newpage 

\thispagestyle{plain} 

\vspace*{-1.6in}

\setlength{\LTleft}{-20cm plus -1fill}

\setlength{\LTright}{\LTleft} 

\begin{longtable}[h]{| C{.6cm}| L{5cm} |  R{1.6cm}  |  R{.3cm} |  C{.6cm}| L{5cm} |  R{1.6cm} | }

\caption{Automation and Control Systems}\\ 

\hhline{|-|-|-|~|-|-|-|}

\multicolumn{1}{|C{.6cm}|}{\textbf{No.}} & \multicolumn{1}{C{5cm}|}{\textbf{Word}} & \multicolumn{1}{C{1.6cm}|}{\textbf{RIG}} &   &\multicolumn{1}{C{.6cm}|}{\textbf{No.}} & \multicolumn{1}{C{5cm}|}{\textbf{Word}} & \multicolumn{1}{C{1.6cm}|}{\textbf{RIG}} \\ \hhline{|-|-|-|~|-|-|-|} 

\endhead 

\hhline{|-|-|-|~|-|-|-|}

\endfoot
 1 & paper & $ 9.7\times10^{-2} $ &   & 51 & scheme & $ 1\times10^{-2} $ \\ \hhline{|-|-|-|~|-|-|-|} 
 
 2 & propos & $ 7\times10^{-2} $ &   & 52 & examin & $ 1\times10^{-2} $ \\ \hhline{|-|-|-|~|-|-|-|} 
 
 3 & system & $ 5.7\times10^{-2} $ &   & 53 & trajectori & $ 9.9\times10^{-3} $ \\ \hhline{|-|-|-|~|-|-|-|} 
 
 4 & control & $ 4.4\times10^{-2} $ &   & 54 & year & $ 9.9\times10^{-3} $ \\ \hhline{|-|-|-|~|-|-|-|} 
 
 5 & robot & $ 3.9\times10^{-2} $ &   & 55 & constraint & $ 9.5\times10^{-3} $ \\ \hhline{|-|-|-|~|-|-|-|} 
 
 6 & problem & $ 3.8\times10^{-2} $ &   & 56 & background & $ 9.4\times10^{-3} $ \\ \hhline{|-|-|-|~|-|-|-|} 
 
 7 & algorithm & $ 3.7\times10^{-2} $ &   & 57 & converg & $ 9.3\times10^{-3} $ \\ \hhline{|-|-|-|~|-|-|-|} 
 
 8 & simul & $ 3.6\times10^{-2} $ &   & 58 & motion & $ 9.2\times10^{-3} $ \\
\end{longtable}
 \newpage 
 
 \clearpage 

\newpage 

\thispagestyle{plain} 

\vspace*{-1.6in}

\setlength{\LTleft}{-20cm plus -1fill}

\setlength{\LTright}{\LTleft} 

\begin{longtable}[h]{| C{.6cm}| L{5cm} |  R{1.6cm}  |  R{.3cm} |  C{.6cm}| L{5cm} |  R{1.6cm} | }

\caption{Behavioral Sciences}\\ 

\hhline{|-|-|-|~|-|-|-|}

\multicolumn{1}{|C{.6cm}|}{\textbf{No.}} & \multicolumn{1}{C{5cm}|}{\textbf{Word}} & \multicolumn{1}{C{1.6cm}|}{\textbf{RIG}} &   &\multicolumn{1}{C{.6cm}|}{\textbf{No.}} & \multicolumn{1}{C{5cm}|}{\textbf{Word}} & \multicolumn{1}{C{1.6cm}|}{\textbf{RIG}} \\ \hhline{|-|-|-|~|-|-|-|} 

\endhead 

\hhline{|-|-|-|~|-|-|-|}

\endfoot
 1 & behavior & $ 5.2\times10^{-2} $ &   & 51 & bird & $ 1.2\times10^{-2} $ \\ \hhline{|-|-|-|~|-|-|-|} 
 
 2 & anim & $ 4.4\times10^{-2} $ &   & 52 & eat & $ 1.2\times10^{-2} $ \\ \hhline{|-|-|-|~|-|-|-|} 
 
 3 & task & $ 4\times10^{-2} $ &   & 53 & corticosteron & $ 1.2\times10^{-2} $ \\ \hhline{|-|-|-|~|-|-|-|} 
 
 4 & male & $ 3.7\times10^{-2} $ &   & 54 & forag & $ 1.2\times10^{-2} $ \\ \hhline{|-|-|-|~|-|-|-|} 
 
 5 & suggest & $ 3.7\times10^{-2} $ &   & 55 & either & $ 1.2\times10^{-2} $ \\ \hhline{|-|-|-|~|-|-|-|} 
 
 6 & cognit & $ 3.5\times10^{-2} $ &   & 56 & amygdala & $ 1.2\times10^{-2} $ \\ \hhline{|-|-|-|~|-|-|-|} 
 
 7 & brain & $ 3.4\times10^{-2} $ &   & 57 & examin & $ 1.2\times10^{-2} $ \\ \hhline{|-|-|-|~|-|-|-|} 
 
 8 & whether & $ 3.1\times10^{-2} $ &   & 58 & affect & $ 1.1\times10^{-2} $ \\
\end{longtable}
 \newpage 
 
 \clearpage 

\newpage 

\thispagestyle{plain} 

\vspace*{-1.6in}

\setlength{\LTleft}{-20cm plus -1fill}

\setlength{\LTright}{\LTleft} 

\begin{longtable}[h]{| C{.6cm}| L{5cm} |  R{1.6cm}  |  R{.3cm} |  C{.6cm}| L{5cm} |  R{1.6cm} | }

\caption{Biochemical Research Methods}\\ 

\hhline{|-|-|-|~|-|-|-|}

\multicolumn{1}{|C{.6cm}|}{\textbf{No.}} & \multicolumn{1}{C{5cm}|}{\textbf{Word}} & \multicolumn{1}{C{1.6cm}|}{\textbf{RIG}} &   &\multicolumn{1}{C{.6cm}|}{\textbf{No.}} & \multicolumn{1}{C{5cm}|}{\textbf{Word}} & \multicolumn{1}{C{1.6cm}|}{\textbf{RIG}} \\ \hhline{|-|-|-|~|-|-|-|} 

\endhead 

\hhline{|-|-|-|~|-|-|-|}

\endfoot
 1 & chromatographi & $ 5.9\times10^{-2} $ &   & 51 & specif & $ 9.5\times10^{-3} $ \\ \hhline{|-|-|-|~|-|-|-|} 
 
 2 & spectrometri & $ 5.1\times10^{-2} $ &   & 52 & drug & $ 9.4\times10^{-3} $ \\ \hhline{|-|-|-|~|-|-|-|} 
 
 3 & protein & $ 4.2\times10^{-2} $ &   & 53 & analysi & $ 9.4\times10^{-3} $ \\ \hhline{|-|-|-|~|-|-|-|} 
 
 4 & proteom & $ 3.4\times10^{-2} $ &   & 54 & quadrupol & $ 9.4\times10^{-3} $ \\ \hhline{|-|-|-|~|-|-|-|} 
 
 5 & mass & $ 2.8\times10^{-2} $ &   & 55 & molecular & $ 9.2\times10^{-3} $ \\ \hhline{|-|-|-|~|-|-|-|} 
 
 6 & detect & $ 2.8\times10^{-2} $ &   & 56 & molecul & $ 9\times10^{-3} $ \\ \hhline{|-|-|-|~|-|-|-|} 
 
 7 & biolog & $ 2.8\times10^{-2} $ &   & 57 & rapid & $ 8.7\times10^{-3} $ \\ \hhline{|-|-|-|~|-|-|-|} 
 
 8 & quantif & $ 2.7\times10^{-2} $ &   & 58 & limit & $ 8.7\times10^{-3} $ \\
\end{longtable}
 \newpage 
 
 \clearpage 

\newpage 

\thispagestyle{plain} 

\vspace*{-1.6in}

\setlength{\LTleft}{-20cm plus -1fill}

\setlength{\LTright}{\LTleft} 

\begin{longtable}[h]{| C{.6cm}| L{5cm} |  R{1.6cm}  |  R{.3cm} |  C{.6cm}| L{5cm} |  R{1.6cm} | }

\caption{Biochemistry and Molecular Biology}\\ 

\hhline{|-|-|-|~|-|-|-|}

\multicolumn{1}{|C{.6cm}|}{\textbf{No.}} & \multicolumn{1}{C{5cm}|}{\textbf{Word}} & \multicolumn{1}{C{1.6cm}|}{\textbf{RIG}} &   &\multicolumn{1}{C{.6cm}|}{\textbf{No.}} & \multicolumn{1}{C{5cm}|}{\textbf{Word}} & \multicolumn{1}{C{1.6cm}|}{\textbf{RIG}} \\ \hhline{|-|-|-|~|-|-|-|} 

\endhead 

\hhline{|-|-|-|~|-|-|-|}

\endfoot
 1 & protein & $ 1.1\times10^{-1} $ &   & 51 & involv & $ 1.3\times10^{-2} $ \\ \hhline{|-|-|-|~|-|-|-|} 
 
 2 & cell & $ 7\times10^{-2} $ &   & 52 & signal & $ 1.3\times10^{-2} $ \\ \hhline{|-|-|-|~|-|-|-|} 
 
 3 & bind & $ 6.3\times10^{-2} $ &   & 53 & apoptosi & $ 1.2\times10^{-2} $ \\ \hhline{|-|-|-|~|-|-|-|} 
 
 4 & express & $ 5.5\times10^{-2} $ &   & 54 & molecul & $ 1.2\times10^{-2} $ \\ \hhline{|-|-|-|~|-|-|-|} 
 
 5 & gene & $ 5.3\times10^{-2} $ &   & 55 & affin & $ 1.2\times10^{-2} $ \\ \hhline{|-|-|-|~|-|-|-|} 
 
 6 & regul & $ 5.2\times10^{-2} $ &   & 56 & termin & $ 1.2\times10^{-2} $ \\ \hhline{|-|-|-|~|-|-|-|} 
 
 7 & activ & $ 5.2\times10^{-2} $ &   & 57 & domain & $ 1.2\times10^{-2} $ \\ \hhline{|-|-|-|~|-|-|-|} 
 
 8 & inhibit & $ 3.8\times10^{-2} $ &   & 58 & mitochondri & $ 1.1\times10^{-2} $ \\
\end{longtable}
 \newpage 
 
 \clearpage 

\newpage 

\thispagestyle{plain} 

\vspace*{-1.6in}

\setlength{\LTleft}{-20cm plus -1fill}

\setlength{\LTright}{\LTleft} 

\begin{longtable}[h]{| C{.6cm}| L{5cm} |  R{1.6cm}  |  R{.3cm} |  C{.6cm}| L{5cm} |  R{1.6cm} | }

\caption{Biodiversity Conservation}\\ 

\hhline{|-|-|-|~|-|-|-|}

\multicolumn{1}{|C{.6cm}|}{\textbf{No.}} & \multicolumn{1}{C{5cm}|}{\textbf{Word}} & \multicolumn{1}{C{1.6cm}|}{\textbf{RIG}} &   &\multicolumn{1}{C{.6cm}|}{\textbf{No.}} & \multicolumn{1}{C{5cm}|}{\textbf{Word}} & \multicolumn{1}{C{1.6cm}|}{\textbf{RIG}} \\ \hhline{|-|-|-|~|-|-|-|} 

\endhead 

\hhline{|-|-|-|~|-|-|-|}

\endfoot
 1 & speci & $ 1.8\times10^{-1} $ &   & 51 & geograph & $ 1.4\times10^{-2} $ \\ \hhline{|-|-|-|~|-|-|-|} 
 
 2 & habitat & $ 1.1\times10^{-1} $ &   & 52 & south & $ 1.4\times10^{-2} $ \\ \hhline{|-|-|-|~|-|-|-|} 
 
 3 & conserv & $ 9.6\times10^{-2} $ &   & 53 & distribut & $ 1.4\times10^{-2} $ \\ \hhline{|-|-|-|~|-|-|-|} 
 
 4 & biodivers & $ 6.2\times10^{-2} $ &   & 54 & individu & $ 1.4\times10^{-2} $ \\ \hhline{|-|-|-|~|-|-|-|} 
 
 5 & ecolog & $ 5.9\times10^{-2} $ &   & 55 & region & $ 1.4\times10^{-2} $ \\ \hhline{|-|-|-|~|-|-|-|} 
 
 6 & popul & $ 5.7\times10^{-2} $ &   & 56 & cover & $ 1.4\times10^{-2} $ \\ \hhline{|-|-|-|~|-|-|-|} 
 
 7 & ecosystem & $ 5.5\times10^{-2} $ &   & 57 & eastern & $ 1.3\times10^{-2} $ \\ \hhline{|-|-|-|~|-|-|-|} 
 
 8 & forest & $ 5.1\times10^{-2} $ &   & 58 & park & $ 1.3\times10^{-2} $ \\
\end{longtable}
 \newpage 
 
 \clearpage 

\newpage 

\thispagestyle{plain} 

\vspace*{-1.6in}

\setlength{\LTleft}{-20cm plus -1fill}

\setlength{\LTright}{\LTleft} 

\begin{longtable}[h]{| C{.6cm}| L{5cm} |  R{1.6cm}  |  R{.3cm} |  C{.6cm}| L{5cm} |  R{1.6cm} | }

\caption{Biology}\\ 

\hhline{|-|-|-|~|-|-|-|}

\multicolumn{1}{|C{.6cm}|}{\textbf{No.}} & \multicolumn{1}{C{5cm}|}{\textbf{Word}} & \multicolumn{1}{C{1.6cm}|}{\textbf{RIG}} &   &\multicolumn{1}{C{.6cm}|}{\textbf{No.}} & \multicolumn{1}{C{5cm}|}{\textbf{Word}} & \multicolumn{1}{C{1.6cm}|}{\textbf{RIG}} \\ \hhline{|-|-|-|~|-|-|-|} 

\endhead 

\hhline{|-|-|-|~|-|-|-|}

\endfoot
 1 & speci & $ 2.3\times10^{-2} $ &   & 51 & growth & $ 3.9\times10^{-3} $ \\ \hhline{|-|-|-|~|-|-|-|} 
 
 2 & paper & $ 1.4\times10^{-2} $ &   & 52 & conserv & $ 3.8\times10^{-3} $ \\ \hhline{|-|-|-|~|-|-|-|} 
 
 3 & anim & $ 1.2\times10^{-2} $ &   & 53 & pattern & $ 3.7\times10^{-3} $ \\ \hhline{|-|-|-|~|-|-|-|} 
 
 4 & cell & $ 1.1\times10^{-2} $ &   & 54 & transcript & $ 3.7\times10^{-3} $ \\ \hhline{|-|-|-|~|-|-|-|} 
 
 5 & biolog & $ 1.1\times10^{-2} $ &   & 55 & patient & $ 3.6\times10^{-3} $ \\ \hhline{|-|-|-|~|-|-|-|} 
 
 6 & protein & $ 1.1\times10^{-2} $ &   & 56 & vertebr & $ 3.6\times10^{-3} $ \\ \hhline{|-|-|-|~|-|-|-|} 
 
 7 & gene & $ 9.9\times10^{-3} $ &   & 57 & prey & $ 3.6\times10^{-3} $ \\ \hhline{|-|-|-|~|-|-|-|} 
 
 8 & evolutionari & $ 9\times10^{-3} $ &   & 58 & mediat & $ 3.6\times10^{-3} $ \\
\end{longtable}
 \newpage 
 
 \clearpage 

\newpage 

\thispagestyle{plain} 

\vspace*{-1.6in}

\setlength{\LTleft}{-20cm plus -1fill}

\setlength{\LTright}{\LTleft} 

\begin{longtable}[h]{| C{.6cm}| L{5cm} |  R{1.6cm}  |  R{.3cm} |  C{.6cm}| L{5cm} |  R{1.6cm} | }

\caption{Biophysics}\\ 

\hhline{|-|-|-|~|-|-|-|}

\multicolumn{1}{|C{.6cm}|}{\textbf{No.}} & \multicolumn{1}{C{5cm}|}{\textbf{Word}} & \multicolumn{1}{C{1.6cm}|}{\textbf{RIG}} &   &\multicolumn{1}{C{.6cm}|}{\textbf{No.}} & \multicolumn{1}{C{5cm}|}{\textbf{Word}} & \multicolumn{1}{C{1.6cm}|}{\textbf{RIG}} \\ \hhline{|-|-|-|~|-|-|-|} 

\endhead 

\hhline{|-|-|-|~|-|-|-|}

\endfoot
 1 & protein & $ 5.6\times10^{-2} $ &   & 51 & receptor & $ 6.6\times10^{-3} $ \\ \hhline{|-|-|-|~|-|-|-|} 
 
 2 & bind & $ 4.2\times10^{-2} $ &   & 52 & signal & $ 6.5\times10^{-3} $ \\ \hhline{|-|-|-|~|-|-|-|} 
 
 3 & cell & $ 3.8\times10^{-2} $ &   & 53 & phosphoryl & $ 6.4\times10^{-3} $ \\ \hhline{|-|-|-|~|-|-|-|} 
 
 4 & membran & $ 2\times10^{-2} $ &   & 54 & target & $ 6.3\times10^{-3} $ \\ \hhline{|-|-|-|~|-|-|-|} 
 
 5 & paper & $ 1.9\times10^{-2} $ &   & 55 & fold & $ 6.3\times10^{-3} $ \\ \hhline{|-|-|-|~|-|-|-|} 
 
 6 & conform & $ 1.6\times10^{-2} $ &   & 56 & bilay & $ 6.2\times10^{-3} $ \\ \hhline{|-|-|-|~|-|-|-|} 
 
 7 & angstrom & $ 1.6\times10^{-2} $ &   & 57 & suggest & $ 5.8\times10^{-3} $ \\ \hhline{|-|-|-|~|-|-|-|} 
 
 8 & molecular & $ 1.6\times10^{-2} $ &   & 58 & kinas & $ 5.8\times10^{-3} $ \\
\end{longtable}
 \newpage 
 
 \clearpage 

\newpage 

\thispagestyle{plain} 

\vspace*{-1.6in}

\setlength{\LTleft}{-20cm plus -1fill}

\setlength{\LTright}{\LTleft} 

\begin{longtable}[h]{| C{.6cm}| L{5cm} |  R{1.6cm}  |  R{.3cm} |  C{.6cm}| L{5cm} |  R{1.6cm} | }

\caption{Biotechnology and Applied Microbiology}\\ 

\hhline{|-|-|-|~|-|-|-|}

\multicolumn{1}{|C{.6cm}|}{\textbf{No.}} & \multicolumn{1}{C{5cm}|}{\textbf{Word}} & \multicolumn{1}{C{1.6cm}|}{\textbf{RIG}} &   &\multicolumn{1}{C{.6cm}|}{\textbf{No.}} & \multicolumn{1}{C{5cm}|}{\textbf{Word}} & \multicolumn{1}{C{1.6cm}|}{\textbf{RIG}} \\ \hhline{|-|-|-|~|-|-|-|} 

\endhead 

\hhline{|-|-|-|~|-|-|-|}

\endfoot
 1 & gene & $ 5.4\times10^{-2} $ &   & 51 & microalga & $ 8.3\times10^{-3} $ \\ \hhline{|-|-|-|~|-|-|-|} 
 
 2 & strain & $ 3\times10^{-2} $ &   & 52 & biofuel & $ 8.2\times10^{-3} $ \\ \hhline{|-|-|-|~|-|-|-|} 
 
 3 & genom & $ 3\times10^{-2} $ &   & 53 & potenti & $ 8.1\times10^{-3} $ \\ \hhline{|-|-|-|~|-|-|-|} 
 
 4 & product & $ 2.9\times10^{-2} $ &   & 54 & concentr & $ 8\times10^{-3} $ \\ \hhline{|-|-|-|~|-|-|-|} 
 
 5 & ferment & $ 2.9\times10^{-2} $ &   & 55 & virus & $ 7.6\times10^{-3} $ \\ \hhline{|-|-|-|~|-|-|-|} 
 
 6 & cell & $ 2.9\times10^{-2} $ &   & 56 & fold & $ 7.6\times10^{-3} $ \\ \hhline{|-|-|-|~|-|-|-|} 
 
 7 & protein & $ 2.8\times10^{-2} $ &   & 57 & sugar & $ 7.5\times10^{-3} $ \\ \hhline{|-|-|-|~|-|-|-|} 
 
 8 & sequenc & $ 2.7\times10^{-2} $ &   & 58 & 16s & $ 7.3\times10^{-3} $ \\
\end{longtable}
 \newpage 
 
 \clearpage 

\newpage 

\thispagestyle{plain} 

\vspace*{-1.6in}

\setlength{\LTleft}{-20cm plus -1fill}

\setlength{\LTright}{\LTleft} 

\begin{longtable}[h]{| C{.6cm}| L{5cm} |  R{1.6cm}  |  R{.3cm} |  C{.6cm}| L{5cm} |  R{1.6cm} | }

\caption{Business}\\ 

\hhline{|-|-|-|~|-|-|-|}

\multicolumn{1}{|C{.6cm}|}{\textbf{No.}} & \multicolumn{1}{C{5cm}|}{\textbf{Word}} & \multicolumn{1}{C{1.6cm}|}{\textbf{RIG}} &   &\multicolumn{1}{C{.6cm}|}{\textbf{No.}} & \multicolumn{1}{C{5cm}|}{\textbf{Word}} & \multicolumn{1}{C{1.6cm}|}{\textbf{RIG}} \\ \hhline{|-|-|-|~|-|-|-|} 

\endhead 

\hhline{|-|-|-|~|-|-|-|}

\endfoot
 1 & firm & $ 8.4\times10^{-2} $ &   & 51 & surfac & $ 1.2\times10^{-2} $ \\ \hhline{|-|-|-|~|-|-|-|} 
 
 2 & market & $ 8\times10^{-2} $ &   & 52 & advertis & $ 1.2\times10^{-2} $ \\ \hhline{|-|-|-|~|-|-|-|} 
 
 3 & busi & $ 7.8\times10^{-2} $ &   & 53 & focus & $ 1.2\times10^{-2} $ \\ \hhline{|-|-|-|~|-|-|-|} 
 
 4 & research & $ 6.2\times10^{-2} $ &   & 54 & treatment & $ 1.2\times10^{-2} $ \\ \hhline{|-|-|-|~|-|-|-|} 
 
 5 & compani & $ 5.9\times10^{-2} $ &   & 55 & influenc & $ 1.2\times10^{-2} $ \\ \hhline{|-|-|-|~|-|-|-|} 
 
 6 & manag & $ 4.6\times10^{-2} $ &   & 56 & develop & $ 1.2\times10^{-2} $ \\ \hhline{|-|-|-|~|-|-|-|} 
 
 7 & innov & $ 3.6\times10^{-2} $ &   & 57 & govern & $ 1.2\times10^{-2} $ \\ \hhline{|-|-|-|~|-|-|-|} 
 
 8 & organiz & $ 3.6\times10^{-2} $ &   & 58 & sale & $ 1.1\times10^{-2} $ \\
\end{longtable}
 \newpage 
 
 \clearpage 

\newpage 

\thispagestyle{plain} 

\vspace*{-1.6in}

\setlength{\LTleft}{-20cm plus -1fill}

\setlength{\LTright}{\LTleft} 

\begin{longtable}[h]{| C{.6cm}| L{5cm} |  R{1.6cm}  |  R{.3cm} |  C{.6cm}| L{5cm} |  R{1.6cm} | }

\caption{Business, Finance}\\ 

\hhline{|-|-|-|~|-|-|-|}

\multicolumn{1}{|C{.6cm}|}{\textbf{No.}} & \multicolumn{1}{C{5cm}|}{\textbf{Word}} & \multicolumn{1}{C{1.6cm}|}{\textbf{RIG}} &   &\multicolumn{1}{C{.6cm}|}{\textbf{No.}} & \multicolumn{1}{C{5cm}|}{\textbf{Word}} & \multicolumn{1}{C{1.6cm}|}{\textbf{RIG}} \\ \hhline{|-|-|-|~|-|-|-|} 

\endhead 

\hhline{|-|-|-|~|-|-|-|}

\endfoot
 1 & market & $ 1.1\times10^{-1} $ &   & 51 & default & $ 1.2\times10^{-2} $ \\ \hhline{|-|-|-|~|-|-|-|} 
 
 2 & financi & $ 1.1\times10^{-1} $ &   & 52 & account & $ 1.2\times10^{-2} $ \\ \hhline{|-|-|-|~|-|-|-|} 
 
 3 & firm & $ 7\times10^{-2} $ &   & 53 & surfac & $ 1.2\times10^{-2} $ \\ \hhline{|-|-|-|~|-|-|-|} 
 
 4 & price & $ 6.1\times10^{-2} $ &   & 54 & privat & $ 1.2\times10^{-2} $ \\ \hhline{|-|-|-|~|-|-|-|} 
 
 5 & investor & $ 5.9\times10^{-2} $ &   & 55 & method & $ 1.1\times10^{-2} $ \\ \hhline{|-|-|-|~|-|-|-|} 
 
 6 & stock & $ 5.4\times10^{-2} $ &   & 56 & european & $ 1.1\times10^{-2} $ \\ \hhline{|-|-|-|~|-|-|-|} 
 
 7 & bank & $ 4.6\times10^{-2} $ &   & 57 & treatment & $ 1.1\times10^{-2} $ \\ \hhline{|-|-|-|~|-|-|-|} 
 
 8 & asset & $ 4.5\times10^{-2} $ &   & 58 & clinic & $ 1.1\times10^{-2} $ \\
\end{longtable}
 \newpage 
 
 \clearpage 

\newpage 

\thispagestyle{plain} 

\vspace*{-1.6in}

\setlength{\LTleft}{-20cm plus -1fill}

\setlength{\LTright}{\LTleft} 

\begin{longtable}[h]{| C{.6cm}| L{5cm} |  R{1.6cm}  |  R{.3cm} |  C{.6cm}| L{5cm} |  R{1.6cm} | }

\caption{Cardiac and Cardiovascular Systems}\\ 

\hhline{|-|-|-|~|-|-|-|}

\multicolumn{1}{|C{.6cm}|}{\textbf{No.}} & \multicolumn{1}{C{5cm}|}{\textbf{Word}} & \multicolumn{1}{C{1.6cm}|}{\textbf{RIG}} &   &\multicolumn{1}{C{.6cm}|}{\textbf{No.}} & \multicolumn{1}{C{5cm}|}{\textbf{Word}} & \multicolumn{1}{C{1.6cm}|}{\textbf{RIG}} \\ \hhline{|-|-|-|~|-|-|-|} 

\endhead 

\hhline{|-|-|-|~|-|-|-|}

\endfoot
 1 & patient & $ 1.8\times10^{-1} $ &   & 51 & hospit & $ 2.8\times10^{-2} $ \\ \hhline{|-|-|-|~|-|-|-|} 
 
 2 & cardiac & $ 1.6\times10^{-1} $ &   & 52 & median & $ 2.7\times10^{-2} $ \\ \hhline{|-|-|-|~|-|-|-|} 
 
 3 & conclus & $ 1.5\times10^{-1} $ &   & 53 & cardiomyopathi & $ 2.7\times10^{-2} $ \\ \hhline{|-|-|-|~|-|-|-|} 
 
 4 & coronari & $ 1.4\times10^{-1} $ &   & 54 & revascular & $ 2.7\times10^{-2} $ \\ \hhline{|-|-|-|~|-|-|-|} 
 
 5 & heart & $ 1.4\times10^{-1} $ &   & 55 & multivari & $ 2.7\times10^{-2} $ \\ \hhline{|-|-|-|~|-|-|-|} 
 
 6 & ventricular & $ 1.4\times10^{-1} $ &   & 56 & confid & $ 2.7\times10^{-2} $ \\ \hhline{|-|-|-|~|-|-|-|} 
 
 7 & myocardi & $ 1.1\times10^{-1} $ &   & 57 & hypertens & $ 2.7\times10^{-2} $ \\ \hhline{|-|-|-|~|-|-|-|} 
 
 8 & arteri & $ 1.1\times10^{-1} $ &   & 58 & month & $ 2.6\times10^{-2} $ \\
\end{longtable}
 \newpage 
 
 \clearpage 

\newpage 

\thispagestyle{plain} 

\vspace*{-1.6in}

\setlength{\LTleft}{-20cm plus -1fill}

\setlength{\LTright}{\LTleft} 

\begin{longtable}[h]{| C{.6cm}| L{5cm} |  R{1.6cm}  |  R{.3cm} |  C{.6cm}| L{5cm} |  R{1.6cm} | }

\caption{Cell and Tissue Engineering}\\ 

\hhline{|-|-|-|~|-|-|-|}

\multicolumn{1}{|C{.6cm}|}{\textbf{No.}} & \multicolumn{1}{C{5cm}|}{\textbf{Word}} & \multicolumn{1}{C{1.6cm}|}{\textbf{RIG}} &   &\multicolumn{1}{C{.6cm}|}{\textbf{No.}} & \multicolumn{1}{C{5cm}|}{\textbf{Word}} & \multicolumn{1}{C{1.6cm}|}{\textbf{RIG}} \\ \hhline{|-|-|-|~|-|-|-|} 

\endhead 

\hhline{|-|-|-|~|-|-|-|}

\endfoot
 1 & stem & $ 2.4\times10^{-1} $ &   & 51 & growth & $ 1.9\times10^{-2} $ \\ \hhline{|-|-|-|~|-|-|-|} 
 
 2 & cell & $ 2\times10^{-1} $ &   & 52 & transcript & $ 1.9\times10^{-2} $ \\ \hhline{|-|-|-|~|-|-|-|} 
 
 3 & mesenchym & $ 1\times10^{-1} $ &   & 53 & cocultur & $ 1.9\times10^{-2} $ \\ \hhline{|-|-|-|~|-|-|-|} 
 
 4 & pluripot & $ 9.3\times10^{-2} $ &   & 54 & extracellular & $ 1.8\times10^{-2} $ \\ \hhline{|-|-|-|~|-|-|-|} 
 
 5 & differenti & $ 8\times10^{-2} $ &   & 55 & factor & $ 1.8\times10^{-2} $ \\ \hhline{|-|-|-|~|-|-|-|} 
 
 6 & tissu & $ 7.5\times10^{-2} $ &   & 56 & regul & $ 1.8\times10^{-2} $ \\ \hhline{|-|-|-|~|-|-|-|} 
 
 7 & progenitor & $ 7.2\times10^{-2} $ &   & 57 & protein & $ 1.8\times10^{-2} $ \\ \hhline{|-|-|-|~|-|-|-|} 
 
 8 & bone & $ 6.9\times10^{-2} $ &   & 58 & sox2 & $ 1.8\times10^{-2} $ \\
\end{longtable}
 \newpage 
 
 \clearpage 

\newpage 

\thispagestyle{plain} 

\vspace*{-1.6in}

\setlength{\LTleft}{-20cm plus -1fill}

\setlength{\LTright}{\LTleft} 

\begin{longtable}[h]{| C{.6cm}| L{5cm} |  R{1.6cm}  |  R{.3cm} |  C{.6cm}| L{5cm} |  R{1.6cm} | }

\caption{Cell Biology}\\ 

\hhline{|-|-|-|~|-|-|-|}

\multicolumn{1}{|C{.6cm}|}{\textbf{No.}} & \multicolumn{1}{C{5cm}|}{\textbf{Word}} & \multicolumn{1}{C{1.6cm}|}{\textbf{RIG}} &   &\multicolumn{1}{C{.6cm}|}{\textbf{No.}} & \multicolumn{1}{C{5cm}|}{\textbf{Word}} & \multicolumn{1}{C{1.6cm}|}{\textbf{RIG}} \\ \hhline{|-|-|-|~|-|-|-|} 

\endhead 

\hhline{|-|-|-|~|-|-|-|}

\endfoot
 1 & cell & $ 1.7\times10^{-1} $ &   & 51 & beta & $ 1.7\times10^{-2} $ \\ \hhline{|-|-|-|~|-|-|-|} 
 
 2 & express & $ 1.1\times10^{-1} $ &   & 52 & factor & $ 1.6\times10^{-2} $ \\ \hhline{|-|-|-|~|-|-|-|} 
 
 3 & protein & $ 1\times10^{-1} $ &   & 53 & extracellular & $ 1.6\times10^{-2} $ \\ \hhline{|-|-|-|~|-|-|-|} 
 
 4 & regul & $ 9.5\times10^{-2} $ &   & 54 & downregul & $ 1.6\times10^{-2} $ \\ \hhline{|-|-|-|~|-|-|-|} 
 
 5 & induc & $ 6.2\times10^{-2} $ &   & 55 & phenotyp & $ 1.6\times10^{-2} $ \\ \hhline{|-|-|-|~|-|-|-|} 
 
 6 & inhibit & $ 5.6\times10^{-2} $ &   & 56 & mutat & $ 1.5\times10^{-2} $ \\ \hhline{|-|-|-|~|-|-|-|} 
 
 7 & pathway & $ 5.3\times10^{-2} $ &   & 57 & embryon & $ 1.5\times10^{-2} $ \\ \hhline{|-|-|-|~|-|-|-|} 
 
 8 & activ & $ 5.2\times10^{-2} $ &   & 58 & involv & $ 1.5\times10^{-2} $ \\
\end{longtable}
 \newpage 
 
 \clearpage 

\newpage 

\thispagestyle{plain} 

\vspace*{-1.6in}

\setlength{\LTleft}{-20cm plus -1fill}

\setlength{\LTright}{\LTleft} 

\begin{longtable}[h]{| C{.6cm}| L{5cm} |  R{1.6cm}  |  R{.3cm} |  C{.6cm}| L{5cm} |  R{1.6cm} | }

\caption{Chemistry, Analytical}\\ 

\hhline{|-|-|-|~|-|-|-|}

\multicolumn{1}{|C{.6cm}|}{\textbf{No.}} & \multicolumn{1}{C{5cm}|}{\textbf{Word}} & \multicolumn{1}{C{1.6cm}|}{\textbf{RIG}} &   &\multicolumn{1}{C{.6cm}|}{\textbf{No.}} & \multicolumn{1}{C{5cm}|}{\textbf{Word}} & \multicolumn{1}{C{1.6cm}|}{\textbf{RIG}} \\ \hhline{|-|-|-|~|-|-|-|} 

\endhead 

\hhline{|-|-|-|~|-|-|-|}

\endfoot
 1 & detect & $ 8.4\times10^{-2} $ &   & 51 & label & $ 1.4\times10^{-2} $ \\ \hhline{|-|-|-|~|-|-|-|} 
 
 2 & chromatographi & $ 8\times10^{-2} $ &   & 52 & simpl & $ 1.4\times10^{-2} $ \\ \hhline{|-|-|-|~|-|-|-|} 
 
 3 & spectrometri & $ 6.9\times10^{-2} $ &   & 53 & voltammetri & $ 1.3\times10^{-2} $ \\ \hhline{|-|-|-|~|-|-|-|} 
 
 4 & sampl & $ 5.3\times10^{-2} $ &   & 54 & use & $ 1.3\times10^{-2} $ \\ \hhline{|-|-|-|~|-|-|-|} 
 
 5 & analyt & $ 4.9\times10^{-2} $ &   & 55 & microfluid & $ 1.3\times10^{-2} $ \\ \hhline{|-|-|-|~|-|-|-|} 
 
 6 & liquid & $ 4.6\times10^{-2} $ &   & 56 & solvent & $ 1.3\times10^{-2} $ \\ \hhline{|-|-|-|~|-|-|-|} 
 
 7 & sensit & $ 4\times10^{-2} $ &   & 57 & methanol & $ 1.3\times10^{-2} $ \\ \hhline{|-|-|-|~|-|-|-|} 
 
 8 & limit & $ 3.9\times10^{-2} $ &   & 58 & esi & $ 1.3\times10^{-2} $ \\
\end{longtable}
 \newpage 
 
 \clearpage 

\newpage 

\thispagestyle{plain} 

\vspace*{-1.6in}

\setlength{\LTleft}{-20cm plus -1fill}

\setlength{\LTright}{\LTleft} 

\begin{longtable}[h]{| C{.6cm}| L{5cm} |  R{1.6cm}  |  R{.3cm} |  C{.6cm}| L{5cm} |  R{1.6cm} | }

\caption{Chemistry, Applied}\\ 

\hhline{|-|-|-|~|-|-|-|}

\multicolumn{1}{|C{.6cm}|}{\textbf{No.}} & \multicolumn{1}{C{5cm}|}{\textbf{Word}} & \multicolumn{1}{C{1.6cm}|}{\textbf{RIG}} &   &\multicolumn{1}{C{.6cm}|}{\textbf{No.}} & \multicolumn{1}{C{5cm}|}{\textbf{Word}} & \multicolumn{1}{C{1.6cm}|}{\textbf{RIG}} \\ \hhline{|-|-|-|~|-|-|-|} 

\endhead 

\hhline{|-|-|-|~|-|-|-|}

\endfoot
 1 & acid & $ 3.9\times10^{-2} $ &   & 51 & contain & $ 7.5\times10^{-3} $ \\ \hhline{|-|-|-|~|-|-|-|} 
 
 2 & prepar & $ 2.3\times10^{-2} $ &   & 52 & algorithm & $ 7.4\times10^{-3} $ \\ \hhline{|-|-|-|~|-|-|-|} 
 
 3 & compound & $ 2.3\times10^{-2} $ &   & 53 & chitosan & $ 7.3\times10^{-3} $ \\ \hhline{|-|-|-|~|-|-|-|} 
 
 4 & patient & $ 1.9\times10^{-2} $ &   & 54 & problem & $ 7.3\times10^{-3} $ \\ \hhline{|-|-|-|~|-|-|-|} 
 
 5 & catalyst & $ 1.8\times10^{-2} $ &   & 55 & zeolit & $ 7.1\times10^{-3} $ \\ \hhline{|-|-|-|~|-|-|-|} 
 
 6 & content & $ 1.8\times10^{-2} $ &   & 56 & liquid & $ 7\times10^{-3} $ \\ \hhline{|-|-|-|~|-|-|-|} 
 
 7 & oil & $ 1.8\times10^{-2} $ &   & 57 & fatti & $ 6.9\times10^{-3} $ \\ \hhline{|-|-|-|~|-|-|-|} 
 
 8 & chromatographi & $ 1.8\times10^{-2} $ &   & 58 & gas & $ 6.7\times10^{-3} $ \\
\end{longtable}
 \newpage 
 
 \input{tablecodes/RIG1_100_34.txt} 
 
 \input{tablecodes/RIG1_100_35.txt} 
 
 \input{tablecodes/RIG1_100_36.txt} 
 
 \input{tablecodes/RIG1_100_37.txt} 
 
 \input{tablecodes/RIG1_100_38.txt} 
 
 \input{tablecodes/RIG1_100_39.txt} 
 
 \input{tablecodes/RIG1_100_40.txt} 
 
 \input{tablecodes/RIG1_100_41.txt} 
 
 \input{tablecodes/RIG1_100_42.txt} 
 
 \input{tablecodes/RIG1_100_43.txt} 
 
 \input{tablecodes/RIG1_100_44.txt} 
 
 \input{tablecodes/RIG1_100_45.txt} 
 
 \input{tablecodes/RIG1_100_46.txt} 
 
 \input{tablecodes/RIG1_100_47.txt} 
 
 \input{tablecodes/RIG1_100_48.txt} 
 
 \input{tablecodes/RIG1_100_49.txt} 
 
 \input{tablecodes/RIG1_100_50.txt} 
 
 \input{tablecodes/RIG1_100_51.txt} 
 
 \input{tablecodes/RIG1_100_52.txt} 
 
 \input{tablecodes/RIG1_100_53.txt} 
 
 \input{tablecodes/RIG1_100_54.txt} 
 
 \input{tablecodes/RIG1_100_55.txt} 
 
 \input{tablecodes/RIG1_100_56.txt} 
 
 \input{tablecodes/RIG1_100_57.txt} 
 
 \input{tablecodes/RIG1_100_58.txt} 
 
 \input{tablecodes/RIG1_100_59.txt} 
 
 \input{tablecodes/RIG1_100_60.txt} 
 
 \input{tablecodes/RIG1_100_61.txt} 
 
 \input{tablecodes/RIG1_100_62.txt} 
 
 \input{tablecodes/RIG1_100_63.txt} 
 
 \input{tablecodes/RIG1_100_64.txt} 
 
 \input{tablecodes/RIG1_100_65.txt} 
 
 \input{tablecodes/RIG1_100_66.txt} 
 
 \input{tablecodes/RIG1_100_67.txt} 
 
 \input{tablecodes/RIG1_100_68.txt} 
 
 \input{tablecodes/RIG1_100_69.txt} 
 
 \input{tablecodes/RIG1_100_70.txt} 
 
 \input{tablecodes/RIG1_100_71.txt} 
 
 \input{tablecodes/RIG1_100_72.txt} 
 
 \input{tablecodes/RIG1_100_73.txt} 
 
 \input{tablecodes/RIG1_100_74.txt} 
 
 \input{tablecodes/RIG1_100_75.txt} 
 
 \input{tablecodes/RIG1_100_76.txt} 
 
 \input{tablecodes/RIG1_100_77.txt} 
 
 \input{tablecodes/RIG1_100_78.txt} 
 
 \input{tablecodes/RIG1_100_79.txt} 
 
 \input{tablecodes/RIG1_100_80.txt} 
 
 \input{tablecodes/RIG1_100_81.txt} 
 
 \input{tablecodes/RIG1_100_82.txt} 
 
 \input{tablecodes/RIG1_100_83.txt} 
 
 \input{tablecodes/RIG1_100_84.txt} 
 
 \input{tablecodes/RIG1_100_85.txt} 
 
 \input{tablecodes/RIG1_100_86.txt} 
 
 \input{tablecodes/RIG1_100_87.txt} 
 
 \input{tablecodes/RIG1_100_88.txt} 
 
 \input{tablecodes/RIG1_100_89.txt} 
 
 \input{tablecodes/RIG1_100_90.txt} 
 
 \input{tablecodes/RIG1_100_91.txt} 
 
 \input{tablecodes/RIG1_100_92.txt} 
 
 \input{tablecodes/RIG1_100_93.txt} 
 
 \input{tablecodes/RIG1_100_94.txt} 
 
 \input{tablecodes/RIG1_100_95.txt} 
 
 \input{tablecodes/RIG1_100_96.txt} 
 
 \input{tablecodes/RIG1_100_97.txt} 
 
 \input{tablecodes/RIG1_100_98.txt} 
 
 \input{tablecodes/RIG1_100_99.txt} 
 
 \input{tablecodes/RIG1_100_100.txt} 
 
 \input{tablecodes/RIG1_100_101.txt} 
 
 \input{tablecodes/RIG1_100_102.txt} 
 
 \input{tablecodes/RIG1_100_103.txt} 
 
 \input{tablecodes/RIG1_100_104.txt} 
 
 \input{tablecodes/RIG1_100_105.txt} 
 
 \input{tablecodes/RIG1_100_106.txt} 
 
 \input{tablecodes/RIG1_100_107.txt} 
 
 \input{tablecodes/RIG1_100_108.txt} 
 
 \input{tablecodes/RIG1_100_109.txt} 
 
 \input{tablecodes/RIG1_100_110.txt} 
 
 \input{tablecodes/RIG1_100_111.txt} 
 
 \input{tablecodes/RIG1_100_112.txt} 
 
 \input{tablecodes/RIG1_100_113.txt} 
 
 \input{tablecodes/RIG1_100_114.txt} 
 
 \input{tablecodes/RIG1_100_115.txt} 
 
 \input{tablecodes/RIG1_100_116.txt} 
 
 \input{tablecodes/RIG1_100_117.txt} 
 
 \input{tablecodes/RIG1_100_118.txt} 
 
 \input{tablecodes/RIG1_100_119.txt} 
 
 \input{tablecodes/RIG1_100_120.txt} 
 
 \input{tablecodes/RIG1_100_121.txt} 
 
 \input{tablecodes/RIG1_100_122.txt} 
 
 \input{tablecodes/RIG1_100_123.txt} 
 
 \input{tablecodes/RIG1_100_124.txt} 
 
 \input{tablecodes/RIG1_100_125.txt} 
 
 \input{tablecodes/RIG1_100_126.txt} 
 
 \input{tablecodes/RIG1_100_127.txt} 
 
 \input{tablecodes/RIG1_100_128.txt} 
 
 \input{tablecodes/RIG1_100_129.txt} 
 
 \input{tablecodes/RIG1_100_130.txt} 
 
 \input{tablecodes/RIG1_100_131.txt} 
 
 \input{tablecodes/RIG1_100_132.txt} 
 
 \input{tablecodes/RIG1_100_133.txt} 
 
 \input{tablecodes/RIG1_100_134.txt} 
 
 \input{tablecodes/RIG1_100_135.txt} 
 
 \input{tablecodes/RIG1_100_136.txt} 
 
 \input{tablecodes/RIG1_100_137.txt} 
 
 \input{tablecodes/RIG1_100_138.txt} 
 
 \input{tablecodes/RIG1_100_139.txt} 
 
 \input{tablecodes/RIG1_100_140.txt} 
 
 \input{tablecodes/RIG1_100_141.txt} 
 
 \input{tablecodes/RIG1_100_142.txt} 
 
 \input{tablecodes/RIG1_100_143.txt} 
 
 \input{tablecodes/RIG1_100_144.txt} 
 
 \input{tablecodes/RIG1_100_145.txt} 
 
 \input{tablecodes/RIG1_100_146.txt} 
 
 \input{tablecodes/RIG1_100_147.txt} 
 
 \input{tablecodes/RIG1_100_148.txt} 
 
 \input{tablecodes/RIG1_100_149.txt} 
 
 \input{tablecodes/RIG1_100_150.txt} 
 
 \input{tablecodes/RIG1_100_151.txt} 
 
 \input{tablecodes/RIG1_100_152.txt} 
 
 \input{tablecodes/RIG1_100_153.txt} 
 
 \input{tablecodes/RIG1_100_154.txt} 
 
 \input{tablecodes/RIG1_100_155.txt} 
 
 \input{tablecodes/RIG1_100_156.txt} 
 
 \input{tablecodes/RIG1_100_157.txt} 
 
 \input{tablecodes/RIG1_100_158.txt} 
 
 \input{tablecodes/RIG1_100_159.txt} 
 
 \input{tablecodes/RIG1_100_160.txt} 
 
 \input{tablecodes/RIG1_100_161.txt} 
 
 \input{tablecodes/RIG1_100_162.txt} 
 
 \input{tablecodes/RIG1_100_163.txt} 
 
 \input{tablecodes/RIG1_100_164.txt} 
 
 \input{tablecodes/RIG1_100_165.txt} 
 
 \input{tablecodes/RIG1_100_166.txt} 
 
 \input{tablecodes/RIG1_100_167.txt} 
 
 \input{tablecodes/RIG1_100_168.txt} 
 
 \input{tablecodes/RIG1_100_169.txt} 
 
 \input{tablecodes/RIG1_100_170.txt} 
 
 \input{tablecodes/RIG1_100_171.txt} 
 
 \input{tablecodes/RIG1_100_172.txt} 
 
 \input{tablecodes/RIG1_100_173.txt} 
 
 \input{tablecodes/RIG1_100_174.txt} 
 
 \input{tablecodes/RIG1_100_175.txt} 
 
 \input{tablecodes/RIG1_100_176.txt} 
 
 \input{tablecodes/RIG1_100_177.txt} 
 
 \input{tablecodes/RIG1_100_178.txt} 
 
 \input{tablecodes/RIG1_100_179.txt} 
 
 \input{tablecodes/RIG1_100_180.txt} 
 
 \input{tablecodes/RIG1_100_181.txt} 
 
 \input{tablecodes/RIG1_100_182.txt} 
 
 \input{tablecodes/RIG1_100_183.txt} 
 
 \input{tablecodes/RIG1_100_184.txt} 
 
 \input{tablecodes/RIG1_100_185.txt} 
 
 \input{tablecodes/RIG1_100_186.txt} 
 
 \input{tablecodes/RIG1_100_187.txt} 
 
 \input{tablecodes/RIG1_100_188.txt} 
 
 \input{tablecodes/RIG1_100_189.txt} 
 
 \input{tablecodes/RIG1_100_190.txt} 
 
 \input{tablecodes/RIG1_100_191.txt} 
 
 \input{tablecodes/RIG1_100_192.txt} 
 
 \input{tablecodes/RIG1_100_193.txt} 
 
 \input{tablecodes/RIG1_100_194.txt} 
 
 \input{tablecodes/RIG1_100_195.txt} 
 
 \input{tablecodes/RIG1_100_196.txt} 
 
 \input{tablecodes/RIG1_100_197.txt} 
 
 \input{tablecodes/RIG1_100_198.txt} 
 
 \input{tablecodes/RIG1_100_199.txt} 
 
 \input{tablecodes/RIG1_100_200.txt} 
 
 \input{tablecodes/RIG1_100_201.txt} 
 
 \input{tablecodes/RIG1_100_202.txt} 
 
 \input{tablecodes/RIG1_100_203.txt} 
 
 \input{tablecodes/RIG1_100_204.txt} 
 
 \input{tablecodes/RIG1_100_205.txt} 
 
 \input{tablecodes/RIG1_100_206.txt} 
 
 \input{tablecodes/RIG1_100_207.txt} 
 
 \input{tablecodes/RIG1_100_208.txt} 
 
 \input{tablecodes/RIG1_100_209.txt} 
 
 \input{tablecodes/RIG1_100_210.txt} 
 
 \input{tablecodes/RIG1_100_211.txt} 
 
 \input{tablecodes/RIG1_100_212.txt} 
 
 \input{tablecodes/RIG1_100_213.txt} 
 
 \input{tablecodes/RIG1_100_214.txt} 
 
 \input{tablecodes/RIG1_100_215.txt} 
 
 \input{tablecodes/RIG1_100_216.txt} 
 
 \input{tablecodes/RIG1_100_217.txt} 
 
 \input{tablecodes/RIG1_100_218.txt} 
 
 \input{tablecodes/RIG1_100_219.txt} 
 
 \input{tablecodes/RIG1_100_220.txt} 
 
 \input{tablecodes/RIG1_100_221.txt} 
 
 \input{tablecodes/RIG1_100_222.txt} 
 
 \input{tablecodes/RIG1_100_223.txt} 
 
 \input{tablecodes/RIG1_100_224.txt} 
 
 \input{tablecodes/RIG1_100_225.txt} 
 
 \input{tablecodes/RIG1_100_226.txt} 
 
 \input{tablecodes/RIG1_100_227.txt} 
 
 \input{tablecodes/RIG1_100_228.txt} 
 
 \input{tablecodes/RIG1_100_229.txt} 
 
 \input{tablecodes/RIG1_100_230.txt} 
 
 \input{tablecodes/RIG1_100_231.txt} 
 
 \input{tablecodes/RIG1_100_232.txt} 
 
 \input{tablecodes/RIG1_100_233.txt} 
 
 \input{tablecodes/RIG1_100_234.txt} 
 
 \input{tablecodes/RIG1_100_235.txt} 
 
 \input{tablecodes/RIG1_100_236.txt} 
 
 \input{tablecodes/RIG1_100_237.txt} 
 
 \input{tablecodes/RIG1_100_238.txt} 
 
 \input{tablecodes/RIG1_100_239.txt} 
 
 \input{tablecodes/RIG1_100_240.txt} 
 
 \input{tablecodes/RIG1_100_241.txt} 
 
 \input{tablecodes/RIG1_100_242.txt} 
 
 \input{tablecodes/RIG1_100_243.txt} 
 
 \input{tablecodes/RIG1_100_244.txt} 
 
 \input{tablecodes/RIG1_100_245.txt} 
 
 \input{tablecodes/RIG1_100_246.txt} 
 
 \input{tablecodes/RIG1_100_247.txt} 
 
 \input{tablecodes/RIG1_100_248.txt} 
 
 \input{tablecodes/RIG1_100_249.txt} 
 
 \input{tablecodes/RIG1_100_250.txt} 
 
 \input{tablecodes/RIG1_100_251.txt} 
 
 \input{tablecodes/RIG1_100_252.txt}

\end{document}